\documentclass[a4paper,12pt,twoside,final]{huthesis}
\usepackage{epsfig,bm,epsf,float}

\usepackage[ruled,vlined]{algorithm2e}
\usepackage{algorithmic}
\usepackage{tabularx, booktabs} 
\newcolumntype{Y}{>{\centering\arraybackslash}X}

\usepackage{epsfig}
\usepackage{graphicx}
\usepackage{cite}
\usepackage{color}
\usepackage{setspace}
\usepackage{subcaption}
\usepackage{afterpage}
\usepackage{emptypage}
\usepackage{enumitem} 

\usepackage[pagebackref=false,breaklinks=true,bookmarks=false]{hyperref}
\usepackage{array}
\newcolumntype{M}{>{\centering\arraybackslash}m{16em}}
\newcolumntype{N}{>{\centering\arraybackslash}m{5em}}
\newcolumntype{P}{>{\centering\arraybackslash}m{8em}}
\newcolumntype{Q}{>{\centering\arraybackslash}m{20em}}

\newcolumntype{R}{>{\centering\arraybackslash}m{3.2em}}
\newcolumntype{T}{>{\centering\arraybackslash}m{10.6em}}

\usepackage{amsmath}
\usepackage{amssymb}
\usepackage{booktabs}
\usepackage{multirow}
\usepackage{mathtools}
\usepackage{url}
\usepackage{pifont}
\usepackage{wrapfig}

\definecolor{brown}{RGB}{128, 76, 37}

\graphicspath{{/}}

\makeatletter
\let\@currsize\normalsize
\makeatother

\newcommand{\B}{\fontseries{b}\selectfont}

\usepackage{graphicx}

\usepackage{tikz}
\usepackage{comment}
\usepackage{amsmath,amssymb} 
\usepackage{color}
\usepackage{amsmath,graphicx}
\usepackage{booktabs}
\usepackage{nicematrix}
\usepackage{tabularray}
\usepackage{multirow}
\usepackage{tabto}
\usepackage{float}

\newcommand {\R}{\mathbb{R}}
\usepackage{array}
\usepackage{xcolor}
\usepackage{color, colortbl}
	
\definecolor{Gray}{gray}{0.9}
\definecolor{LightCyan}{rgb}{0.88,0.95,1}

\makeatletter
\newcommand*{\@rowstyle}{}
\newcommand*{\rowstyle}[1]{
  \gdef\@rowstyle{#1}%
  \@rowstyle\ignorespaces%
}
\newcolumntype{=}{
  >{\gdef\@rowstyle{}}%
}
\newcolumntype{+}{
  >{\@rowstyle}%
}
\makeatother

\usepackage{amssymb}
\usepackage{pifont}

\usepackage{url}
\usepackage{graphicx}
\usepackage{amssymb}
\usepackage{amsfonts}
\usepackage{amsmath}
\usepackage{amsthm}
\usepackage{booktabs}
\usepackage{xcolor}
\usepackage{tabularx, booktabs} 
\usepackage{graphicx}
\usepackage{booktabs}
\usepackage{multirow}
\usepackage{colortbl}
\usepackage{xcolor}
\usepackage{amssymb}
\usepackage{pifont}

\usepackage{placeins}
\usepackage{pifont}

\definecolor{dgreen}{rgb}{0.0, 0.5, 0.0}
\newcommand{\cmark}{\color{dgreen} \ding{51}}%
\newcommand{\xmark}{\color{black} \ding{55}}%
\usepackage{lipsum}

\usepackage{changepage}

\newcommand{\etal}{\emph{et al}.}

\newcommand{\mcc}[1]{\multicolumn{#1}{c}}
\newcommand{\mcp}[1]{\multicolumn{#1}{c}} 

\newcommand{\texta}[1]{$\texttt{#1}$}
\definecolor{Gray}{gray}{0.90}
\newcolumntype{a}{>{\columncolor{Gray}}r}
\newcolumntype{d}{>{\columncolor{Gray}}c}

\usepackage{enumitem}
\newcommand{\subscript}[2]{$#1 _ #2$}
\usepackage{afterpage}

\newcommand\blankpage{%
    \null
    \thispagestyle{empty}%
    \addtocounter{page}{-1}%
    \newpage}

\makeatletter
\def\@makechapterhead#1{%
  \vspace*{50\p@}%
  {\parindent \z@ \raggedright
    \ifnum \c@secnumdepth >\m@ne
      \Huge \bfseries \@chapapp{} \thechapter
      \par
      \vskip 20\p@
    \fi
    \fontsize{22}{26.4}\selectfont \bfseries
    #1\par
    \nobreak
    \vskip 40\p@
  }}
\makeatother

\begin{document}



\dsp
\raggedbottom

\title{{Machine Intelligence that Understands Visual and Linguistic Information and Interacts with Humans and Environments}}
\japtitle{\phantom{x}}
\author{Nguyen Van Quang}
\degreemonth{July}
\degreeday{7}
\degreeyear{2022}
\degree{Doctor of Philosophy} 
\field{Information Sciences}
\department{System Information Sciences} 
\advisor{OKATANI Takayuki}
\maketitle

\afterpage{\blankpage}
\clearpage

{\thispagestyle{empty}}
\pagenumbering{Roman}							
\setcounter{page}{1}


\cleardoublepage
\begin{abstract}
\renewcommand{\baselinestretch}{1.2}\normalsize
Over the past years, Artificial Intelligence has witnessed significant progress in computer vision and natural language processing thanks to deep learning advancements. Inspired by remarkable success in these two independent fields, there has been a growing interest in the problems at the intersection of visual and linguistic understanding. It is believed that advances in solving those mentioned above and related problems would open the door to many real-world applications, bringing fundamental change to society. Take virtual assistants that aid the visually impaired, automatic surveillance systems for querying over visual databases, and in-home robots that perform household tasks as examples. Thus, the integration of vision and language is a viable approach to achieving one of AI's visionary goals: building machines that can understand both the visual and linguistic worlds, communicate with humans in natural language, and further interact with environments.

In this dissertation, we aim to build and improve agents endowed with such intelligence as a continuation of collective efforts by research communities. Specifically, we focus our attention on three representative vision language tasks, namely \textit{image captioning}, \textit{visual dialog}, and \textit{interactive instruction following tasks}. 

In the first part of the work, we revisit how to extract and utilize visual representations, aiming to build a better and faster model for image captioning. In image captioning, understanding visual information is crucial to correctly describing its content in words. Therefore, extracting good visual representations from the input image is necessary. Current state-of-the-art methods employ region-based features extracted by high-performance object detectors, e.g., Faster R-CNN. However, they have several issues, for example, the lack of contextual information, the risk of incorrect detection, and the high computational cost. The first two could be addressed by additionally using grid-based features. However, how to extract and integrate these two types of features was uncharted. We propose a transformer-only neural architecture, dubbed GRIT (Grid and Region-based Image captioning Transformer), that can effectively extract and integrate the two visual features to generate better captions for input images. Specifically, GRIT replaces the CNN-based detector employed in previous methods with a DETR-based one, making it computationally faster and end-to-end trainable. We find that the proposed method brings about significant performance improvement, outperforming previous methods in inference accuracy and speed.

In the second part of this work, we tackle the visual dialog task, which requires agents to maintain a meaningful conversation with humans about the content of input images by answering questions. Unlike image captioning, the agent must handle multiple inputs, i.e., an image, a question, a dialog history, or even its individual dialog components. Thus, the key to success lies in how to model all the interactions between these inputs effectively and efficiently. We introduce a neural architecture, LTMI (dubbed Light-weight Transformer for Many Inputs), that can efficiently deal with all the interactions between multiple inputs in the visual dialog. It has a block structure similar to the Transformer and employs the same design for attention computation. With a similar setting on visual dialog, a layer built upon the proposed attention block has less than one-tenth of the parameters compared with its counterpart, a natural Transformer extension. It has only a small number of parameters yet has sufficient representational power for the purpose. The experimental results on the VisDial dataset validate the effectiveness of our proposed method.

In the last part of this work, we study interactive instruction-following tasks. An embodied AI agent is required to perform a sequence of actions to accomplish a complicated task in the interactive environment by following natural language directives. Recent studies have tackled the problem using ALFRED, a well-designed dataset for the task, but have obtained only very low accuracy. To this end, we propose a novel method based on a combination of several new ideas, which surpasses the existing methods by a large margin. One is a two-stage interpretation of the given instructions. The method first chooses and decodes an instruction without visual information, yielding a tentative sequence of object and action predictions. It then integrates this prediction with the visual information to generate the final prediction of an action and an object. It can localize the object of interest accurately from the input image.
Furthermore, the proposed method utilizes multiple egocentric views of the environment and extracts crucial information by applying hierarchical attention conditioned on the selected instruction. It leads to better accuracy in predicting navigation actions. Our proposed method attains an unseen success rate of 8.37\%. 

\end{abstract}

\cleardoublepage 
\pagenumbering{roman}
\setcounter{page}{1}
\addcontentsline{toc}{section}{Table of Contents}
\tableofcontents

\cleardoublepage
\listoffigures
\listoftables


\cleardoublepage
\pagenumbering{arabic}
\setcounter{page}{1}
\chapter{Introduction} \label{chapter:ch1}


\section{Artificial Intelligence Overview}
\subsection{Artificial Intelligence Progress}
Artificial Intelligence (AI), or Machine Intelligence, has advanced rapidly in recent years.
This achievement is arguably attributed to tremendous progress in AI sub-fields, including computer vision (CV) and natural language processing (NLP), thanks to deep learning advancements.
Computer vision has seen many remarkable achievements in many tasks, such as image recognition \cite{krizhevsky2012imagenet,simonyan2014very,he2016deep,dosovitskiy2020image}, object detection \cite{ren2015faster}, semantic segmentation \cite{he2017mask}, using large labeled datasets \cite{deng2009imagenet}, or employing self-supervision \cite{jing:2019} on large-scale unlabeled data.
Similarly, NLP has experienced unprecedented studies using deep learning, achieving remarkable performance on many downstream tasks powered by neural networks pre-trained on large-scale text corpora \cite{devlin2018bert,radford2018improving,brown2020language}. 
Consequently, there is also a growing interest in the problems at the intersection between these two independent fields that require visual and linguistic understanding.

\subsection{Integration of Vision and Language} \label{sec:vl}
Integrating the two fields of computer vision and natural language processing is a viable approach toward one long-standing goal of AI: building machines that can perceive the worlds of vision and language, communicate with humans in natural language, and further interact with the physical environments around us. 
Inspired by the tremendous success in CV and NLP, an increasing amount of attention has been paid to the problems lying at the intersection between vision and language domains, taking further steps towards this visionary goal.
Many pilot tasks in this intersecting region have been designed and introduced to the research community, together with datasets. 
It remains challenging as the tasks require future machine intelligence not only to (1) acquire a comprehensive understanding of visual and/or linguistic information but also to (2) generate descriptions or stories about the visual content \cite{karpathy2014deep,vinyals2015show}, (3) specify salient regions and objects and their relationships in the image to reason about, or answer arbitrary questions about its content \cite{antol2015vqa,das2017visual}, (4) navigate through and interact with physical environments by leveraging natural language instructions \cite{anderson2018vision,fried2018speaker,zhu2020vision,shridhar2020alfred}, etc.
Methods that can deal with and translate between various modalities (for example, visual and linguistic inputs) are classified as a sub-category of multi-modal models, which were originally described \cite{mogadala:2015}.

\paragraph{Practical Applications} 
Advances in tackling the aforementioned and other related challenges are expected to open the door to a broad range of practical applications, bringing about a radical change in society.
For example, the visually impaired can be helped better by a future generation of virtual assistants. They can obtain helpful information about a scene from generated descriptions and by being able to ask questions about it.
It can be used in automatic surveillance for querying from enormous image and video databases using natural language and in personal navigation systems that can process or even generate navigation instructions in natural language \cite{das2017visual,anderson2018vision}. 
It also includes in-home robots that perform household tasks \cite{shridhar2020alfred}. 
Lastly, these problems play a critical role in evaluating the performance of AI systems that contribute to the progress of designing better machines with more collectively comprehensive intelligence than independent ones in CV and NLP.


\section{Our Research Problems}
In this dissertation, we aim to build and improve agents endowed with such the intelligence as a continuation of collective efforts by research communities. In particular, we study three representative vision language tasks, namely \textbf{image captioning} \cite{karpathy2014deep}, \textbf{visual dialog} \cite{das2017visual}, and \textbf{interactive instruction following} \cite{shridhar2020alfred}.

\paragraph{Challenges} When tackling tasks that require visual and linguistic understanding with the translation between the two modalities, humans can do it with ease. 
For example, a human can point out and convey an enormous quantity of information about a visual scene with just a cursory look at it. On the other hand, computers find it difficult when they deal with unstructured data, e.g., images and text.
Next, we will describe the three tasks, the challenges we confront when designing agents, and the shortcomings of previous studies, and summarize our proposed approaches.

\subsection{Image Captioning} 
\begin{figure}[ht]
\centering
\includegraphics[width=1.0\linewidth]{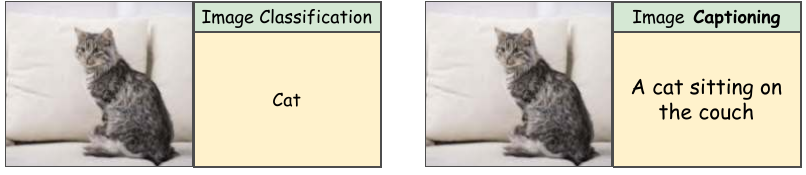}
\caption{From {Image Classification} to {Image Captioning}. Left) Predict an category for an image; Right) Generate a description in a sentence for the image.}
\label{fig:rec_cap_chap1}
\end{figure}

The task of image captioning requires generating a sentence describing the content of a scene for a given image. We usually represent an image by an array of pixels with intensity values in the color channel(s) (e.g., RGB images have three color channels while black-white images have a single channel); a typical image might have millions of pixels.
To recognize the objects in the image (e.g., a cat as in Fig.\ref{fig:rec_cap_chap1}), the agent must convert these raw intensity values into high-level concepts of objects. On the other hand, many objects share similar low-level patterns (e.g., from cats and dogs to carpets and coats, they all have fur). It makes the recognition of objects in the image by manually programming unattainable.

Generating a sentence in natural language is also a challenging task. Unlike the object recognition task that assigns one or a few labels to the image, computers must output a sequence of words in a large vocabulary to reflect their understanding of the visual content (see Fig.~\ref{fig:rec_cap_chap1}). Therefore, image captioning requires a complex pattern recognition process of identifying salient objects and regions and annotating them with a sequence of integers to represent words in the caption.

It is believed in the community that extracting visual representations from an input image plays a crucial role in generating better captions. Identifying existing objects and their relationships in the image is especially beneficial for precisely describing its visual content. 
The state-of-the-art methods utilize the region-based features obtained from CNN-based detectors, such as Faster R-CNN \cite{ren2015faster}, since they encode detected objects directly. 
However, the region-based features have several issues, such as a lack of contextual information, risk of false detection, and expensive computation. The grid-based features are the high-level feature maps extracted from the entire image. They thus represent contextual information while being free from the risk of incorrect object detection.

In this dissertation, we revisit how to extract these dual visual features from input images and how to combine such region and grid features in an integrated manner, aiming to build a better and faster model for image captioning. The underlying idea is that appropriate integration of the two visual features will provide a better representation of the input image since they are complementary, as explained above. 
In particular, in Chapter \ref{chapter:ch3}, we propose a Transformer-only neural architecture, dubbed GRIT (Grid and Region-based Image captioning Transformer), that effectively utilizes the two visual features to generate better captions. GRIT replaces the CNN-based detector employed in previous methods with a DETR-based one, making it computationally faster. Moreover, the monolithic design consisting only of Transformers makes it end-to-end trainable. We find that the proposed method obtains considerable performance gain, surpassing previous methods in both inference accuracy and speed.

\subsection{Visual Dialog}
\begin{figure}[ht]
\centering
\includegraphics[width=1.0\linewidth]{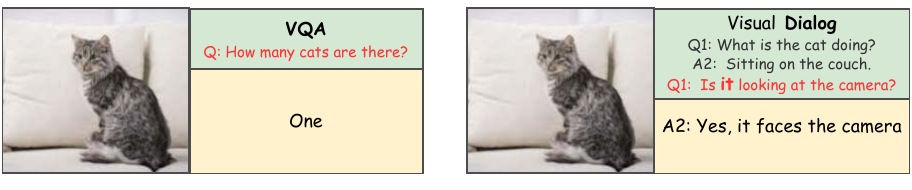}
\caption{From Visual Question Answering to Visual Dialog. Left) Answer a single question; Right) Answer multiple questions in dialog.}
\label{fig:vqa_dialog_chap1}
\end{figure}
Unlike the image captioning task, where agents perform only one-way communication by returning only the image description to humans, visual dialog agents communicate with humans in a two-way manner (see Fig.~\ref{fig:model_overview_chap1}).
Specifically, the task of visual dialog demands an agent to maintain a meaningful conversation with humans in natural language by answering questions about the visual content. 
Visual dialog has been developed aiming at a higher level of vision-language interactions \cite{das2017visual}, as compared with VQA (visual question answering) \cite{antol2015vqa} and VCR (visual commonsense reasoning); see Figure \ref{fig:vqa_dialog_chap1}. 
It extends VQA to multiple rounds; given an image and a history of question-answer pairs about the image, an agent is asked to answer a new question.
This task requires multiple sub-problems ranging from visual understanding to natural language tasks such as language generation, co-reference resolution, etc.
For example, to answer the question {\it `What color are they?'}, the agent needs to understand the context from a dialog history to know what {\it `they'} refers to and look at the relevant image region to find out the color.

The key to success in visual dialog lies in how to model the interactions between multiple inputs, i.e., the input image, the dialog history, and a question. To this end, we propose in Chapter \ref{chapter:ch4} a neural architecture named Light-weight Transformer for Many Inputs (LTMI) that can efficiently deal with all the interactions between multiple such inputs in visual dialog. It has a block structure similar to the Transformer and employs the same design for attention computation. In contrast, it has only a small number of parameters yet has sufficient representational power for the purpose. With a similar setting on visual dialog, a layer built upon the proposed attention block has less than one-tenth of the parameters compared with its counterpart, a natural Transformer extension. The experimental results on the VisDial dataset validate the effectiveness of the proposed method.

\subsection{Interactive Instruction Following}
\begin{figure}[ht]
\centering
\includegraphics[width=0.6\linewidth]{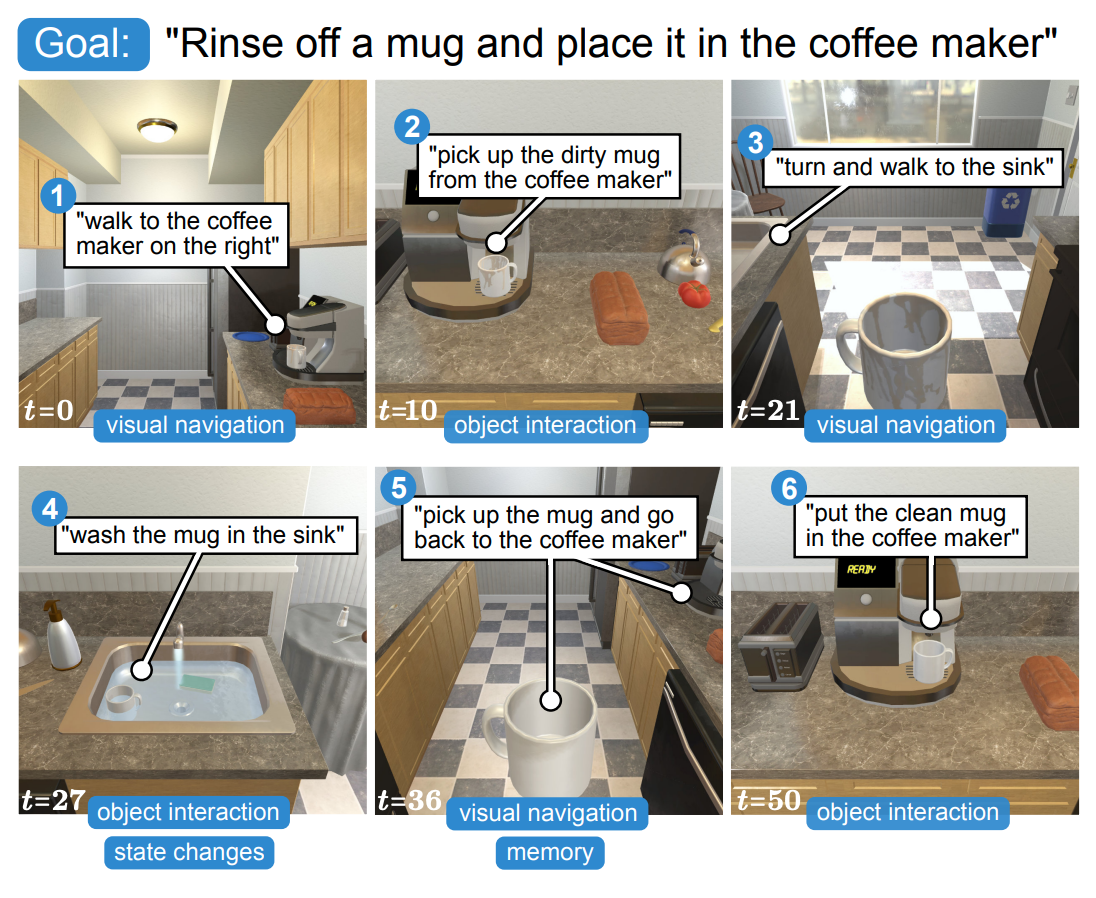}
\caption{An example of the ALFRED task with highlighted frames corresponding to a portion of accompanying instructions. Source: \cite{shridhar2020alfred}.}
\label{fig:alfred_example_chap1}
\end{figure}
The image captioning and visual dialog problems are beneficial for stimulating and evaluating progress on the problems involving both vision and language domains. They do, however, share a flaw: they are both passive. In these tasks, the agent is not allowed to move or manipulate the camera or interact with the environment. Therefore, the visual inputs are static, i.e., fixed images.
It underestimates one of the most important aspects of several practical applications stated in Section \ref{sec:vl}, as each of these examples (e.g., in-home robots) requires an embodied agent. 

In the last part of our work, we focus on the interactive instruction following tasks.
We examine a more complex problem by addressing a recently constructed benchmark known as ALFRED \cite{shridhar2020alfred}. 
An agent must do a household task in an interactive environment by following verbal directions. In comparison with a close problem called vision-language navigation (VLN) \cite{anderson2018vision}, ALFRED is more difficult because the agent must (1) reason over a larger number of instructions and (2) predict actions from a wider action space in order to complete a task across longer time horizons. The agent must also (3) predict the pixel-wise masks to localize the objects of interest. Previous studies (e.g., \cite{shridhar2020alfred}) employ a Seq2Seq model, which performs well on the VLN tasks \cite{ma2019selfmonitoring}. However, it does not work well on ALFRED.
Consequently, previous approaches have poor performance and a large gap compared with humans. 

In Chapter \ref{chapter:ch5}, we present a new method, which surpasses the prior methods by a significant margin. We propose an embodied agent based on several new ideas. One is a two-stage interpretation of the given instructions. The method first selects and decodes an instruction without visual information, yielding a tentative sequence prediction of objects and actions. It then combines this prediction with the visual information, obtaining the final prediction of an action and an object with better accuracy. Furthermore, our method utilizes multiple ego-centric views and extracts crucial information using hierarchical attention conditioned on the selected instruction.

\subsection{Research Questions}

As a result, we will address three main research questions (\textbf{RQ}s) while studying the three research problems:
\begin{enumerate}[label=\subscript{\textbf{RQ}}{{\arabic*}}:]
    \item In vision-language tasks, understanding visual information is of importance. It is reasonably valid for image captioning, in which agents must grasp visual information before describing an input image in words. To this end, we need a form of representation for visual inputs so that neural networks can learn helpful information to caption an image effectively. We raise a question: \textbf{how do we extract good visual representations from input images?}
    
    \item In \subscript{\textbf{RQ}}{1}, neural networks process only the visual inputs. Meanwhile, visual dialog tasks require neural networks to gain understanding from both visual and linguistic inputs in order to answer questions. We deal with multiple inputs in these tasks, e.g., an image, a dialog history, and a question. It is thus essential to address a question: \textbf{how do we model the interactions between multiple inputs efficiently?}
    
    \item Agents process information passively in the above problems, i.e., agents only receive and process the inputs given by humans. We ask a question: \textbf{how do we build embodied agents that interact with physical environments and collect visual inputs actively itself?} We attempt to answer this question by tackling ALFRED, a well-defined embodied problem.
\end{enumerate}

\section{Dissertation Outline and Contributions} \label{sec:outline_of_dissertation}
\begin{figure}[t]
\centering
\includegraphics[width=0.6\linewidth]{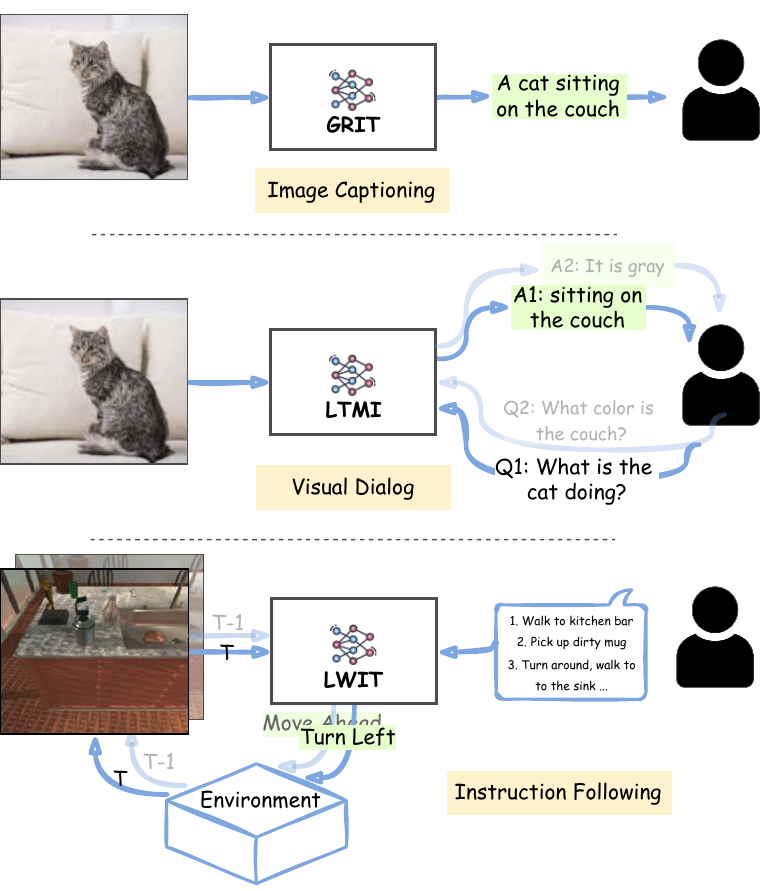}
\caption{Illustrations of the research problems with our corresponding proposed agents in this dissertation. 
Each proposed agent/model is encapsulated with a black bounding box, outputting the predictions highlighted in green.
\textbf{Top}) We propose GRIT, a Grid- and Region-based Transformer for Image captioning.
\textbf{Middle}) We propose LTMI, a lightweight transformer that handles multiple inputs in Visual Dialog.
\textbf{Bottom}) We propose LWIT, a neural agent that performs household tasks following humans' instructions.}
\label{fig:model_overview_chap1}
\end{figure}

In this dissertation, we develop new models for machine intelligence that understand visual and linguistic information and interact with humans and environments, focusing on the three representative vision-language problems; See the overview in Figure \ref{fig:model_overview_chap1}. 
We present the dissertation's outline and summarize each chapter's contributions (if any) as follows.

\paragraph{Chapter \ref{chapter:ch2}--Background} This chapter provides some fundamental background that is applied throughout the rest of the dissertation.

\paragraph{Chapter \ref{chapter:ch3}--GRIT: Integrating Dual Visual Features for Image Captioning
} In this chapter, we revisit the representation of visual features, aiming to build a better and faster image captioning model. We then propose GRIT, an \textbf{G}rid- and \textbf{R}egion-based \textbf{I}mage-captioning \textbf{T}ransformer that effectively utilizes the two visual features to generate better captions. GRIT replaces the CNN-based detector with a DETR-based one, making it computationally faster and end-to-end trainable. The experimental results show that GRIT outperforms the previous methods by a large margin in inference speed and accuracy.

\paragraph{Chapter \ref{chapter:ch4}--LTMI: Light-weight Transformer for Many Inputs in Visual Dialog
} 
In this chapter, we tackle the visual dialog task, which requires the agent to handle multiple inputs, i.e., an image, a question, a dialog history, or its components. We introduce a neural architecture named \textbf{L}ight-weight \textbf{T}ransformer for \textbf{M}any \textbf{I}nputs (LTMI) that can efficiently deal with all the interactions between multiple such inputs in visual dialog. LTMI possesses sufficient representational power with much fewer parameters and a similar block structure to Transformer. The experimental results on the VisDial dataset validate the effectiveness of our proposed method.

\paragraph{Chapter \ref{chapter:ch5}--LWIT: Improving Performance on Instruction Following Tasks
} In this chapter, we tackle ALFRED, the interactive instruction following tasks. Previous methods only show limited performance on the task; there is a huge gap with human performance. We propose LWIT (\textbf{L}ook \textbf{W}ide \textbf{I}nterpret \textbf{T}wice), a new method which outperforms the previous methods by a large margin. It is based on a combination of several new ideas, such as a two-stage interpretation of the provided instructions, using multiple egocentric views of the environment and extracting essential information by applying hierarchical attention conditioned on the current instruction, etc.

\paragraph{Chapter \ref{chapter:ch6}--Conclusion and Future Directions}
We conclude the dissertation by summarizing the main contributions and discussing future research toward improving this machine intelligence.

\cleardoublepage
\chapter{Preliminaries}
\label{chapter:ch2}

This chapter summarizes the introductory knowledge on recent deep learning that we will apply throughout the dissertation. To acquire a broader and deeper understanding, we highly recommend the Deep Learning textbook \cite{goodfellow2016deep} which gives a more comprehensive introduction.

\section{Preliminary Background}
The proposed methods in this dissertation are mainly built upon modern deep learning techniques.
Recent deep learning advances provide a powerful framework for supervised learning. Throughout this dissertation, we tackle several supervised learning problems that require learning an input-to-output mapping from training examples. We will present a generic supervised learning problem, optimization techniques to learn the mapping, and a typical workflow for building deep learning algorithms.

\subsection{Supervised Learning}
A supervised learning problem requires learning a mapping $f : X \rightarrow Y$, where $X$ and $Y$ represent the space for input and output, respectively.
Supervised learning algorithms are those that are able to learn from a labeled dataset. Each sample in the dataset is represented by a vector $x$ and associated with a label $y$, forming a datapoint $(x, y)$. 
A classical example of supervised learning problems is Iris classification, in which each datapoint is associated with a plant. We represent a plant by a single vector $x$ which is composed of four measurements: the petal length, the sepal length, the petal width, and the sepal width. The species form the output space $Y$. 
Take image captioning as a more complicated supervised learning example in this dissertation. The input $x$ is an image, while the output $y$ is a sentence describing the content of the image.
It can often be challenging to manually define the mapping $f$ that can be explicitly programmed using standard techniques. As an instance, take writing a hand-coded program that can describe an image in natural language. 
On the other hand, supervised learning algorithms provide an alternate approach by learning from many training examples $(x, y) \in X \times Y$ that can be collected easily in practice.

Formally, a supervised learning model can be defined as $y = f(x; \theta)$ where $\theta$ are the parameters learned from the training examples that result in the best approximation of $f^\star$. Given $n$ training examples $\{(x_1, y_1), (x_2, y_2), \dots, (x_n, y_n)\}$, the best approximation $f^\star$ is defined as 
\begin{equation} \label{eq:ch2-opt}
f^\star = \underset{\theta}{\mathrm{argmin}}\sum_i^n \mathcal{L}(f(x_i; \theta), y_i),
\end{equation}
where $\mathcal{L}$ is the per-sample loss (e.g., the mean square error (MSE), cross entropy). The loss function $J(\theta)$ measuring the model performance over the entire training samples is defined by
\begin{equation}
    J(\theta) = \sum_i^n \mathcal{L}(f(x_i; \theta), y_i).
\end{equation}

\subsection{Optimization}
As of now, it becomes an optimization problem by minimizing the loss function, e.g., $J(\theta)$. 
Maximization can be performed via a minimization algorithm by minimizing its negative function, e.g., $-J(\theta)$. In deep learning, the loss function can be the cost function, objective function, or training error. 
\paragraph{Gradient Descent} Suppose a function $y = f(x)$, where $x$ and $y$ are real numbers. 
The slope of $f(x)$ at the point $x$ is given by its derivative, denoted as $f'(x)$ or $\frac{dy}{dx}$. It reflects how $f(x)$ changes in relation to the change in $x$.
Using this knowledge, we can minimize $f(x)$ by moving $x$ in small steps in the opposite direction of the derivative's sign. This is known as the \textbf{gradient descent} algorithm. It is natural to apply this algorithm to the optimization problem in \ref{eq:ch2-opt} in order to find a function $f^\star$ by adjusting $\theta$ values. It is required computing the following gradient
\begin{equation}\label{eq:ch2-grad}
\nabla_\theta = \sum_i^n \nabla_\theta \mathcal{L}(f(x_i; \theta), y_i).
\end{equation}

\paragraph{Stochastic Gradient Descent} In many practical problems, a sufficiently large dataset is required to seek a better model with sufficient generalizability. As a result, applying gradient descent directly to Eq.\ref{eq:ch2-opt} is computationally expensive, slowing down the optimization. In practice, to train a deep learning model on a large dataset, we usually use its extension, Stochastic Gradient Descent (SGD). 
The loss function is computed over a small number of training examples, which are sampled randomly during the training phase. 
Specifically, at one optimization step, we sample a \textbf{mini-batch} of $m$ samples $\{(x_i, y_i)\}_{i=1}^m$ from the training set. Thus, rather than computing the gradient in \ref{eq:ch2-grad} across all training examples, we only need to compute the gradient shown below.
\begin{equation}
    \nabla_\theta = \sum_i^m \nabla_\theta \mathcal{L}(f(x_i; \theta), y_i),
\end{equation}
where $m$ is a relatively smaller number that is independent of the number of training examples $n$. As a result, the model's parameters $\theta$ are updated as follows:
\begin{equation}
    \theta \longleftarrow \theta - \epsilon  \nabla_\theta,
\end{equation}
where $epsilon$ represents the learning rate. It is critical to set the learning rate to a suitable value that allows the optimization to properly converge. Inappropriate learning rates will slow convergence or cause the optimization to diverge.

\paragraph{Advanced Optimization Algorithms} Several advanced gradient-based optimization algorithms have been developed that result in faster convergence. The \textbf{momentum} methods \cite{polyak1964some, sutskever2013importance} can be used to accelerate training with SGD. It accumulates an exponentially decaying moving average of previous gradients and keeps moving in their direction. Several approaches, including Adam \cite{kingma2014adam}, AdaGrad \cite{duchi2011adaptive}, and RMSProp \cite{Hinton06}, use adaptive learning rates to speed up optimization. It should be noted that this dissertation mostly employs Adam as the primary optimizer choice.

\subsection{Deep Learning Workflow}
We already know how to solve a generic supervised learning problem using optimization. To solve this optimization problem, we can use the SGD algorithm to find the best $f^\star$ possible. In general, any deep learning algorithm can be thought of as an instance of a simple recipe that combines a dataset, a cost function, a model, and an optimization function.  The following is a typical workflow for applying any deep learning model:

\paragraph{Data preparation} The first step in most deep learning workflows is to obtain and prepare the dataset. The dataset is divided into training, development, and testing splits. In this dissertation, we focus on well-defined tasks with datasets that are primarily divided into training and development/validation splits. Each dataset also includes a testing split without any annotations; the result of this split can only be obtained by submitting it to the dataset's online server. Following that, it is critical to inspect a dataset to gain an initial understanding of the task at hand, such as its distribution, how samples are collected and labeled, existing issues, and so on.

\paragraph{Data preprocessing} Preprocessing data usually speeds up the training procedure, allowing the optimization to converge more quickly. Images are preprocessed, for example, by normalizing pixel values in each dimension with a given mean and standard deviation. This step is necessary in two situations: (1) when we extract features or finetune a pretrained backbone on a larger dataset, such as ImageNet, and (2) when we preprocess the samples during deployment using the same estimate statistics as training samples.

\paragraph{Architecture design} All the deep learning algorithms require specifying the function space or the architecture family the optimization will seek. Based on the understanding of the dataset, one can make a heuristic on which kind of architecture can be employed. For example, CNNs are commonly used for grid-like data (e.g., images), whereas RNNs can handle sequential data (e.g., text), and so on. A heuristic like this would help us quickly build viable architectures and establish initial baselines. Section \ref{sec:ch2-dnn} will present several common neural architectures upon which we build our proposed methods. Designing architecture is not limited by prior experience; thus, we can experiment with various architectures through trials and errors.

\paragraph{Training and Validation} 
During the training steps, the designed model learns patterns from the training samples by utilizing optimization algorithms. While there are many popular optimization algorithms, researchers recommend using Adam with the default learning rate and first and second moment coefficients at the very first training. It is also common to use learning rate schedulers, which adjust the learning rate throughout training. We can also validate the model's prediction ability on the validation split during the training steps.

\paragraph{Hyper-parameter optimization} Deep learning algorithms involve numerous hyper-parameter decisions. Because it is difficult to find the optimal combination for all hyper-parameters, it is common to begin with the default settings in previous studies. Grid search and random search are the two standard techniques for performing hyper-parameter search efficiently. By evaluating the model on the validation set, we perform a hyper-parameter search. The final best model with the best validation performance is then chosen to perform on the test split in order to measure its performance.

\section{Deep Neural Networks} \label{sec:ch2-dnn}
Neural networks are well-known in the deep learning regime for approximating $f$. In the previous section, we defined the arbitrary function $f$ that uses an optimization algorithm to learn the mapping $X \rightarrow Y$ for supervised learning problems. This section will explore several deep neural networks commonly used for visual and linguistic understanding.

\subsection{FeedForward Neural Networks}
The development of Feedforward Neural Networks dates back to 1958 when Frank Rosenblatt introduced the first Artificial Neural Network \cite{rosenblatt1958perceptron}, named \textbf{perceptron}. 
The concept of the perceptron was inspired by the operation of a biological brain.
Perceptrons were not learning efficiently until 1986 when David et al. \cite{rumelhart1985learning} proposed the back-propagation algorithm, which can compute the gradient efficiently.

Feedforward neural networks are constructed by combining multiple functions. The most common method of building feedforward neural networks is by stacking multiple functions in a chain structure. For example, a two-layer network can be formed as $f(x) = W_2\sigma(W_1 x + b_1) + b_2$, where $W_1$ and $W_2$ are learnable matrices, $b_1$ and $b_2$ are learnable biases, and $\sigma$ is a non-linear function (e.g., tanh, sigmoid). 
A function transforming input $x$ into $Wx + b$ with learnable parameters $W$ and $b$ forms a \textbf{Fully-Connected} (FC) layer. 
It is noted that an \textbf{activation function}, denoted by $\sigma$, is added to introduce non-linearity and enhance the representational power of the neural network (its learning capacity). The number of layers in these structures indicates the depth of neural networks, giving birth to the terminology \textbf{Deep Neural Networks} (DNN).

Feedforward neural networks are the foundation of many architecture designs. The specialized feedforward neural network types include convolutional neural networks, recurrent neural networks, and transformers. In the sections that follow, we will introduce briefly the neural architecture of these networks.

\subsection{Convolutional Neural Networks}
Simple neural networks composed of multiple FC layers make no assumptions about the input data.
It becomes inefficient when processing high-dimensional input data, such as images with several hundred pixels per dimension. Many statistical properties of natural images are invariant to translation.

\begin{figure}[ht]
    \centering
    \includegraphics[width=0.9\linewidth]{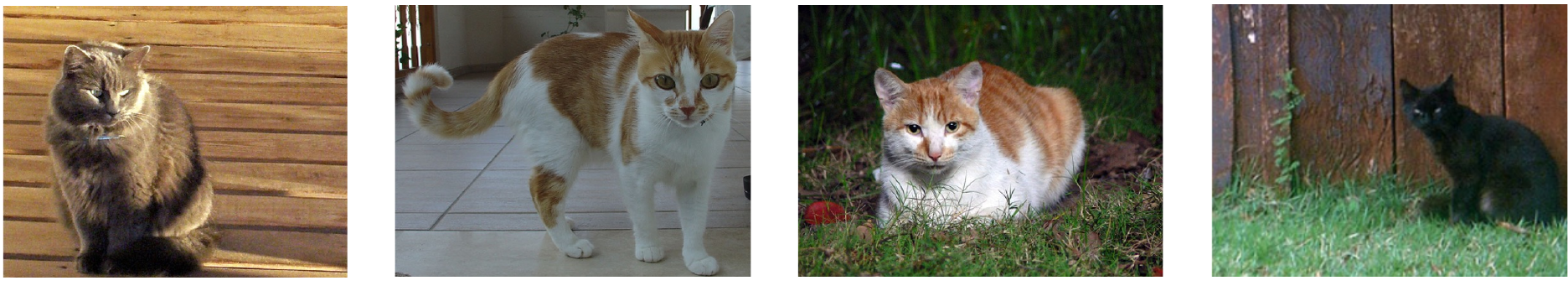}
    \caption{Examples of cat images. They are of different cat breed, position, size, color intensity, etc. Source: Kaggle dataset.} 
    \label{fig:cat_images}
\end{figure}
For instance, we classify the cat images as ``cat'' regardless of their breed, scale, and location in the image; see Figure \ref{fig:cat_images}.
Convolutional neural networks (CNN) \cite{lecun1998mnist} are proposed to deal with these grid-like data (e.g., images, videos, etc).
CNNs take this invariance into account by performing \textbf{convolution} across different locations in the input representations with the kernels using shared parameters. 

CNNs possess a number of characteristics that make them ideal for grid-like data \cite{goodfellow2016deep}. First, because convolutions are translation invariant, objects and other characteristics can be recognized regardless of where they are in the image. Second, CNNs use the same kernels at every input location. Therefore, convolutional architecture is computationally more efficient than architectures with fully connected layers. Lastly, interactions in a convolutional layer are sparse, as only a few neighboring tensors are convolved with the kernels.

\begin{figure}[ht]
\centering
\includegraphics[width=1.0\linewidth]{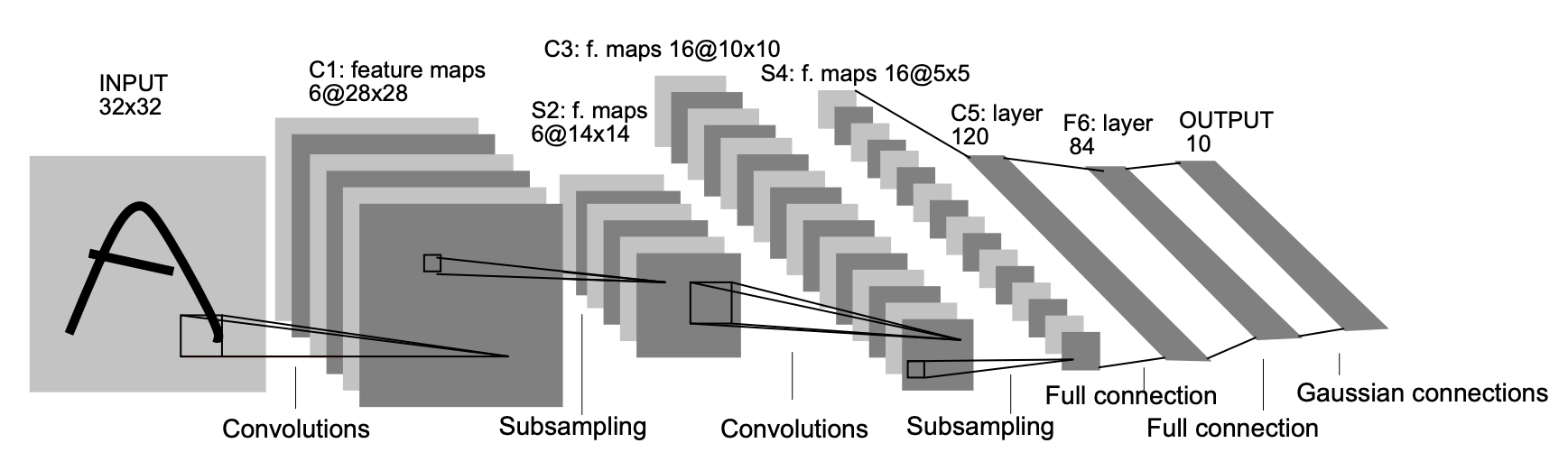}
\caption{The architecture of LeNet-5, the first convolutional neural network introduced by Lecun et al. \cite{lecun1998mnist} for character recognition. The architecture of LeNet-5 consists of two convolutional layers, two subsampling (pooling) layers, and two fully connected layers. Soure: \cite{lecun1998mnist}.}
\label{fig:chap-vgg}
\end{figure}

A typical convolutional neural network is formed by stacking \textbf{convolutional layers} and \textbf{pooling layers}. LetNet-5 \cite{lecun1998mnist} has two convolutional layers, two subsampling (pooling) layers, two fully connected layers, and two fully connected layers. All the layers are stacked in a chain structure as shown in Figure \ref{fig:chap-vgg}.

\paragraph{Convolutional layers} The input to the convolutional layer is a tensor. It then produces an output tensor by convolving the input with a set of filters. We can convolve the filter by sliding it across all spatial positions of the input tensor and computing a dot product between it and a small region of the input tensor at each spatial position. This will result in an activation map.
As a specific example, consider a $256 \times 256 \times 3$ input that is processed with a layer having 32 $5 \times 5 \times 3$ filters with padding of 2 and stride of 1. In this instance, the output would be $256 \times 256 \times 32$, representing the firing of all filters at all spatial locations.

Following are characteristics of a typical convolutional layer.
\begin{itemize}
    \item It takes a tensor of size $W_1\times H_1 \times D_1$ (e.g., 3D tensors).
    \item It has four hyper-parameters: $K$ is the number of filters, $F$ is their spatial extent, $S$ is the stride, and $P$ is zero padding on the input's borders.
    \item It produces an output having a size of  $W_2\times H_2 \times D_2$, in which $W_2 = (W_1 - F + 2P)/S + 1; H2 = (H_1 - F + 2P)/S + 1$, and $D_2 = K$.
    \item Each filter is $(F * F * D_1)$ parameters, for a total of $(F* F* D_1 * K)$ weights and $K$ biases. It is noted that the receptive field of the filters is small with an area of $F\times F$, but that it always traverses the entire depth of the input tensor ($D_1$).
    \item The size of the $d$-th of the output tensor is $(W_2\times H_2)$. It is the result of convolving the $d$-th filter over the input tensor with a stride of $S$ and then offsetting by the $d$-th bias.
\end{itemize}

\paragraph{Pooling layers}
In addition to convolutional layers, pooling layers are typically used to downsample feature maps. Pooling layers, unlike convolutional layers, transform input tensors without employing any learnable parameters. In particular, the pooling layers perform independently on each channel (activation map) and spatially downsample them. 
\emph{Max pooling layer} and \emph{average pooling layer} are frequently used as the building blocks in designing CNN architectures.
Consider a max pooling layer with $2\times2$ filters and a stride of 2, where each filter performs the max operation over four numbers. 
Therefore, an input tensor is precisely downscaled by a factor of two in both width and height, and the representation size is reduced by a factor of four at the cost of some local spatial information loss.

\subsection{Recurrent Neural Networks}
\paragraph{Recurrent Neural Networks} (RNN) \cite{rumelhart1988learning} is a type of neural network designed to process sequential data.
Unlike other feedforward neural networks, RNNs contain one or more feedback loops and a hidden state (memory) that is updated as each element in an input or output sequence is processed. This structure enables the network to process and remember signals, thereby enabling the model to learn sequential data dependencies.
Given an input vector $x_t$ at time step $t$, the hidden state $h_t$ is computed by using a recurrent formulation:
\begin{equation} \label{eq:rnn1}
    {h_t} = f_{\theta}({h_{t-1}}, {x_t}),
\end{equation}
where $f$ represents the computation of RNNs by using the same learnable parameters $\theta$ for all the time steps. Thus, it can process any sequence of arbitrary length.
The hidden state ${h_{t-1}}$ can be interpreted as a running memory from all previous time steps. Specifically, $h_t$ in a \textbf{Vanilla Recurrent Neural Network} is computed as follows:
\begin{equation} \label{eq:rnn2}
{h_t} = \text{tanh}({W_{xh}}{x_t} + {W_{hh}}{h_{t-1}}),
\end{equation}
where ${W_{xh}}$ and ${W_{hh}}$ are the learnable parameters, and $\textrm{tanh}(\cdot)$ is a hyperbolic activation function. Equation \ref{eq:rnn1} and \ref{eq:rnn2} omits bias vectors for the sake of brevity. 

\begin{figure}[ht]
\centering
\includegraphics[width=0.75\linewidth]{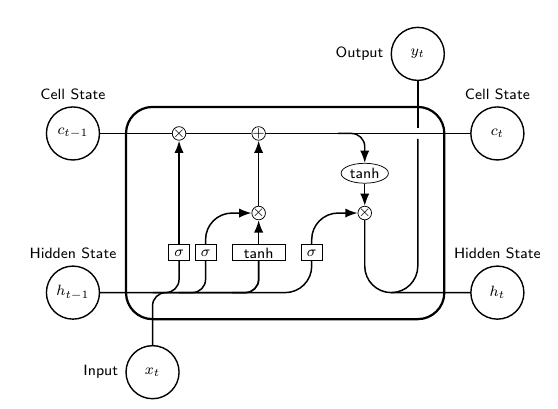}
\caption{Illustration of a Long Short-Term Memory cell.}
\label{fig:prelim-lstm}
\end{figure}
\paragraph{Long Short-Term Memory Networks} The undesirable nature of vanilla RNNs is that the gradients tend to either vanish or explode over long time steps. Long Short-Term Memory (LSTM) networks \cite{hochreiter1997long} are a particular RNN implementation which is designed to handle the RNN limitations. 
In addition to the hidden state $h_t$, it also maintains the memory cell state $c_t$ as follows:
\begin{equation}\label{eq_lstm}
    h_t, c_t = \textrm{LSTM}( x_t , h_{t-1}, c_{t-1}).
\end{equation}
Figure \ref{fig:prelim-lstm} illustrates all the feed-forward computations and internal updates of LSTM. At each time step, the LSTM can choose to read from, write to, or reset the cell using explicit gating mechanisms. Specifically, given $x_t$ as input to a LSTM layer of $N$ hidden units, the $N$-dimensional input gate $i_t$, forget gate $f_t$, output gate $o_t$, and input modulation gate $g_t$ at time step $t$ are updated as
\begin{align} 
i_t &= \textrm{sigm} ( W_{xi} x_t + W_{hi} h_{t-1} + b_i ), \\
f_t &= \textrm{sigm} ( W_{xf} x_t + W_{hf} h_{t-1} + b_f), \\
o_t &= \textrm{sigm} ( W_{xo} x_t + W_{ho} h_{t-1} + b_o), \\
g_t &= \textrm{tanh} (W_{xc} x_t + W_{hc} h_{t-1} + b_c),
\end{align}
where $h_{t-1} \in \mathbb{R}^N$ is the hidden state from the previous time step, $W$ and $b$ are learned weights and biases, and $\textrm{sigm}(\cdot)$ and $\textrm{tanh}(\cdot)$ are sigmoid and tanh functions, respectively. The above gates control the memory cell activation vector $c_t \in \mathbb{R}^N$ and output $h_t \in \mathbb{R}^N$ of the LSTM as follows:
\begin{align}
    c_t &= f_t \odot c_{t-1} + i_t \odot g_t   \\
    h_t &= o_t \odot \textrm{tanh}(c_t),
\end{align}
where $\odot$ represents element-wise multiplication.

\subsection{Transformer Neural Networks}
\label{prelim-transformer}
Transformers \cite{vaswani2017attention} are powerful neural networks with remarkable success in numerous areas across different data modalities, from language \cite{devlin2018bert, radford2018improving} to images \cite{dosovitskiy2020image,liu2021swin}, etc. 
The self-attention mechanism of transformers is their most important success factor.
This mechanism learns the self-alignment between the tokens by calculating the similarity of a given token to all other tokens.
Each token is then updated with a weighted representation of all tokens (including itself). It is observed that the weight value is proportional to each token pair's affinity score. It will be described in detail in the text that follows.

\begin{figure}[ht]
\centering
\includegraphics[width=0.75\linewidth]{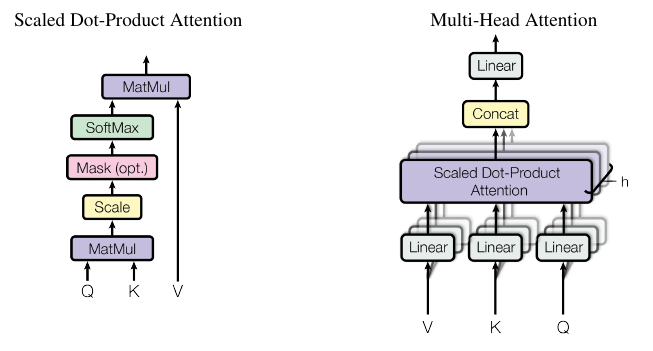}
\caption{(left) Scaled Dot-Product Attention. (right) Multi-Head Attention consists of several attention layers running in parallel. Source: \cite{vaswani2017attention}.}
\label{fig:prelim-attention}
\end{figure}

\textbf{Self-Attention} A self-attention function requires a query $Q$ of dimension $d_k$ and a set of key-value pairs (i.e., $K$, $V$) of dimension $d_v$ for the sequence as inputs. It outputs a weighted sum of the values, in which we associate each value with a weight by computing the similarity between its respective query and key. Specifically, the function computes the dot products of the query with all the keys, divides each by $\sqrt{d_k}$, then applies a softmax function to obtain the normalized weights on the values. The attention function is given by
\begin{equation}
    \text{Attention}({Q}, {K}, {V}) = \text{softmax}(\frac{{Q}{K}^T}{\sqrt{d_k}}){V}.
\end{equation}

Transformers utilize a multi-head attention mechanism, as shown in Figure \ref{fig:prelim-attention}, to learn multiple attended features from different sub-spaces at different positions.
\begin{equation}
\text{MultiHead}({Q}, {K}, {V}) = \text{Concat}(head_1, \ldots, head_h){W^O},
\end{equation}
where $head_i = \text{Attention}({Q}{W^{Q}_{i}}, {K}{W^{K}_{i}}, {V}{W^{V}_{i}})$, and the projections are parameter matrices ${W^{Q}_{i}} \in \mathbb{R}^{d_{m} \times d_k}$, ${W^{K}_{i}} \in \mathbb{R}^{d_{m} \times d_k}$, ${W^{V}_{i}} \in \mathbb{R}^{d_{m} \times d_v}$, and ${W^{O}_{i}} \in \mathbb{R}^{d_{m} \times d_v}$.

\textbf{Position-wise Feed-Forward Networks.} In addition to the Self-Attention sub-layer, a typical transformer layer also includes a Feedforward Network (FFN) that is applied separately and identically to each position. The network consists of two linear transformations separated by a ReLU activation. Its computation is provided by
\begin{equation}
    \text{FFN}({x}) = \text{ReLU}({x}{W_1} + b_1){W_2} + b_2.
\end{equation}

\cleardoublepage
\chapter{GRIT: Integrating Dual Visual Features for Image Captioning}
\label{chapter:ch3}

\section{Introduction}

This chapter presents the first of three research contributions in this dissertation.
As outlined in Chapter~\ref{chapter:ch1}, we study three vision-language tasks that form a progression from passive scene understanding to active embodied interaction: image captioning, visual dialog, and interactive instruction following.
Underlying all three tasks is the need for an agent to form accurate and expressive representations of what it sees; without reliable visual perception, no sophistication in language modeling or decision-making can compensate.
We therefore begin our investigations with image captioning, a task whose performance depends critically on the quality of the visual features extracted from the input image, and one that provides a clean, well-established testbed for studying this dependence.
Beyond its own merits, the work in this chapter serves a broader purpose within the dissertation.
The visual representations developed here, and in particular the insights about how region-based and grid-based features complement each other, carry forward into the subsequent chapters: region-based, object-centric features form the visual backbone of the embodied agent in Chapter~\ref{chapter:ch5}, and the principle of integrating multiple levels of visual abstraction is equally relevant to the multi-input reasoning problem addressed in Chapter~\ref{chapter:ch4}.

The performance of image captioning depends critically on how well the visual features capture what is in the scene. Most existing methods follow a two-stage pipeline: they first extract visual features from the input image and then use those features to generate a description. Two primary approaches to feature extraction have been developed, referred to as grid features \cite{xu2015show,rennie2017self,lu2017knowing} and region features \cite{anderson2018bottom}. Grid features are local representations extracted at regular grid points, typically obtained directly from a higher-layer feature map of a CNN or ViT. Region features are a set of local representations corresponding to regions (i.e., bounding boxes) detected by an object detector.

\begin{figure}[t]
\begin{center}
\includegraphics[width=1.0\linewidth]{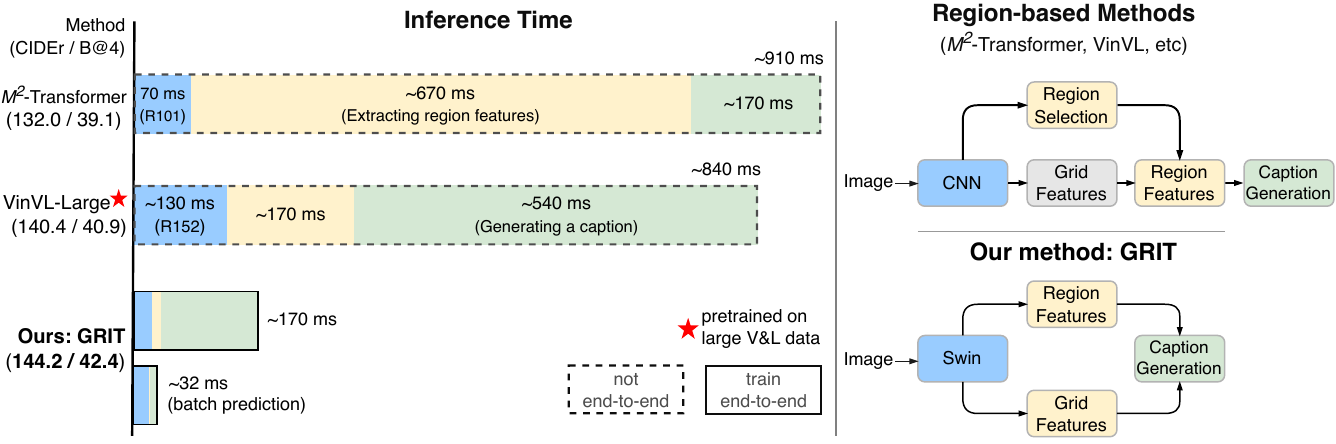}
\end{center}
   \vspace{-0.3cm}
   \caption{Comparison of GRIT and other region-based methods for image captioning. Left: Running time per image {\color{black} of performing inference with beam size of five and the maximum length of 20 on a V100 GPU. Right: Their architectures}
   }
\label{fig:tradeoff}
\end{figure}

The current state-of-the-art methods employ region features because they encode detected object regions directly, and identifying objects and their relations is useful for generating accurate descriptions. However, region features carry three notable limitations. First, they do not convey contextual information such as relations between objects, because the detected regions do not cover the areas between objects. Second, there is a risk of erroneous detection in which important objects may be missed. Third, computing region features is costly, particularly when using a high-performance CNN-based detector such as Faster R-CNN \cite{ren2015faster}.

Grid features, extracted from the entire image as a higher-layer feature map of a backbone network, are free from the first two of these limitations. They naturally capture contextual information across the full image and are not subject to detection errors. Their drawback is that they lack explicit object-level information.

This complementarity motivates integrating both feature types to obtain a more complete image representation. While a few recent studies have explored such integration \cite{luo2021dual,xian2022dual}, the best strategy for combining them remains unclear. In this chapter, we reconsider how to extract each type of feature and then design a principled method for integrating them.

A further limitation concerns CNN-based detectors and training. At the final stage of computation, such detectors apply non-maximum suppression (NMS) to remove redundant bounding boxes. This step makes end-to-end training of the full model difficult, because jointly minimizing a single loss over both the caption decoder and the detector is not straightforward when NMS is involved. Recent studies therefore train the two components separately, first training the detector on object detection and then training the decoder on captioning. This decoupled procedure may limit the final captioning performance.

To overcome these limitations, we adopt the framework of DETR \cite{carion2020end}, which eliminates the need for NMS. We choose Deformable DETR \cite{zhu2021deformable} for its high performance, and replace the original CNN backbone with Swin Transformer \cite{liu2021swin} to extract initial features from the input image. Grid features are also obtained from the same Swin Transformer: its last-layer features are passed through a lightweight self-attention Transformer to model spatial interactions among grid positions, recovering contextual information that the region features do not provide.

The two types of features are then passed to the caption generator, which we design as a lightweight Transformer that produces a caption autoregressively. A cross-attention mechanism within the generator computes and applies attention from both visual feature types to the caption words being decoded.

These components together form a Transformer-only neural architecture we call GRIT (\textbf{G}rid- and \textbf{R}egion-based \textbf{I}mage-captioning \textbf{T}ransformer). Experiments on the COCO benchmark \cite{lin2014microsoft} show that GRIT establishes a new state of the art among methods without vision-and-language (V\&L) pretraining, and performs on par with SimVLM$_\mathrm{huge}$ \cite{wang2021simvlm}, which leverages pretraining on 1.8 billion image-text pairs.

\section{Related Work} 
\subsection{Visual Representations for Image Captioning}

Recent image captioning methods typically employ an encoder-decoder architecture. Specifically, given an image, the encoder extracts visual features; the decoder receives the visual features as inputs and generates a sequence of words. Early methods use a CNN to extract a global feature as a holistic representation of the input image \cite{vinyals2015show,karpathy2015deep}. Although it is simple and compact, this holistic representation suffers from information loss and insufficient granularity. To cope with this, several studies \cite{xu2015show,rennie2017self,lu2017knowing} employed more fine-grained grid-based features to represent input images and also used attention mechanisms to utilize the granularity for better caption generation. Later, Anderson et al. \cite{anderson2018bottom} introduced the method of using an object detector, such as Faster R-CNN, to extract object-oriented features, called region features, showing that this leads to performance improvement in many V\&L tasks, including image captioning and visual question answering. Since then, region features have become the de facto choice of visual representation for image captioning. Pointing out the high computational cost of the region features, Jiang et al. \cite{jiang2020defense} showed that the grid features extracted by an object detector perform well on the VQA task. RSTNet \cite{zhang2021rstnet} has recently applied these grid features to image captioning. 

\subsection{Application of Transformer in Vision/Language Tasks}

Transformer has long been a standard neural architecture in natural language processing \cite{vaswani2017attention,devlin2018bert,radford2018improving}, and started to be
extended to computer vision tasks. 
Besides ViT \cite{dosovitskiy2020image} for image classification,
it was also applied to object detection, leading to DETR \cite{carion2020end}, followed by several variants \cite{zhu2021deformable,fang2021you,song2021vidt}. A recent study \cite{xu2021e2e} applied the framework of DETR to pretraining for various V\&L tasks, where they did not use it to obtain the region features. 

Transformer has been applied to image captioning, where it is used as an encoder for extracting and encoding visual features and a decoder for generating captions. Specifically, Yang et al. \cite{yang2019learning} proposed to use the self-attention mechanism to encode visual features. Li et al. \cite{li2019entangled} used Transformer for obtaining the region features in combination with a semantic encoder that exploits knowledge from an external tagger. Several following studies proposed several variants of Transformer tailored to image captioning, such as Attention on Attention \cite{huang2019attention}, X-Linear Attention \cite{pan2020x}, Memory-augmented Attention \cite{cornia2020meshed}, etc. Transformer is naturally employed also as a caption decoder \cite{herdade2019image,guo2020normalized,luo2021dual,wang2021simvlm}.

\section{Proposed Method}
This section describes the architecture of GRIT (Grid- and Region-based Image-captioning Transformer). It consists of two parts, one for extracting the dual visual features from an input image (Sec.~\ref{sec:extraction}) and the other for generating a caption sentence from the extracted features (Sec.~\ref{sec:generation}). 

\begin{figure}[t]
\begin{center}
\includegraphics[width=1.0\linewidth]{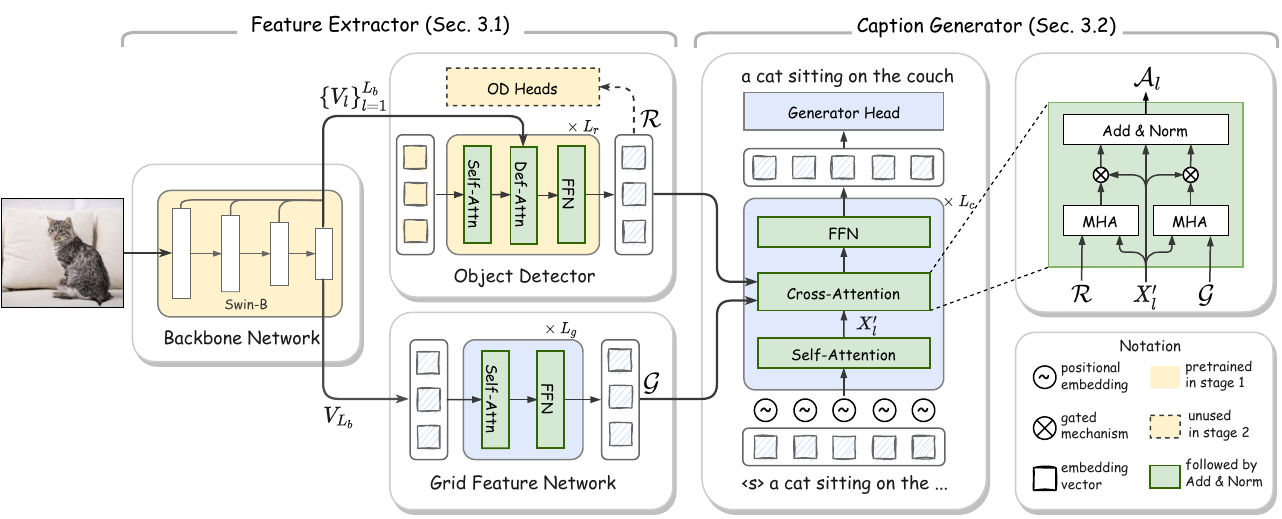}
\end{center}
   \caption{Overview of the architecture of GRIT
   }
\label{fig:overview}
\end{figure}

\subsection{Extracting Visual Features from Images}
\label{sec:extraction}

\subsubsection{Backbone Network for Extracting Initial Features}

A lot of efforts have been made to apply the Transformer architecture to various computer vision tasks since ViT~\cite{dosovitskiy2020image} applied it to image classification. ViT divides an input image into small patches and computes global attention over them. This is not suitable for tasks requiring spatially dense prediction, e.g., object detection since the computational complexity increases quadratically with the image resolution.

Swin Transformer~\cite{liu2021swin} mitigates this issue to a great extent by incorporating operations such as patch reduction and shifted windows that support local attention. It is currently a de facto standard as a backbone network for various computer vision tasks. We employ it to extract initial visual features from the input image in our model. 

We briefly summarize its structure, explaining how we extract features from the input image and send them to the components following the backbone. Given an input image of resolution $H\times W$, Swin Transformer computes and updates feature maps through multiple stages; it uses the {\color{black} patch merging layer} after every stage (but the last stage) to downsample feature maps in their spatial dimension by the factor of 2. We apply another patch merging layer to downsample the last layer's feature map. 
We then collect the feature maps from all the stages, obtaining four multi-scale feature maps, i.e., $\{ V_{l}\}_{l=1}^{L_b}$ where ${\color{black} L_b}=4$, which have the resolution from $H/8 \times W/8$ to $H/64 \times W/64$. These are inputted to the subsequent modules, i.e.,  the object detector and the network for generating grid features.

\subsubsection{Generating Region Features}

{\color{black} As in previous image captioning methods, ours also rely on an object detector to create region features. }
However, we employ a Transformer-based decoder framework, i.e., DETR \cite{carion2020end} instead of CNN-based detectors, such as Faster-RCNN, which is widely employed by the SOTA image captioning models \cite{anderson2018bottom}. DETR formulates object detection as a direct set prediction problem, which makes the model free of the unideal computation for us, i.e., NMS and RoI alignment. This enables the end-to-end training of the entire model from the input image to the final output, i.e., a generated caption, and also leads to a significant reduction in computational time while maintaining the model’s performance on image captioning compared with the SOTA models. 

Specifically, we employ Deformable DETR \cite{zhu2021deformable}, a variant of DETR. Deformable DETR extracts multi-scale features from an input image with its encoder part, which are fed to the decoder part. We use only the decoder part, to which we input the multi-scale features from the Swin Transformer backbone. This leads to further reduction in computational time. 
We will refer this decoder part as ``object detector’’ in what follows; see Fig.~\ref{fig:overview}.

The object detector receives two inputs: the multi-scale feature maps generated by the backbone, and {\color{black} $N$ learnable object queries $R_0 = \{r_i\}_{i=1}^{N}$, in which $r_i \in \R^d$}. Before forwarding them into the object detector, we apply linear transformation to the multi-scale feature maps, mapping them into $d$-dimensional vectors {\color{black} as  $V_l\leftarrow  W_l^r V_l$, where $\{W_l^r\}_{l=1}^{L_b}$ is a learnable projection matrix}.

Receiving these two inputs, the object detector updates the object queries through a stack of $L_r$ deformable layers, yielding $R_{L_r}\in \R^{N\times d}$ from the last layer; see \cite{zhu2021deformable} for details. We use $R_{L_r}\in \R^{N\times d}$ as our region features $\mathcal{R}$. We forward this to the caption generator. 

Although we train it as a part of our entire model, we pretrain our ``object detector'' including the vision backbone on object detection before the training of image-captioning.
For the pretraining, we follow the procedure of Deformable DETR; placing {\color{black} a three-layer MLP and a linear layer on its top to predict box coordinates and class category, respectively.}
We then minimize a set-based global loss that forces unique predictions via bipartite matching.

Following \cite{anderson2018bottom,zhang2021vinvl}, 
we pretrain the model (i.e., our object detector including the vision backbone) in two steps. 
We first train it on object detection following the training method of Deformable DETR. We then fine-tune it on a joint task of object detection and object attribute prediction, aiming to make it learn fine-grained visual semantics with the following loss:
\begin{equation}
\mathcal{L}_{v}(y,\hat{y}) = \sum_{i=1}^{N}[\underbrace{-{\rm log} \hat{p}_{\hat{\sigma}(i)}(c_i) + \mathbf{1}_{c_i\neq\varnothing}
\mathcal{L}_{box} (b_{i}, \hat{b}_{\hat{\sigma}(i)})}_{\rm object \ detection} \underbrace{-{\rm log} \hat{p}_{\hat{\sigma}(i)}(a_i)}_{\rm attribute\ prediction}],
\end{equation}
where $\hat{p}_{\hat{\sigma}(i)}(a_i)$ and $\hat{p}_{\hat{\sigma}(i)}(c_i)$ are the attribute and class probabilities, $\mathcal{L}_{box}(b_{i}$,$\hat{b}_{\hat{\sigma}(i)})$ is the loss for normalized bounding box regression for object $i$~\cite{zhu2021deformable}.
\subsubsection{Grid Feature Network}

This network receives the last one of the multi-scale feature maps from the Swin Transformer backbone, i.e., $V_{L_b}\in \R^{M \times d_{L_b}}$, where $M = H/64 \times W/64$. As with the input to the object detector, we apply a linear transformation with a learnable matrix $W^g\in \R^{d\times d_{L_b}}$ to $V_{L_b}$, obtaining $G_0= W^g V_{L_b}$
We employ the standard self-attention Transformer having $L_g$ layers. This network updates $V_{L_b}$ through these layers, yielding our grid features $\mathcal{G}$ represented as a $M\times d$ matrix. We intend to extract contextual information hidden in the input image by modeling the spatial interaction between the grid features. 

\subsection{Caption Generation Using Dual Visual Features} 
\label{sec:generation}

\subsubsection{Overall Design of Caption Generator}

The caption generator receives the two types of visual features, the region features $\mathcal{R} \in \R^{N \times d}$ and the grid features $\mathcal{ G} \in \R^{M \times d}$, as inputs. Apart from this, we employ the basic design employed in previous studies \cite{vaswani2017attention,herdade2019image} that is based on the Transformer architecture. It generates a caption sentence in an autoregressive manner; receiving the sequence of predicted words (rigorously their embeddings) at time $t-1$, it predicts the next word at time $t$. We employ the sinusoidal positional embedding of time step $t$ \cite{vaswani2017attention}; we add it to the word embedding to obtain the input $x^t_0 \in \R^d$ at $t$. 

The caption generator consists of a stack of $L_c$ identical layers. {\color{black} The initial layer receives the sequence of predicted words and the output from the last layer is input to a linear layer whose output dimension equals the vocabulary size to predict the next word.}

Each transformer layer has a sub-layer of masked self-attention over the sentence words and a sub-layer(s) of cross-attention between them and the visual features in this order, followed by a feedforward network (FFN) sub-layer.
The masked self-attention sub-layer at the $l$-th layer receives an input sequence $\{{x^i_{l-1}}\}_{i=0}^{t}$ 
at time step $t$, 
and computes and applies self-attention over the sequence to update the tokens with the attention mask to
prevent the interaction from the future words during training.

\begin{figure}[t]
\begin{center}
\includegraphics[width=1.0\linewidth]{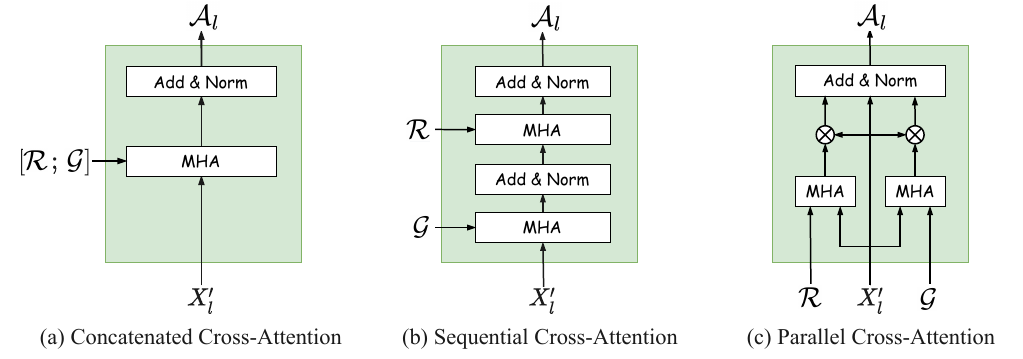}
\end{center}
   \caption{Three designs of cross-attention mechanism to use dual visual features}
\label{fig:cross_attn}
\end{figure}

The cross-attention sub-layer {\color{black}in the layer $l$}, located after the self-attention sub-layer, fuses its output with the dual visual features by cross-attention between them, {\color{black}yielding $\mathcal{ A}_l$.}
We consider the three design choices shown in Fig.~\ref{fig:cross_attn} and described below. We examine their performance through experiments. 
\subsubsection{Cross-attention between Caption Word and Dual Visual Features}
We show three designs of cross-attention between the {\color{black}caption word features} and the dual visual features (i.e., the region features $\mathcal{R}$ and the grid features $\mathcal{G}$) as below. 
\paragraph{Concatenated Cross-Attention}
The simplest approach is to concatenate the two visual features and use the resultant features as keys and values in the standard multi-head attention {\color{black}sub-}layer, where the sentence words serve as queries; see Fig.~\ref{fig:cross_attn}(a).
\paragraph{Sequential Cross-Attention}
Another approach is to perform cross-attention computation separately for the two visual features. The corresponding design is to place two independent multi-head attention {\color{black}sub-}layers in a sequential fashion, and uses one for the grid features and the other for the region features (or the opposite combination); see Fig.~\ref{fig:cross_attn}(b). Note that their order could affect the performance. 
\paragraph{Parallel Cross-Attention}
The third approach is to perform multi-head attention computation on the two visual features in parallel. To do so, we use two multi-head attention mechanisms with independent learnable parameters. The detailed design is as follows. Let $X_{l-1}=\{x^{l-1}_i\}$ be the {\color{black}word features} inputted to the meta-layer $l$ containing this cross attention sub-layer. As shown in Fig.~\ref{fig:overview}, they are first input to the self-attention sub-layer, converted into $X_l'=\{x_i'\}$ (layer index $l$ omitted for brevity) and then input to this cross attention sub-layer. In this sub-layer, multi-head attention (MHA) is computed with $\{x_i'\}$ as queries and the region features $\mathcal{R}$ as keys and values, yielding attended features $\{a^r_i\}$. The same computation is performed in parallel with the grid features $\mathcal{G}$ as keys and values, yielding $\{a^g_i\}$. Next, we concatenate them with $x_i'$ as $[a^r_i;x_i']$ and $[a^g_i;x_i']$, projecting them back to $d$-dimensional vector using learnable affine projections. Normalizing them with sigmoid into probabilities $\{c_i^r\}$ and $\{c_i^g\}$, respectively, we have 
\begin{align} 
    c_i^g &= \mathrm{sigmoid}(W^g[{a^{g}_{i}}; x^{\prime}_{i}] + b^g), \\
    c_i^r &= \mathrm{sigmoid}(W^r[{a^{r}_{i}}; x^{\prime}_{i}] + b^r).
\end{align}
We then multiply them with $\{a^r_i\}$ and $\{a^g_i\}$, add the resultant vectors to $\{x_i'\}$, and finally feed to layer normalization, obtaining $\mathcal{ A}_l=\{a^{(l)}_i\}$ as follows:
\begin{align} 
    a^{(l)}_{i} &= \mathrm{LN}(c^{g}_i \otimes a^{g}_i + c^{r}_i \otimes a^{r}_i + 
    {\color{black}x}^{\prime}_{i}).
    \label{eq:agg}
 \end{align}
\subsubsection{Caption Generator Losses}
Following a standard practice of image captioning studies, we pre-train our model with a cross-entropy loss (XE) and finetune it using the CIDEr-D optimization with self-critical sequence training strategy \cite{rennie2017self}. 
Specifically, the model is first trained to predict the next word $x^*_{t}$ at $t=1..T$, given the ground-truth sentence $x^*_{1:T}$. This is equal to minimize the following XE loss with respect to the model's parameter $\theta$:
\begin{equation}
\mathrm \mathcal{ L}_{XE}(\theta)=-\sum_{t=1}^{T} \log \left(p_{\theta}\left(x_{t}^{*} \mid x^{*}_{0: t-1}\right)\right).
\end{equation}
We then finetune the model with the CIDEr-D optimization, where we use the CIDEr score as the reward and the mean of the rewards as the reward baseline, following \cite{cornia2020meshed}.
The loss for self-critical sequence training is given by
\begin{equation}
    \mathcal{ L}_{RL}(\theta) = -\frac{1}{k}\sum_{i=1}^k (r(\bm{w}^i)-b) \log p(\bm{w}^i),
\end{equation}
where $\bm{w}^i$ is the $i$-th sentence in the beam; $r(\cdot)$ is the reward function; and $b$ is the reward baseline; and $k$ is the number of samples in the batch. 

\section{Experiments}
\subsection{Datasets}
\subsubsection{Object Detection}

As mentioned earlier, we train our object detector (including the backbone) in two steps. In the first step, we train it on object detection using either Visual Genome \cite{krishnavisualgenome} or a combination \cite{zhang2021vinvl} of four datasets: COCO \cite{lin2014microsoft}, Visual Genome, Open Images \cite{OpenImages}, and Object365 \cite{shao2019objects365}, depending on what previous methods we experimentally compare. In the second step, we train the model on object detection plus attribute prediction using Visual Genome. 
Note that following the standard practice, we exclude the duplicated samples appearing in the testing and validation splits of the COCO and {nocaps} \cite{agrawal2019nocaps} datasets to remove data contamination. See the supplementary material for more details.

When pretraining our model on the four datasets (i.e., Visual Genome (VG), COCO, OpenImages, and Objects365), we follow \cite{zhang2021vinvl} to build a unified training corpus with the statistics shown in Table \ref{tab:stat} except that we do not use the annotations from COCO stuff \cite{caesar2018cvpr}. The resultant corpus has 2.49M unique images with 1848 categories.
\begin{table}[H]
    \centering
    \setlength{\tabcolsep}{3.pt}
    \caption{Statistics of the pretraining datasets for object detection.}
    \begin{tabular}{l c  c c c  c}
        \toprule
        Source & &  VG & COCO & Objects365 & OpenImages \\
        \midrule
        Images  & &  97k & 111k & 609k & 1.67M \\
        Categories & &  1594 & 80 & 365 & 500 \\
        Sampling & & $\times 8$ & $\times 8$ & $\times 2$ & $\times 1$ \\
        \bottomrule
    \end{tabular}
    \label{tab:stat}
\end{table}

\subsubsection{Image Captioning}
We conduct our experiments on the COCO dataset, the standard for the research of image captioning  \cite{lin2014microsoft}. The dataset contains 123,287 images, each annotated with five different captions. For offline evaluation, we follow the widely adopted Karpathy split \cite{karpathy}, where {\color{black}113,287, 5,000, and 5,000 images} are used for training, validation, and testing respectively. 

To test our method's effectiveness on other image captioning datasets, we also report the performances on the {nocaps} dataset and the Artemis dataset \cite{achlioptas2021artemis}. See the supplementary material for more details.

\subsection{Implementation Details}
\subsubsection{Evaluation Metrics} 
We employ the standard evaluation protocol for the evaluation of methods. Specifically, we use the full set of captioning metrics: BLEU@N \cite{papineni2002bleu}, METEOR \cite{banerjee2005meteor}, ROUGE-L \cite{lin2004rouge}, CIDEr \cite{vedantam2015cider}, and SPICE \cite{anderson2016spice}.
We will use the abbreviations, B@N, M, R, C, and S, to denote BLEU@N,  METEOR, ROUGE-L, CIDEr, and SPICE, respectively.

\subsubsection{Hyperparameters Settings}
In our model, we set the dimension $d$ of each layer to $512$, the number of heads to eight. We employ dropout with the dropout rate of $0.2$ on the output of {each MHA and FFN sub-layer} following \cite{vaswani2017attention}. 

For the object detector, we set the number of queries $N=150$ and the hidden dimension $d=512$. The backbone network weights are intialized by the weights of Swin-Base ($384\times384$) pretrained on ImageNet21K \cite{liu2021swin}. Following \cite{zhu2021deformable}, the loss for normalized bounding box regression for object $i$, $\mathcal{L}_{box}(b_{i}$,$\hat{b}_{\hat{\sigma}(i)})$ is computed as the weighted summation of a box distance $\mathcal{L}_{l_1}$ and a GIoU loss $\mathcal{L}_{iou}$:
\begin{align} 
    \mathcal{L}_{l_1}(b_{i},\hat{b}_{\hat{\sigma}(i)}) &= ||b_i - \hat{b}_{\hat{\sigma}(i)}||_1, \\
    \mathcal{L}_{iou}(b_{i},\hat{b}_{\hat{\sigma}(i)}) &= 1 -  \big( \frac{|b_{i} \cap  \hat{b}_{\sigma(i)}|}{|b_{i} \cup  \hat{b}_{\sigma(i)}|} - \frac{|{\sf B}(b_{i}, \hat{b}_{\sigma(i)}) \symbol{92} b_{i} \cup  \hat{b}_{\sigma(i)}| }{|{\sf B}(b_{i}, \hat{b}_{\sigma(i)})|} \big), \\
    \mathcal{L}_{box}(b_{i},\hat{b}_{\hat{\sigma}(i)}) &= \alpha_{l_1}\mathcal{L}_{l_1}(b_{i},\hat{b}_{\hat{\sigma}(i)}) + \alpha_{iou}\mathcal{L}_{iou}(b_{i},\hat{b}_{\hat{\sigma}(i)}),
\end{align}
where $\alpha_{l_1} = 5$, $\alpha_{iou} = 2$, and ${\sf B}$ outputs the largest box covering $b_{i}$ and $\hat{b}_{\hat{\sigma}(i)}$. We also employ two training strategies, i.e., iterative bounding box refinement and auxiliary losses; see \cite{zhu2021deformable} for details.

We set the number of layers as $L_r = 6$ for the object detector, as $L_g=3$ for the grid feature network, and as $L_c=3$ for the caption generator. Following previous studies, we convert all the captions to lower-case, remove punctuation characters, and perform tokenization with the SpaCy toolkit \cite{spacy2}. We build the vocabularies, excluding the words which appear less than five times in the training and validation splits.

\subsection{Training Details}

\paragraph{First Stage}
In the first stage, we pretrain the object detector with the backbone. We consider several existing region-based methods for comparison, which employ similar pretraining of an object detector but use different datasets. For a fair comparison, we consider two settings. One uses Visual Genome for training, following most previous methods. We train our detector for 150,000 iterations with a batch size of 32. The other (results indicated with $\dagger$ in what follows) uses the four datasets mentioned above, {\color{black} following \cite{zhang2021vinvl}}. We train the detector for 125,000 iterations with a batch size of 256. In both settings, the input image is resized so that the maximum for the shorter side is 800 and for the longer side is 1333. We use Adam optimizer \cite{kingma2015adam} with a learning rate of $10^{-4}$, decreased by 10 at iteration 120,000 and 100,000 in the first and second settings, respectively. We follow \cite{zhu2021deformable} for other training procedures. After this, we finetune the models on object detection plus attribute prediction using Visual Genome for additional five epochs with a learning rate of $10^{-5}$, following \cite{anderson2018bottom,zhang2021vinvl}. The supplementary material presents the details of implementation and experimental results on object detection.

\paragraph{Second Stage}
We train the entire model for the image-captioning task in the second stage. We employ the standard method for word representation, i.e., linear projections of one-hot vectors to vectors of dimension $d$ = 512.  
In this stage, we resize all the input images so that the maximum dimensions for the shorter side and longer side are 384 and 640, respectively. We train models, as explained earlier. Specifically, we train models with the cross-entropy loss $\mathcal{L}_{XE}$ for ten epochs, in which we warm up the learning rates for the grid feature network and the caption generator from $10^{-5}$ to $10^{-4}$ {\color{black} in the first epoch}, while we fix those for the backbone network and the object detector at $10^{-5}$. Then, we finetune the model based on the CIDEr-D optimization for ten epochs, where we set the fixed learning rate to $5\times 10^{-6}$ for the entire model.
We use the Adam optimizer~\cite{kingma2015adam} with a batch size of $128$. For the inference inside the CIDEr-D optimization, we use beam search with a beam size of five and a maximum length of 20.

\subsection{Object Detection Results}
\vspace{-1em}
\begin{table}[H]
    \centering
    \caption{Performance of object detection on the COCO and Visual Genome datasets. `4DS' denotes the four object detection datasets.}
    \setlength{\tabcolsep}{3.pt}
    \begin{tabular}{l c  c c c  c}
        \toprule
        Model & &  Training Data & mAP (COCO) & mAP$^{50}$ (VG)  \\
        \midrule
        BUTD \cite{anderson2018bottom} & &  VG & - & 10.2 \\
        VinVL \cite{zhang2021vinvl} & &  4DS & 50.5 & 13.8 \\
        \rowcolor{LightCyan}
        GRIT & & VG  & 33.6 & 14.2 \\
        \rowcolor{LightCyan}
        GRIT$^\dag$ & & 4DS & 50.8 & 15.1 \\
        \bottomrule
    \end{tabular}
    \label{tab:od_results}
\end{table}

Table \ref{tab:od_results} shows the performance on the COCO validation split and the Visual Genome test split of our object detector compared with VinVL and BUTD \cite{anderson2018bottom}. It is seen that the object detector of GRIT attains comparable or higher performance on the two datasets as compared with BUTD and VinVL when pretrained on the similar datasets.

\subsection{Performance of Different Configurations }

\begin{table}[H]
    \caption{Results of ablation tests on the COCO test split. All the models are trained with the XE loss and finetuned by the CIDEr optimization 
    }
    \resizebox{1.0\columnwidth}{!}{
    \setlength{\tabcolsep}{2.pt}
    \begin{tabular}[!t]{cccc}
            \begin{tabular}[!t]{l|c|cc}
            \multicolumn{4}{c}{\normalsize \centering (a)}\\
            \toprule
            Factor & Choice & CIDEr & B@4  \\

            \midrule
            (1) \textbf{Backbone Network}       &  ImageNet & 135.5             & 41.5 \\
            \quad - Training data               &  VG       & 142.3             & 41.9 \\
                                                &  4DS   & \textbf{144.2}    & \textbf{42.4} \\
            \midrule
            (2) \textbf{Region features}            & 50     & 141.4       & 41.9 \\
            \quad - Number of vectors               & 100    & 141.8       & 41.5 \\
                    \quad \ \ (trained on VG)                 & 150    & \textbf{142.3}       & \textbf{41.9} \\
            \midrule 
            (3) \textbf{Training strategy}        && \\
            \quad - {\color{black}End-to-end training} & Yes  & \textbf{144.2}   & {42.4} \\
             & No  & 139.6   & \textbf{42.7} \\
            \bottomrule 
            \end{tabular}
    & & &
        \begin{tabular}{l|c|cc}
            \multicolumn{4}{c}{\normalsize \centering (b)}\\
            \toprule
            Cross Attention  & Choice & CIDEr & B@4  \\
            \midrule

            (1) \textbf{Concatenated} &  $\mathcal{G}$             & 142.1 & 41.7 \\
             \quad - Visual features        
                                            & $\mathcal{R}$             & 142.9 & 41.9 \\
                                            & [$\mathcal{G}$ ; $\mathcal{R}$]       & 143.1 & 41.9 \\
            \midrule
            (2) \textbf{Sequential} & & \\
            \quad - Sequential order    &  $\mathcal{G}$ $\rightarrow$ $\mathcal{R}$  & 144.0 & 42.1 \\
                                &  $\mathcal{R}$ $\rightarrow$ $\mathcal{G}$  & 143.6 & 42.1 \\
            \midrule
            (3) \textbf{Parallel} & & \\
            \quad - Gated activation         &  Sigmoid  & \textbf{144.2} & \textbf{42.4} \\
                                          &  Identity & 143.9 & 41.6 \\
        \bottomrule
        \end{tabular}
    \end{tabular}
    \label{tab:ablation}
    }
\end{table}

Our method has several design choices. We conduct experiments to examine which configuration is the best. The results are shown in Table \ref{tab:ablation}.
We used an identical configuration unless otherwise noted. Specifically, we use the feature extractor pretrained on the four datasets and parallel cross-attention for fusing the region and grid features.

\paragraph{Effects of Object Detection Datasets}
The first block of Table \ref{tab:ablation}(a) shows the effects of different (pre)training strategies of the visual backbone on image-captioning performance. The `ImageNet' column shows the result of the model using a Swin Transformer backbone pretrained on ImageNet21K and the grid features alone; `VG' and `4DS' indicate the models with a detector pretrained on Visual Genome and the four datasets, respectively. They show that using more datasets leads to better performance.

\paragraph{Effects of Number of Region Features}
The second block of Table \ref{tab:ablation}(a) shows the effects of the number of object queries, or equivalently region features. The performance increases as they vary as 50, 100, and 150. 
We also confirmed that {\color{black}the }performance is saturated for  more region features, while the computational cost and false detection increase. 

\paragraph{Impact of End-to-end Training}
The third block shows the effect of the end-to-end training of the entire model. `Yes' indicates the end-to-end training of the entire model and `No' indicates training the model but the vision backbone. The results show that
the end-to-end training considerably improves CIDEr score (from 139.6 to 144.3) with little sacrifice of B@4. This validates our expectation about the effectiveness of the end-to-end training; it arguably helps reduce the domain gap between object detection and image captioning. 

\paragraph{Effects of Fusion of Dual Visual Features}
The first block of Table \ref{tab:ablation}(b) shows the performances of the model employing the concatenated cross-attention and its two variants using the grid features alone or the region features alone. They show that the region features alone work better than the grid features alone, and their fusion achieves the highest performance.  

\paragraph{Effects of Cross-attention Architecture} 

The three blocks of Table \ref{tab:ablation}(b) show the performances of the three cross-attention architectures explained in Sec.~\ref{sec:generation}. The second block shows the two variants of the sequential cross-attention, and the third block shows the two variants of the parallel cross-attention with different {\color{black} gated activation functions}, i.e., sigmoid and identity. By identity activation, we mean setting all the values of $c^g_l$ and $c^r_l$ in Eq.(\ref{eq:agg}) to one. These results show that the parallel cross-attention with sigmoid activation function performs the best; the sequential cross-attention in the order $\mathcal{G} \rightarrow \mathcal{R}$ attains the second best result. 

\subsection{Results on the COCO Dataset}

We next show complete results on the COCO dataset by 
the offline and online evaluations. We present example results in the supplementary material.

\begin{table}[H]
\caption{Offline results evaluated on the COCO Karpathy test split. `V. E. type' indicates the type of visual features; 
{`\# VL Data' is the number of image-text pairs used for vision-language pretraining. 
}
}
    \centering
    \resizebox{\textwidth}{!}{
    \begin{tabular}{l c c c c c c c c c}
    \toprule
        \multirow{2}{*}{Method} &  & {V. E.} & {\# VL} & 
        \multicolumn{6}{c}{Performance Metrics}\\
        \cmidrule(lr){5-10}
         &  & Type & Data &  B@1 & B@4& M & R& C & S\\
        
    \midrule
        {\color{black} {w/ VL pretraining}}\\
        {\color{black} UVLP} \cite{zhou2020unified} &  & {\color{black} $\mathcal{R}$} & {\color{black}3.0M} & {\color{black}-} & {\color{black}39.5} & {\color{black}29.3} & {\color{black}-} & {\color{black}129.3} & {\color{black}23.2} \\ 
        
        {\color{black} Oscar$_{\mathrm{base}}$} \cite{li2020oscar} &  &{\color{black}$\mathcal{R}$} & {\color{black}6.5M} & {\color{black}-} & {\color{black}40.5} & {\color{black}29.7} & {\color{black}-} & {\color{black}137.6} & {\color{black}22.8} \\
        
        {\color{black} VinVL$^\dag_{\mathrm{large}}$\cite{zhang2021vinvl}} &  & {\color{black} $\mathcal{R}$} & {\color{black}8.9M} &-& {\color{black}\textbf{41.0}} & {\color{black}31.1} & - & {\color{black}140.9} & {\color{black}{25.2}} \\
        
        {\color{black} SimVLM$_{\mathrm{huge}}$ \cite{wang2021simvlm}} &  & {\color{black}$\mathcal{G}$} & {\color{black}1.8B} & - & {\color{black}40.6} & {\color{black}\textbf{33.7}} & - & {\color{black}\textbf{143.3}} & \textbf{25.4} \\
        
    
    \midrule
        {w/o VL pretraining}\\
        SAT \cite{vinyals2015show} &  & $\mathcal{G}$ &- & - &31.9 & 25.5 & 54.3&106.3 & - \\
        SCST \cite{rennie2017self} &  & $\mathcal{G}$ &- & - &34.2 & 26.7 & 55.7&114.0 & - \\
        LSTM-A \cite{yao2017boosting}&  & $\mathcal{G}$ &- &78.6& 35.5 & 27.3 & 56.8&118.3 & 22.0 \\
        
        RSTNet \cite{zhang2021rstnet} &  & $\mathcal{G}$ & - & 81.8 &  40.1 & 29.8 & 59.5 & 135.6  & 23.0 \\ 
    
        Up-Down \cite{anderson2018bottom} &  & $\mathcal{R}$ &- &79.8& 36.3 & 27.7 & 56.9&120.1 & 21.4 \\
        RFNet \cite{ke2019reflective} &  & $\mathcal{R}$ &- &79.1& 36.5 & 27.7 & 57.3&121.9 & 21.2 \\
        GCN-LSTM \cite{yao2018exploring} &  & $\mathcal{R}$ & -&80.5& 38.2 & 28.5 & 58.3&127.6 &22.0 \\
        LBPF \cite{qin2019look} &  & $\mathcal{R}$ &- &80.5& 38.3 & 28.5 & 58.4&127.6 & 22.0 \\
        
        SGAE \cite{yang2019auto} &  & $\mathcal{R}$ &- &80.8& 38.4 & 28.4 & 58.6&127.8 & 22.1 \\
        AoA \cite{huang2019attention} &  & $\mathcal{R}$ & - &80.2 & 38.9 & 29.2 & 58.8 & 129.8 & 22.4 \\
        GET \cite{ji2021improving} &  & $\mathcal{R}$ & - & 81.5 & 39.5 & 29.3 & 58.9 & 131.6 & 22.8 \\
        
        ORT \cite{herdade2019image} &  & $\mathcal{R}$ & -& 80.5 & 38.6 & 28.7 & 58.4&128.3 & 22.6 \\
        ETA \cite{li2019entangled} &  & $\mathcal{R}$ & -& 81.5 & 39.3 & 28.8 & 58.9& 126.6  & 22.6 \\
        $\mathcal{M}^{2}$ Transformer \cite{cornia2020meshed} &  & $\mathcal{R}$ & -& 80.8 & 39.1 & 29.2 & 58.6 & 131.2 & 22.6 \\
        X-LAN \cite{pan2020x} &  & $\mathcal{R}$ & -& 80.8 & 39.5 & 29.5 & 59.2 & 132.0 & 23.4 \\
        TCIC \cite{fan2021tcic} &  & $\mathcal{R}$ & - & 81.8 & 40.8 & 29.5 & 59.2 & 135.4 & 22.5 \\
        Dual Global \cite{xian2022dual} &  & $\mathcal{R}$+$\mathcal{G}$ & -& 81.3 & 40.3 & 29.2 & 59.4 & 132.4 & 23.3 \\
        DLCT \cite{luo2021dual} &  & $\mathcal{R}$+$\mathcal{G}$ &-& 81.4 & 39.8 & 29.5 & 59.1 & 133.8 & 23.0 \\
        \rowcolor{LightCyan}
        GRIT  &  &$\mathcal{R}$+$\mathcal{G}$ & - & 83.5 & 41.9 & 30.5  & 60.5 & 142.2 & {24.2} \\ 
        \rowcolor{LightCyan}
        GRIT$^\dag$ &  &$\mathcal{R}$+$\mathcal{G}$ & - &\textbf{84.2}&\textbf{42.4} &\textbf{30.6} &\textbf{60.7} &\textbf{144.2} & \textbf{24.3} \\ 
    \bottomrule
    \end{tabular}}
    \label{tab:offline_test}
\end{table}

\paragraph{Offline Evaluation} 

Table \ref{tab:offline_test} shows the performances
of our method and the current state-of-the-art methods on the offline Karpathy test split.
The compared methods are as follows: grid-based methods \cite{vinyals2015show,rennie2017self,yao2017boosting,zhang2021rstnet}, region-based methods \cite{anderson2018bottom,ke2019reflective,yao2018exploring,qin2019look,yang2019auto,huang2019attention,huang2019attention,guo2020normalized,ji2021improving,herdade2019image,li2019entangled,cornia2020meshed,pan2020x,fan2021tcic}, the methods employing both grid and region features \cite{xian2022dual,luo2021dual}, and also the methods relying on large-scale pretraining on vision and language (V\&L) tasks using a large image-text corpus \cite{zhou2020unified,li2020oscar,zhang2021vinvl}, including SimVLM$_\mathrm{huge}$, a model 
pretrained on an extremely large dataset (i.e., 1.8 billion image-caption pairs) \cite{wang2021simvlm}.

For fair comparison with the region-based methods, we report the results of two variants of our model,  one with the object detector pretrained on Visual Genome alone and the other (marked with $^\dagger$) with the object detector pretrained on the four datasets, as explained earlier. 
It is seen from Table \ref{tab:offline_test} that our models, regardless of the datasets used for the detector's pretraining, outperform all the methods that do not use large-scale pretraining of vision and language tasks (i.e., the methods in the second block entitled `w/o VL pretraining'). Moreover, our model with the detector pretrained solely on Visual Genome (i.e., `GRIT') performs better than those relying on large-scale V\&L pretraining but SimVLM$_\mathrm{huge}$. Finally, our model with the pretrained detector on multiple datasets (i.e., `GRIT$^\dagger$') outperforms SimVLM$_\mathrm{huge}$ leveraging large-scale V\&L pretraining in CIDEr score (i.e., 144.2 vs 143.3). 

\paragraph{Online Evaluation} 

{\color{black}
We also evaluate our models (i.e., a single model and an ensemble of six models) on the 40K testing images by submitting their results on the official evaluation server. Table \ref{tab:online_test} shows the results and those of all the published methods on the leaderboard. Table \ref{tab:online_test} presents the metric scores based on five (c5) and 40 reference captions (c40) per image. We can see that our method achieves the best scores for all the metrics. Note that even our single model outperforms all the published methods that use ensembles.}

\begin{table}[H]
\caption{Online evaluation results on the COCO image captioning dataset.
}
\centering
\resizebox{\textwidth}{!}{
\begin{tabular}{lcllllllllllllll}
\toprule
\multicolumn{1}{l}{\multirow{2}{*}{Method}} & \multirow{2}{*}{Ensemble} & \multicolumn{2}{c}{B-1}   & \multicolumn{2}{c}{B-2}   & \multicolumn{2}{c}{B-3}   & \multicolumn{2}{c}{B-4}   & \multicolumn{2}{c}{M} & \multicolumn{2}{c}{R} & \multicolumn{2}{c}{C} \\
\multicolumn{1}{c}{}& & \multicolumn{1}{c}{c5} & \multicolumn{1}{c}{c40} & \multicolumn{1}{c}{c5} & \multicolumn{1}{c}{c40} & \multicolumn{1}{c}{c5} & \multicolumn{1}{c}{c40} & \multicolumn{1}{c}{c5} & \multicolumn{1}{c}{c40} & \multicolumn{1}{c}{c5} & \multicolumn{1}{c}{c40} & \multicolumn{1}{c}{c5} & \multicolumn{1}{c}{c40} & \multicolumn{1}{c}{c5} & \multicolumn{1}{c}{c40} \\ \hline
{\color{black} w/ VL pretraining}\\[0.05cm]
{\color{black}VinVL$_\mathrm{large}$ \cite{zhang2021vinvl}} & {\color{black}\xmark} & {\color{black}81.9} & {\color{black}96.9} & {\color{black}66.9} & {\color{black}92.4} & {\color{black}52.6} & {\color{black}84.7} & {\color{black}40.4} & {\color{black}74.9} & {\color{black}30.6} & {\color{black}40.8} & {\color{black}60.4} & {\color{black}76.8} & {\color{black}134.7}  & {\color{black}138.7} \\
\midrule
 {w/o VL pretraining}\\
SCST \cite{rennie2017self} & \checkmark & 78.1& 93.7 & 61.9& 86.0 & 47.0& 75.9 & 35.2& 64.5 & 27.0& 35.5 & 56.3& 70.7 & 114.7   & 116.7\\
Up-Down \cite{anderson2018bottom}& \checkmark & 80.2& 95.2 & 64.1& 88.8 & 49.1& 79.4 & 36.9& 68.5 & 27.6& 36.7 & 57.1& 72.4 & 117.9   & 120.5\\
HAN \cite{wang2019hierarchical} &\checkmark & 80.4& 94.5 & 63.8& 87.7 & 48.8   & 78.0& 36.5& 66.8 & 27.4& 36.1 & 57.3& 71.9 & 115.2   & 118.2\\
GCN-LSTM \cite{yao2018exploring} &\checkmark & 80.8& 95.2 & 65.5& 89.3 & 50.8& 80.3 & 38.7& 69.7 & 28.5& 37.6 & 58.5& 73.4 & 125.3   & 126.5\\
SGAE \cite{yang2019auto} & \checkmark &81.0& 95.3 & 65.6& 89.5 & 50.7& 80.4 & 38.5& 69.7 & 28.2& 37.2 & 58.6& 73.6 & 123.8   & 126.5\\
AoA \cite{huang2019attention}  & \checkmark &81.0& 95.0 & 65.8& 89.6 & 51.4& 81.3 & 39.4& 71.2 & 29.1& 38.5 & 58.9& 74.5 & 126.9   & 129.6\\
$\mathcal{M}^{2}$Trans. \cite{cornia2020meshed}  &\checkmark & 81.6& 96.0 & 66.4& 90.8 & 51.8& 82.7 & 39.7& 72.8 & 29.4& 39.0 & 59.2& 74.8 & 129.3   & 132.1\\
X-LAN \cite{pan2020x} &\checkmark & 81.9& 95.7 & 66.9& 90.5 & 52.4& 82.5 & 40.3& 72.4 & 29.6& 39.2 & 59.5& 75.0 & 131.1   & 133.5   \\
DLCT \cite{luo2021dual} &\checkmark & {82.4}  & {96.6}   & {67.4}  & {91.7}   & {52.8}  & {83.8}   & {40.6}  & {74.0}   & {29.8}  & {39.6}   & {59.8}  & {75.3}   & {133.3} & {135.4}  \\ 
\rowcolor{LightCyan}
GRIT$^\dag$ & \xmark & {83.7} & {97.4}   & {68.5}  & {92.8}   & {53.9}  & {85.3}   & {41.5}  & {75.6}   & {30.3}  & {40.2}   & {60.2}  & {75.9}   & {138.3} & {141.8}  \\ 
\rowcolor{LightCyan}
GRIT$^\dag$ & \checkmark & \textbf{84.1} & \textbf{97.6}   & \textbf{69.4}  & \textbf{93.5}   & \textbf{54.9}  & \textbf{86.3}   & \textbf{42.5}  & \textbf{76.8}   & \textbf{30.9}  & \textbf{41.0}   & \textbf{61.2}  & \textbf{77.1}   & \textbf{141.3} & \textbf{143.8}  \\ 
\bottomrule
\label{tab:online_test}
\end{tabular}
}
\end{table}

\subsection{Results on the ArtEmis and {nocaps} Datasets}

As explained above, we evaluate our method on the ArtEmis and nocaps datasets. For ArtEmis, we train the model in the same way as COCO except for the number of training epochs, precisely, five epochs each for the training with the XE loss and that with the CIDEr-D optimization. For nocaps, we evaluate zero-shot inference performance, i.e., the performance of the model trained on COCO. 

Table \ref{tab:other}(a) shows the results of our method on the test split of ArtEmis \cite{achlioptas2021artemis}. It also show the results of existing methods {\color{black} reported in \cite{achlioptas2021artemis}}, which are grid-based \cite{mathews2016senticap,vinyals2015show}, region-based \cite{cornia2020meshed}, and a nearest neighbor method using a holistic vector to encode images (denoted as $\mathcal{H}$).  Our method outperforms all these methods by a large margin.

Table \ref{tab:other}(b) shows the results on the nocaps dataset, including the baseline methods reported in \cite{agrawal2019nocaps,cornia2020meshed}. All the models are trained on the training split of the COCO datasets and tested on the validation split of nocaps, which consists of images with novel objects and captions with unseen vocabularies.
Our method surpasses all the other methods including region-based methods \cite{lu2018neural,anderson2018bottom,cornia2020meshed} in both in-domain and out-of-domain images.

In brief, our model achieves the best performance across all the metrics in the both datasets among the compared methods. These results confirm the effectiveness of our method on datasets in different domains. See the supplementary material for the full results.

\begin{table}[H]
    \caption{Performance on the ArtEmis and {nocaps} datasets.}
    \vskip -0.05in
    \resizebox{1.0\columnwidth}{!}{
    \setlength{\tabcolsep}{1.pt}
\begin{tabular}[!t]{cccc}
    \begin{tabular}{l c c c c c c c}
        \multicolumn{8}{c}{a) Performance on the ArtEmis test split}\\
        \toprule
        \multirow{2}{*}{Method} & 
        {V. E.} & 
        \multicolumn{6}{c}{Performance Metrics}\\
        \cmidrule(lr){3-8}
         & Type &  B@1 & B@2 & B@3 & B@4 & M & R \\
        \midrule
        NN \cite{achlioptas2021artemis} & $\mathcal{H}$ & 36.4 & 13.9 & 5.4 & 2.2 & 10.2 & 21.0 \\
        ANP \cite{achlioptas2021artemis} & $\mathcal{G}$ & 39.6 & 13.4 & 4.2 & 1.4 & 8.8 & 20.2 \\
        SAT \cite{achlioptas2021artemis} & $\mathcal{G}$ & 53.6 & 29.0 & 15.5 & 8.7 & 14.2 & 29.7  \\
        $\mathcal{M}^{2}$Trans. \cite{achlioptas2021artemis} & $\mathcal{R}$ & 50.7  & 28.2  & 15.9 & 9.5 & 13.7 & 28.0 \\
        \rowcolor{LightCyan}
        GRIT$^\dag$ &$\mathcal{R}$+$\mathcal{G}$ &\textbf{70.1}&\textbf{40.1} &\textbf{20.9} &\textbf{11.3} &\textbf{16.8} & \textbf{33.3} \\ 
    \bottomrule
    \end{tabular}
    & & &
    \begin{tabular}{l c c c c c c c}
        \multicolumn{8}{c}{b) Performance on the {nocaps} validation split}\\
        \toprule
        \multirow{2}{*}{Method}  &V.E& \multicolumn{2}{c}{In-Domain} & \multicolumn{2}{c}{Out-Domain} & \multicolumn{2}{c}{Overall}  \\
        \cmidrule(lr){3-4} \cmidrule(lr){5-6} \cmidrule(lr){7-8}
         & Type &  C & S & C & S & C & S \\
        \midrule
        NBT \cite{agrawal2019nocaps} & $\mathcal{R}$ & 62.7 & 10.1 & 54.0 & 8.6 & 53.9 & 9.2 \\
        Up-down \cite{agrawal2019nocaps} & $\mathcal{R}$ & 78.1 & 11.6 & 31.3 & 8.3 & 55.3 & 10.1 \\
        Trans. \cite{cornia2020meshed} & $\mathcal{R}$ & 78.0 & 11.0 & 29.7 & 7.8 & 54.7 & 9.8  \\
        $\mathcal{M}^{2}$Trans. \cite{cornia2020meshed} & $\mathcal{R}$ & 85.7  & 12.1  & 38.9 & 8.9 & 64.5 & 11.1 \\
        \rowcolor{LightCyan}
        GRIT$^\dag$ &$\mathcal{R}$+$\mathcal{G}$  &\textbf{105.9}&\textbf{13.6} &\textbf{72.6} &\textbf{11.1} &\textbf{90.2} & \textbf{12.8} \\ 
    \bottomrule
    \end{tabular}
\end{tabular}

    \label{tab:other}
    
    }
\end{table}

\subsection{Computational Efficiency} 
\paragraph{Technical Measurement Details} We measured the inference time of GRIT and two representative region-based methods, VinVL \cite{zhang2021vinvl} and $\mathcal{M}^2$ Transformer \cite{cornia2020meshed}. It is the computational time per image from image input to caption generation. Specifically, we measured the time to generate a caption of length 20 with a beam size of five on a V100 GPU and averaged it over 1K times. The input image resolution was set to $800 \times 1333$ for VinVL and $\mathcal{M}^2$ Transformer as reported in \cite{anderson2018bottom,zhang2021vinvl}. We set it to $384\times640$ for GRIT since it already achieves higher accuracy.

We measured the inference time of GRIT and two representative region-based methods, VinVL \cite{zhang2021vinvl} and $\mathcal{M}^2$ Transformer \cite{cornia2020meshed}, on the same machine having a Tesla V100-SXM2 of 16GB memory with CUDA version 10.0 and Driver version 410.104. It has Intel(R) Xeon(R) Gold 6148 CPU. The comparison was conducted following \cite{jiang2020defense,kim2021vilt}. Specifically, we excluded the time of preprocessing the image and loading it to the GPU device. Also, the images are rescaled to the resolutions such that all the compared methods achieve its highest performance for image captioning.
For the compared methods, we used the official implementations of $\mathcal{M}^2$ Transformer\footnote{\url{https://github.com/aimagelab/meshed-memory-transformer}} and VinVL\footnote{\url{https://github.com/pzzhang/VinVL}}.
Regarding feature extraction, we extracted the region features from Faster R-CNN using the original implementation\footnote{\url{https://github.com/peteanderson80/bottom-up-attention}} used by $\mathcal{M}^2$ Transformer and another implementation\footnote{\url{https://github.com/microsoft/scene\_graph\_benchmark}} used by VinVL.

\paragraph{Inference Time Comparison} Figure ~\ref{fig:tradeoff} shows the breakdown of the inference time for the three methods. GRIT reduces the time for feature extraction by a factor of \textbf{10 and more} compared with the others.
Similar to $\mathcal{M}^2$ Transformer, GRIT has a lightweight caption generator and thus spends much less time than VinVL for generating a caption after receiving the visual features. GRIT can run with minibatch size up to $64$ on a single V100 GPU, while others cannot afford large minibatch. With minibatch size $\geq32$, the per-image inference time decreases to about 32ms. 

It is seen that VinVL and $\mathcal{M}^2$ Transformer spend considerable time on feature extraction due to the forward pass through the CNN backbone with high resolution inputs and the computationally expensive regional operations. It is also noted that VinVL introduced class-agnostic NMS operations, which reduce a great amount of time consumed by class-aware NMS operations in the standard Faster R-CNN. On the other hand, we employ a Deformable DETR-based detector to extract region features without using all such operations. Table \ref{tab:extraction} shows the comparison on feature extraction.

\begin{table}[H]
    \centering
    \setlength{\tabcolsep}{3.pt}
    \caption{The inference time on feature extraction of different methods.}
    \begin{tabular}{l c c c c c c c r} 
        \toprule
        Method &  Backbone & Detector & Regional Operations & Inference Time \\
        \midrule
        VinVL$_\mathrm{large}$\cite{zhang2021vinvl} & ResNeXt-152 & Faster R-CNN & Class-Agnostic NMS & 304 ms \\
        & & & RoI Align, etc &  \\
        $\mathcal{M}^2$ Trans. \cite{cornia2020meshed} & ResNet-101 & Faster R-CNN & Class-Aware NMS & 736 ms \\
        & & & RoI Align, etc &  \\
        \rowcolor{LightCyan}
        GRIT & Swin-Base & DETR-based & - & 31 ms \\
        \bottomrule
    \end{tabular}
    \label{tab:extraction}
\end{table}

Regarding caption generation, all the methods use beam search as the decoding strategy, with beam size of 5 and the maximum caption length of 20. 
Both $\mathcal{M}^2$ Transformer and GRIT employ a lightweight caption generator (caption decoder) having only 3 transformer layers with hidden dimension of 512 while VinVL$_\mathrm{large}$ has 24 transformer layers with hidden dimension of 1024; see Table \ref{tab:generator}. Thus, with the visual features as inputs, $\mathcal{M}^2$ Transformer and GRIT spend less inference time generating words than VinVL$_\mathrm{large}$ in the autoregressive manner.

\begin{table}[H]
    \centering
    \setlength{\tabcolsep}{3.pt}
    \caption{The inference time on caption generation of different methods.}
    \begin{tabular}{l c c c c c c c} 
        \toprule
        Method &  No. of Layers & Hidden Dim. & Inference Time \\
        \midrule
        VinVL$_\mathrm{large}$\cite{zhang2021vinvl} & 24 & 1024 & 542 ms \\
        $\mathcal{M}^2$ Transformer \cite{cornia2020meshed} & 3 & 512 & 174 ms \\
        \rowcolor{LightCyan}
        GRIT & 3 & 512 & 138 ms \\
        \bottomrule
    \end{tabular}
    \label{tab:generator}
\end{table}

\begin{figure}[ht]
\begin{center}
\includegraphics[height=0.9\textheight]{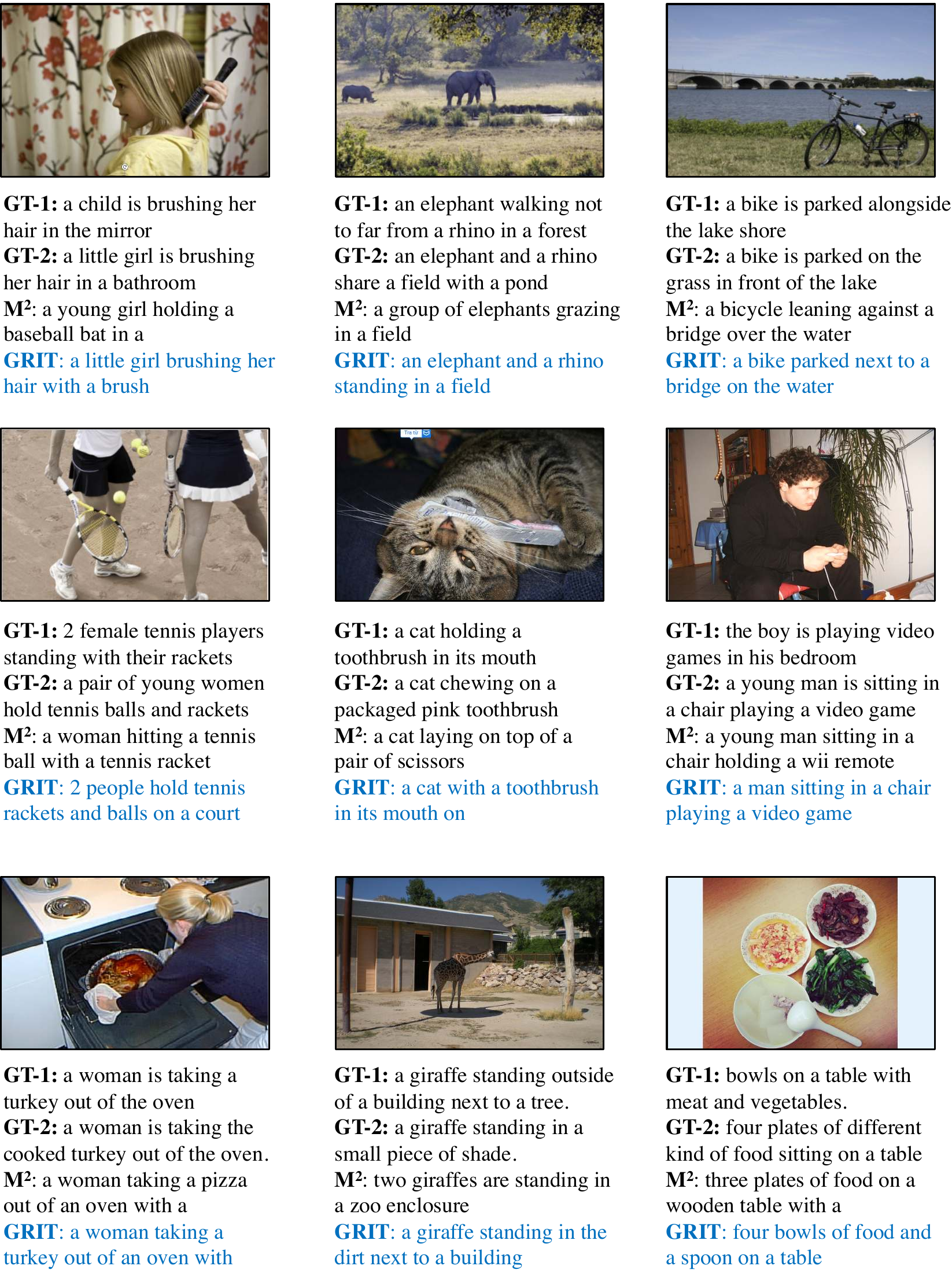}
\end{center}
   \caption{Qualitative examples from our method (GRIT) and a region-based method ($\mathcal{M}^2$ Transformer) on the COCO test images. Zoom in for  better view.}
\label{fig:res1}
\end{figure}

\begin{figure}[ht]
\begin{center}
\includegraphics[height=0.9\textheight]{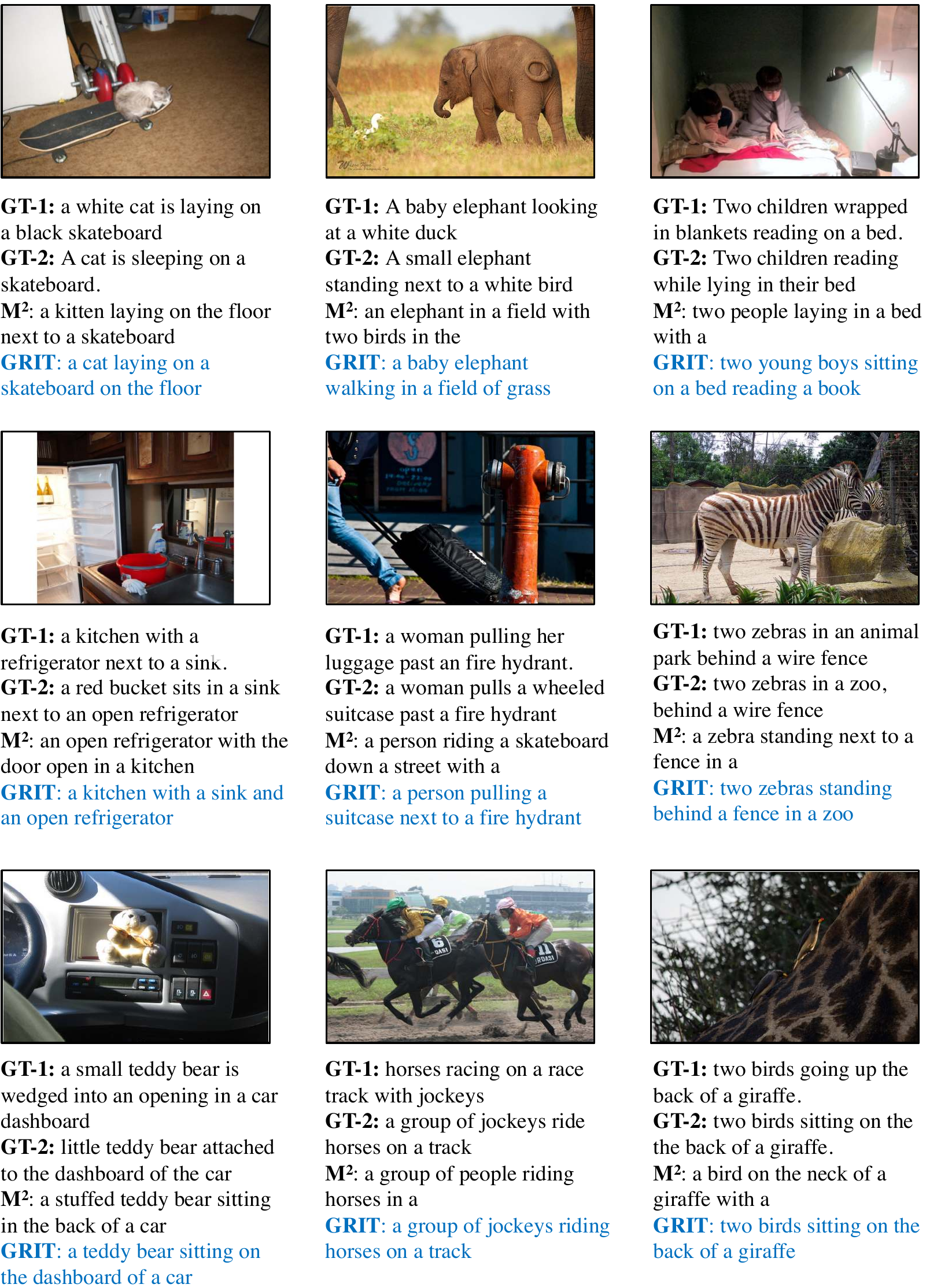}
\end{center}
   \caption{Qualitative examples from our method (GRIT) and a region-based method ($\mathcal{M}^2$ Transformer) on the COCO test images. Zoom in for  better view.}
\label{fig:res2}
\end{figure}
\begin{figure}[ht]
\begin{center}
\includegraphics[height=0.9\textheight]{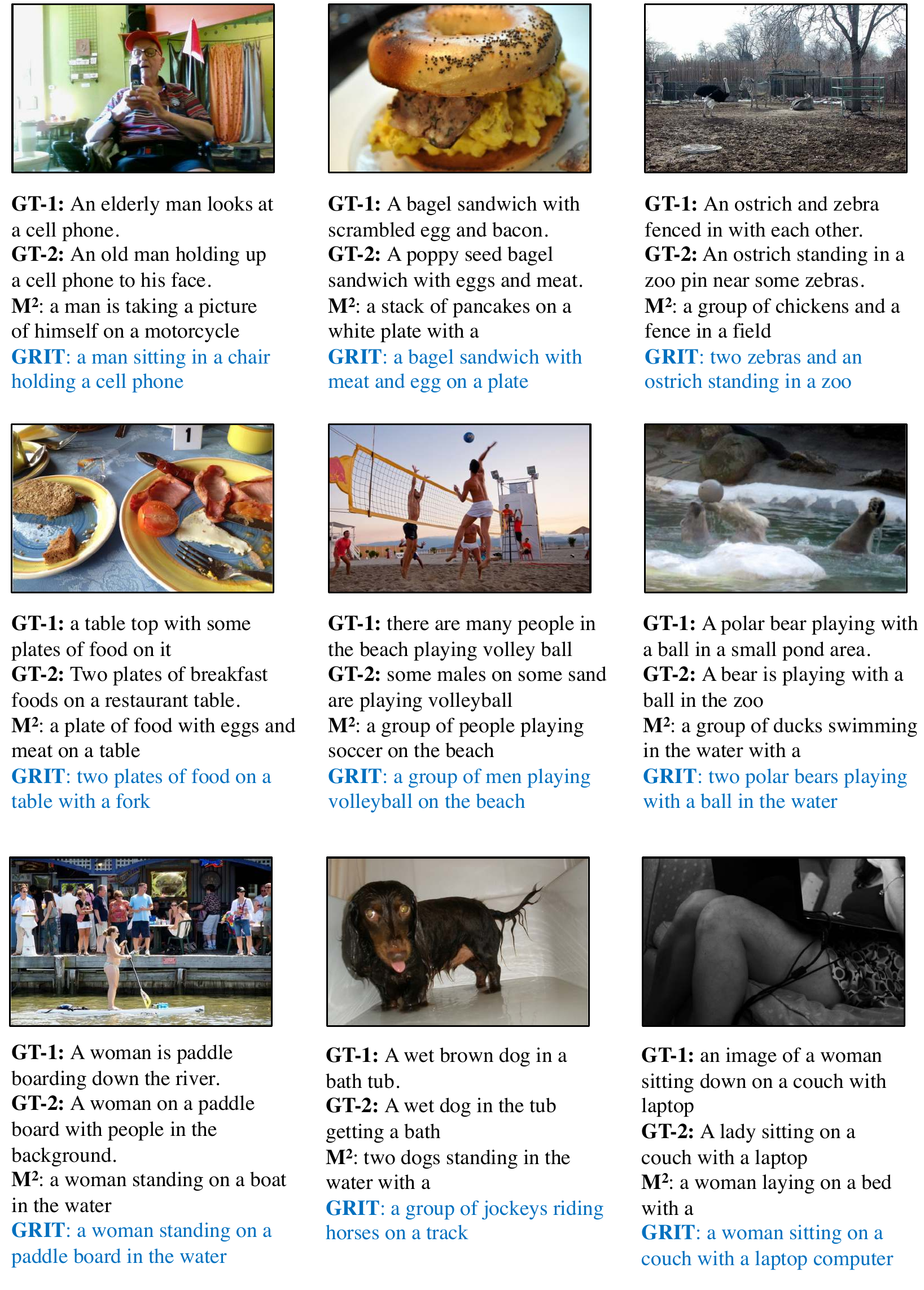}
\end{center}
   \caption{Qualitative examples from our method (GRIT) and a region-based method ($\mathcal{M}^2$ Transformer) on the COCO test images. Zoom in for  better view.}
\label{fig:res3}
\end{figure}
\begin{figure}[ht]
\begin{center}
\includegraphics[height=0.9\textheight]{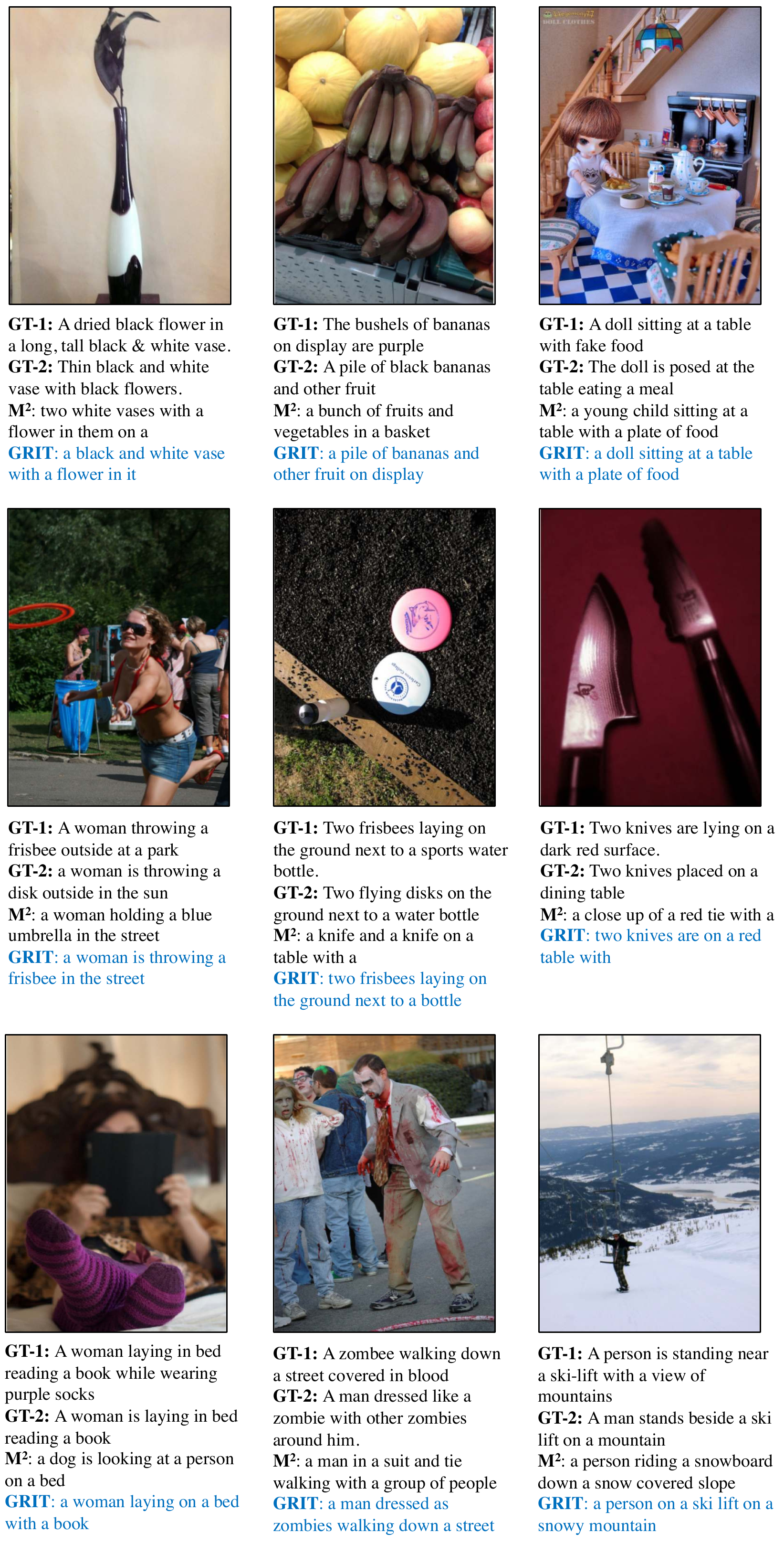}
\end{center}
   \caption{Qualitative examples from our method (GRIT) and a region-based method ($\mathcal{M}^2$ Transformer) on the COCO test images. Zoom in for  better view.}
\label{fig:res4}
\end{figure}

\subsection{Qualitative Results} 
Figure \ref{fig:res1}, \ref{fig:res2}, \ref{fig:res3}, and \ref{fig:res4} show some examples of the captions generated by our proposed method (GRIT) and another region-based method ($\mathcal{M}^2$ Transformer) given the same input images from the COCO test split. 
It is observed that the generated captions from GRIT are qualitatively better than those generated by the baseline method in terms of detecting and counting objects as well as describing their relationships in the given images. The inaccuracy of the captions generated by the baseline method might be due to the drawbacks of the region features extracted by a frozen pretrained object detector which produces wrong detection and lacks of contextual information.

\section{Summary and Conclusion} 

In this chapter, we have proposed a Transformer-based architecture for image captioning named GRIT. It integrates the region features and the grid features extracted from an input image to extract richer visual information from input images. Previous SOTA methods employ a CNN-based detector to extract region features, which prevents the end-to-end training of the entire model and makes to high computational costs. Using the Swin Transformer for a backbone extracting the initial visual feature, GRIT resolves these two issues by employing a DETR-based detector. Furthermore, GRIT obtains grid features by updating the feature from the same backbone using a self-attention Transformer, aiming to extract richer context information complementing the region feature. These two features are fed to the caption generator equipped with a unique cross-attention mechanism, which computes and applies attention from the dual features on the generated caption sentence. The integration of all these components led to significant performance improvement. The experimental results validated our approach, showing that GRIT outperforms all published methods by a large margin in inference accuracy and speed.

\cleardoublepage
\chapter{
LTMI: Lightweight Transformer for Many Inputs in Visual Dialog
}
\label{chapter:ch4}

\newcommand{\NDCG}{NDCG$\uparrow$}
\newcommand{\MRR}{MRR$\uparrow$}
\newcommand{\ROne}{R@1$\uparrow$}
\newcommand{\RFive}{R@5$\uparrow$}
\newcommand{\RTen}{R@10$\uparrow$}
\newcommand{\Mean}{Mean$\downarrow$}

\section{Introduction} \label{introduction}

The preceding chapter demonstrated that integrating dual visual features, combining the contextual awareness of grid-based representations with the object-level precision of region-based ones, yields substantially better scene understanding, as measured by image captioning performance.
Image captioning, however, is inherently a one-directional task: the agent produces a single description of an image and the exchange ends there.
In many practical applications, such as assisting visually impaired users, supporting human-robot collaboration, or enabling conversational search over image collections, a more useful form of intelligence involves the ability to engage in ongoing, multi-round conversations about visual content, answering follow-up questions, resolving ambiguous references through dialog history, and updating its interpretation as the conversation evolves.
This richer interaction regime is the focus of the present chapter.
Transitioning from image captioning to visual dialog introduces a qualitatively different modeling challenge: the agent must now reason jointly over the image, the current question, and a potentially long sequence of prior question-answer pairs, capturing all the relevant interactions among these multiple heterogeneous inputs simultaneously.
Unlike the previous chapter, where the central question was \textit{what to extract} from a single image, the central question here is \textit{how to model} the complex dependencies among many inputs, efficiently and without privileging any particular ordering or interaction path.

To reason over this setting concretely, we follow \cite{Schwartz2019FactorGA} and use the term {\em utility} to refer to each of the heterogeneous inputs—the image, the current question, and the individual question-answer pairs in the dialog history—since the term {\em modality} is inconvenient to distinguish between the question and the dialog history.
For example, to answer the question {\it `What color are they?'}, the agent must draw on the dialog history to resolve the reference of {\it `they'} and on the image to retrieve the color, illustrating why all pairwise interactions among utilities matter.

Existing studies approach this by chaining attention along a fixed, hypothesized path, such as ``question $\rightarrow$ history $\rightarrow$ image'' \cite{Kang2019DualAN,lu2017best} or ``question $\rightarrow$ image $\rightarrow$ history $\rightarrow$ question'' \cite{Gan2019MultistepRV,wu2018you}.
These predetermined paths leave out interactions that may be crucial for a given question.
A more principled approach \cite{Schwartz2019FactorGA} captures all possible interactions via a factor graph, but constructing the graph is computationally inefficient, and this inefficiency becomes a bottleneck as the dialog history grows long.

The Transformer \cite{vaswani2017attention} has emerged as the dominant architecture for natural language processing, with pre-trained variants such as BERT \cite{devlin2018bert} achieving strong results across many tasks, and extensions to bi-modal vision-language problems have also shown promise \cite{chen2019uniter,gao2019dynamic,li2019visualbert,lu2019vilbert,yu2019deep}.
It is therefore natural to ask whether the Transformer can be extended to handle all pairwise utility interactions.
The difficulty is one of scale: even in the simplest bi-modal case with utilities $X$ and $Y$, there are four attention patterns ($X\rightarrow Y$, $Y\rightarrow X$, $X\rightarrow X$, $Y\rightarrow Y$), each requiring an independent block.
Extending this naively to many utilities makes the total parameter count grow with the square of the number of utilities, and it is not clear how to aggregate the outputs of all the resulting blocks.

To cope with this, we propose a neural architecture named {\em Lightweight Transformer for Many Inputs} (LTMI) that can deal with all the interactions between many utilities. While it has a block structure similar to the Transformer and shares the core design of attention computation, it differs in the following two aspects. One is the difference in the implementation of multi-head attention. Multi-head attention in the Transformer projects the input feature space linearly to multiple lower-dimensional spaces, enabling to handle multiple attention maps, where the linear mappings are represented with learnable parameters. In the proposed model, we instead split the input feature space to subspaces mechanically according to its indexes, removing all the learnable parameters from the attention computation. 

The other difference from the Transformer is that LTMI is designed to receive multiple utilities and compute all the interactions to one utility from all the others including itself. This yields the same number of attended features as the input utilities, which are then concatenated in the direction of the feature space dimensions and then linearly projected back to the original feature space. We treat the parameters of the last linear projection 
as only learnable parameters in LTMI.
This design makes it possible to retain sufficient representational power with a much fewer number of parameters, as compared with a natural extension of the Transformer block to many utilities.
By using the same number of blocks in parallel as the number of utilities, we can deal with all the interactions between the utilities; see Fig.~\ref{fig:simplesymbol} for example.
Assuming three utilities and the feature space dimensionality of $512$, a layer consisting of
LTMI has 2.38M parameters, whereas its counterpart based on  naive Transformer extension has 28.4M parameters.

\section{Related Work}
\subsection{Attention Mechanisms for Vision-Language Tasks}

Attention mechanisms are currently indispensable to build neural architectures for vision-language tasks, such as VQA \cite{chen2015abc,ilievski2016focused,kim2018bilinear,lu2016hierarchical,nguyen2018improved,yang2016stacked,yu2017multi,yu2018beyond} and visual grounding \cite{deng2018visual,yu2018mattnet,zhuang2018parallel}, etc. Inspired by the recent success of the Transformer for language tasks \cite{devlin2018bert,vaswani2017attention}, several studies have proposed its extensions to bi-modal vision-language tasks \cite{chen2019uniter,gao2019dynamic,li2019visualbert,lu2019vilbert,tan2019lxmert,yu2019deep}. Specifically, for VQA, it is proposed to use intra-modal and inter-modal attention blocks and stack them alternately to fuse question and image features \cite{gao2019dynamic}; it is also proposed to use a cascade of modular co-attention layers that compute the self-attention and guided-attention of question and image features \cite{yu2019deep}. The method of pre-training a Transformer model used in BERT \cite{devlin2018bert} is employed along with Transformer extension to bi-modal tasks for several vision-language tasks \cite{chen2019uniter,li2019visualbert,lu2019vilbert}. They first pre-train the models on external datasets, such as COCO Captions \cite{chen2015microsoft} or Conceptual Captions dataset \cite{sharma2018conceptual}, and then fine-tune them on several target tasks.

\subsection{Visual Dialog}
The task of visual dialog has recently been proposed by two groups of researchers concurrently \cite{das2017visual,de2017guesswhat}. De Vries et al. introduced the GuessWhat?! dataset, which is built upon goal-oriented dialogs held by two agents to identify unknown objects in an image through a set of yes/no questions \cite{de2017guesswhat}. Das et al. released the VisDial dataset, which is built upon dialogs consisting of pairs of a question and an answer about an image that are provided in the form of natural language texts  \cite{das2017visual}. Kottur et al. recently introduced CLEVR-Dialog as the diagnostic dataset for visual dialog \cite{kottur2019clevr}.

Most of the existing approaches employ an encoder-decoder architecture \cite{sutskever2014sequence}. They can be categorized into the following three groups by the design of the encoder: i) fusion-based methods, e.g., LF \cite{das2017visual} and HRE \cite{das2017visual}, which fuses the inputs by their concatenation followed by the application of a feed-forward or recurrent network, and Synergistic \cite{Guo_2019_CVPR}, which fuses the inputs at multiple stages;
ii) attention-based methods that compute attended features of the input image, question, and history utilities, e.g., MN \cite{das2017visual}, CoAtt \cite{wu2018you}, HCIAE \cite{lu2017best}, Synergistic \cite{Guo_2019_CVPR}, ReDAN \cite{Gan2019MultistepRV}, FGA \cite{Schwartz2019FactorGA}, and CDF \cite{kim2020modality}; ReDAN compute the attention over several reasoning steps, FGA models all the interactions over many utilities via a factor graph; 
iii) methods that attempt to resolve visual co-reference, e.g., RvA \cite{Niu_2019_CVPR} and  CorefNMN \cite{kottur2018visual}, which use neural modules to form an attention mechanism, DAN \cite{Kang2019DualAN}, which employs a network having two attention modules, and AMEM \cite{seo2017visual}, which utilizes a  memory mechanism for attention. As for the decoder, there are two designs: i) discriminative decoders that rank the candidate answers using the cross-entropy loss \cite{das2017visual} or the n-pair loss \cite{lu2017best}; and ii) generative decoders that yield an answer by using a MLE loss \cite{das2017visual}, weighted likelihood estimation \cite{Zhang2019}, or a combination with adversarial learning \cite{lu2017best,wu2018you}, which trains a discriminator on both positive and negative answers, then transferring it to the generator with auxiliary adversarial learning.

Other approaches include GNN \cite{Zheng2019ReasoningVD}, which models relations in a dialog by an unknown graph structure;
the employment of reinforcement learning \cite{chattopadhyay2017evaluating,das2017learning}; 
and HACAN \cite{Yang2019MakingHM} which adopts policy gradient to learn the impact of history by intentionally imposing the wrong answer into dialog history. 
In \cite{wang2020vd,murahari2019large}, pre-trained vision-language models 
are adopted, which consist of many Transformer blocks with hundreds of millions parameters, leading to some performance gain.
Qi et al. \cite{qi2020two} present model-agnostic principles for visual dialog to maximize performance. 

\section{Lightweight Transformer for Many Utilities}

\subsection{Attention Mechanism of Transformer}
As mentioned earlier, the Transformer has been applied to several bi-modal vision-language tasks, yielding promising results.
The Transformer computes and uses attention from three types of inputs, $Q$ (query), $K$ (key), and $V$ (value). Its computation is given by
\begin{equation}
    \mathcal{A}(Q,K,V)=\mbox{softmax}\left(\frac{Q K^\top}{\sqrt{d}} \right) V,
    \label{eqn:trm_attn}
\end{equation}
where $Q$, $K$, and $V$ are all collection of features, each of which is represented by a $d$-dimensional vector. To be specific, $Q=[q_1,\ldots,q_M]^\top\in \mathbb{R}^{M\times d}$ is a collection of $M$ features; similarly, $K$ and $V$ are each a collection of $N$ features, i.e.,  $K, V\in \mathbb{R}^{N\times d}$. In Eq.(\ref{eqn:trm_attn}), $V$ is attended with the weights computed from the similarity between $Q$ and $K$.

The above computation is usually multi-plexed in the way called multi-head attention. It enables to use a number of attention distributions in parallel, aiming at an increase in representational power. The outputs of $H$ `heads' are concatenated, followed by linear transformation with learnable weights $W^O\in\mathbb{R}^{d\times d}$ as
\begin{equation}
    \mathcal{A}^{\mathrm{M}}(Q,K,V)=\begin{bmatrix}
    \mathrm{head}_1,\cdots,\mathrm{head}_H
    \end{bmatrix}W^O.
\end{equation}
Each head is computed as follows:
\begin{equation}
    \mathrm{head}_h = \mathcal{A}(QW_h^Q, KW_h^K, VW_h^V), \;\;h=1,\ldots,H, 
\end{equation}
where $W_h^Q$, $W_h^K$, $W_h^V \in \mathbb{R}^{d\times d_H}$ each are learnable weights inducing a linear projection from the feature space of $d$-dimensions to a lower space of $d_H(=d/H)$-dimensions. Thus, one attentional block $\mathcal{A}^{\mathrm{M}}(Q,K,V)$ has the following learnable weights:
\begin{equation}
    (W_1^Q, W_1^K, W_1^V),\cdots,(W_H^Q, W_H^K, W_H^V)\;\; \mbox{and} \;\; W^O.
    \label{eqn:ma_weights}
\end{equation}

\begin{figure}[t]
\centering
\includegraphics[width=0.85\columnwidth]{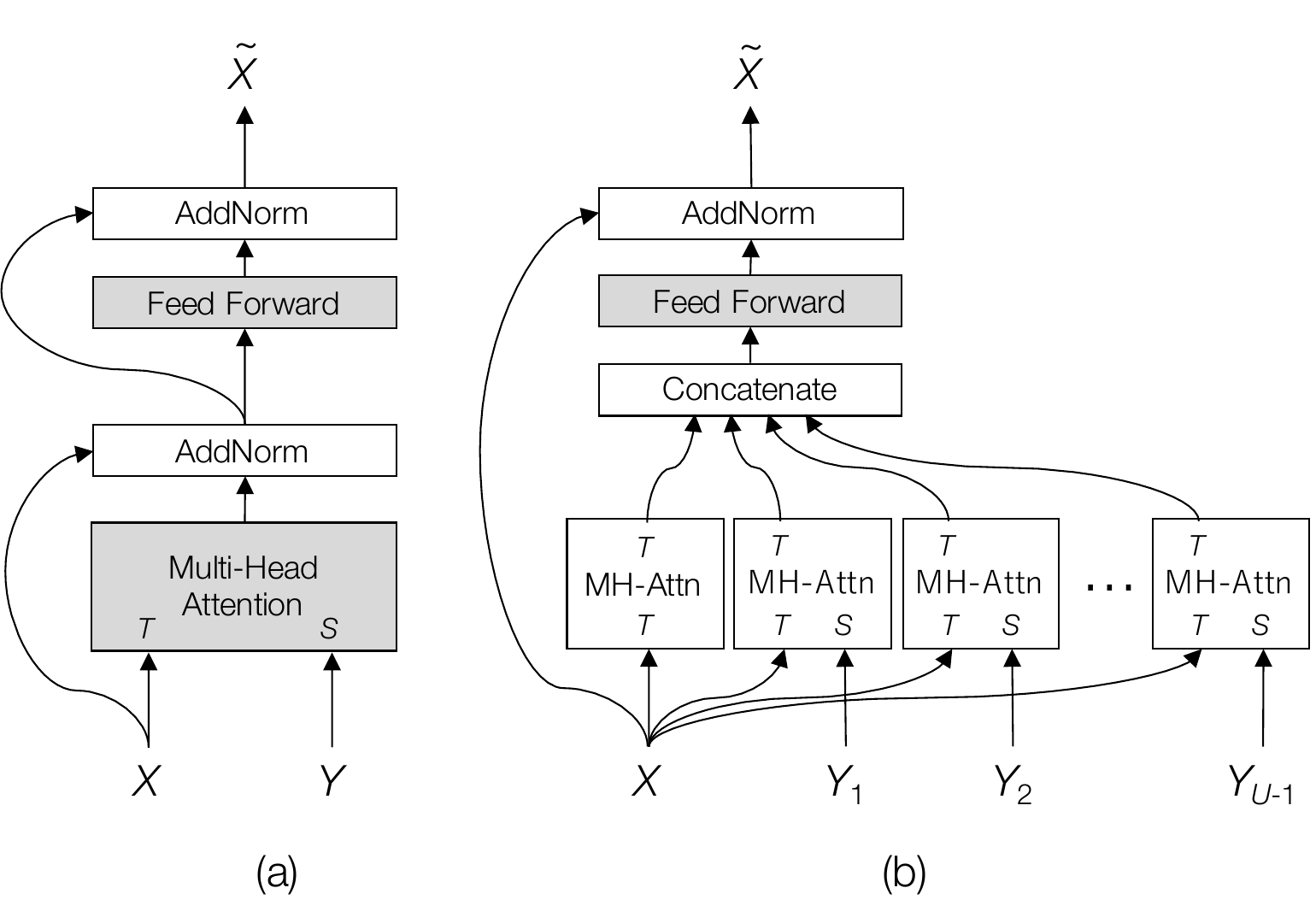}
\vspace*{-0.1in}
\caption{(a) Source-to-target attention for bi-modal problems implemented by the standard Transformer block; the source $Y$ is attended by weights computed from the similarity between the target $X$ and $Y$. (b) The proposed block that can deal with many utilities; the source features $\{Y_1,\ldots,Y_{U-1}\}$ are attended by weights computed between them and the target $X$. Shaded boxes have learnable weights }
\label{fig:comparison_models}
\end{figure}

\subsection{Application to Bi-Modal Tasks}
While $Q$, $K$, and $V$ in NLP tasks are of the same modality (i.e., language), the above mechanism has been extended to bi-modality and applied to vision-language tasks in recent studies \cite{chen2019uniter,gao2019dynamic,li2019visualbert,lu2019vilbert,tan2019lxmert,yu2019deep}. 
They follow the original idea of the Transformer, considering attention from source features $Y$ to target features $X$ as
\begin{equation}\label{eqn:basic_attn}
   \mathcal{A}_Y(X) = \mathcal{A}^{\mathrm{M}}(X, Y, Y).
\end{equation}
In MCAN \cite{yu2019deep}, language feature is treated as the source and visual feature is as the target. 
In \cite{li2019visualbert} and others \cite{chen2019uniter,gao2019dynamic,lu2019vilbert,tan2019lxmert}, co-attention, i.e., attention in the both directions, is considered. Self-attention, i.e., the attention from features to themselves, is given as a special case by 
\begin{equation}
    \mathcal{A}_X(X) = \mathcal{A}^{\mathrm{M}}(X, X, X).
    \label{eqn:a_xx_attn}
\end{equation}
In the above studies, the Transformer block with the source-to-target attention and that with the self-attention are independently treated and are stacked, e.g., alternately or sequentially.

\subsection{Lightweight Transformer for Many Inputs} 

Now suppose we wish to extend the above attention mechanism to a greater number of utilities\footnote{As we stated in Introduction, we use the term {\em utility} here to mean a collection of features.};
we denote the number by $U$. If we consider every possible source-target pairs, 
there are $U(U-1)$ cases in total, as there are $U$ targets, for each of which $U-1$ sources exist. Then we need to consider attention computation $\mathcal{A}_{Y}(X)$ over $U-1$ sources $Y$'s for each target $X$.
Thus, the straightforward extension of the above attention mechanism to $U$ utilities needs $U(U-1)$ times the number of parameters listed in Eq.(\ref{eqn:ma_weights}). If we stack the blocks, the total number of parameters further increases proportionally.

To cope with this, we remove all the weights from Eq.(\ref{eqn:basic_attn}). To be specific, for each head $h(=1,\ldots,H)$, we choose and freeze $(W_h^Q, W_h^K, W_h^V)$ as
\begin{equation}
W^Q_h=W^K_h=W^V_h = [\underbrace{O_{d_H},\cdots,O_{d_H}}_{(h-1)d_H},I_{d_H},\underbrace{O_{d_H},\cdots,O_{d_H}}_{(H-h)d_H}]^\top,
    \label{eqn:freeze_weights}
\end{equation}
where $O_{d_H}$ is a $d_H\times d_H$ zero matrix and $I_{d_H}$ is a $d_H\times d_H$ identity matrix. 
In short, the subspace for each head is determined to be one of $H$ subspaces obtained by splitting the $d$-dimensional feature space with its axis indexes. 
Besides, we set $W^O=I$, which is the linear mapping applied to the concatenation of the heads' outputs. 
Let $\bar{\mathcal{A}}_Y(X)$ denote this simplified attention mechanism.

Now let the utilities be denoted by $\{X,Y_1,\ldots,Y_{U-1}\}$, where $X\in \mathbb{R}^{M\times d}$ is the chosen target and others  $Y_i\in \mathbb{R}^{N_i\times d}$ are the sources. Then, we compute all the source-to-target attention as $\bar{\mathcal{A}}_{Y_1}(X),\cdots, \bar{\mathcal{A}}_{Y_{U-1}}(X)$. In the standard Transformer block (or rigorously its natural extensions to bi-modal problems), the attended features are simply added to the target as $X + \mathcal{A}_Y(X)$, followed by normalization and subsequent computations. To recover some of the loss in representational power due to the simplification yielding $\bar{\mathcal{A}}_Y(X)$, we propose a different approach to aggregate $\bar{\mathcal{A}}_{Y_1}(X),\cdots, \bar{\mathcal{A}}_{Y_{U-1}}(X)$ and $X$. 
Specifically, we concatenate all the source-to-target attention plus the self-attention $\bar{\mathcal{A}}_{X}(X)$ from $X$ to $X$ as
\begin{equation}
    X_{\mathrm{concat}} = [\bar{\mathcal{A}}_{X}(X), \bar{\mathcal{A}}_{Y_1}(X), \cdots,\bar{\mathcal{A}}_{Y_{U-1}}(X)],
\end{equation}
where $X_{\mathrm{concat}}\in\mathbb{R}^{M\times Ud}$. We then apply linear transformation to it given by 
$W\in \mathbb{R}^{Ud\times d}$ and $b\in \mathbb{R}^d$ with a single fully-connected layer, followed by the addition of the original $X$ and layer normalization as
\begin{equation}
    \tilde{X} = \mathrm{LayerNorm}( \mathrm{ReLU}(X_{\mathrm{concat}} W 
    +\mathbf{1}_{M} \cdot b^\top) + X),
\end{equation}
where $\mathbf{1}_M$ is  $M$-vector with all ones. With this method, we aim at recovery of representational power as well as the effective aggregation of information from all the utilities. 

\begin{figure}[t]
\centering
\includegraphics[width=0.85\columnwidth]{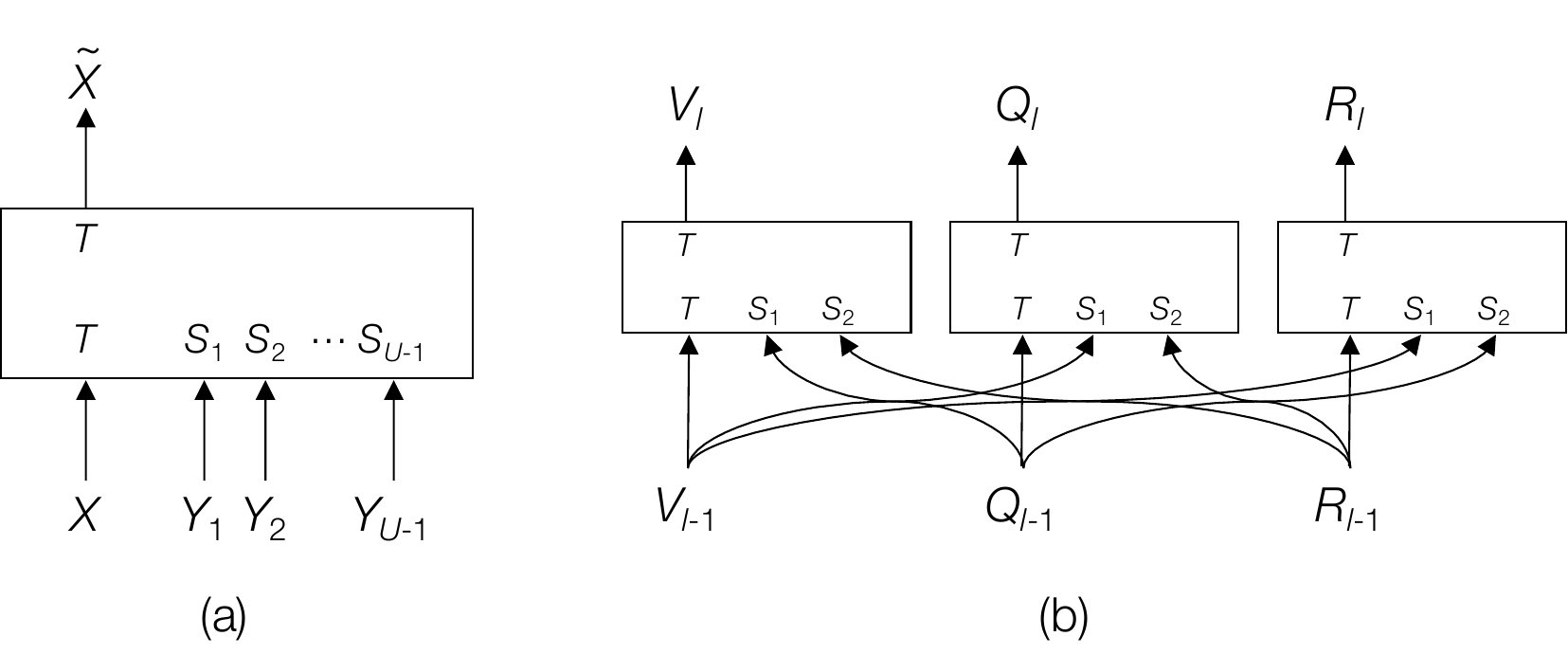} 
\vspace*{-0.1in}
\caption{(a) Simplified symbol of the proposed block shown in Fig.~\ref{fig:comparison_models}(b). (b) Its application to Visual Dialog }
\label{fig:simplesymbol}
\end{figure}

\paragraph{Memory features} When computing $\bar{\mathcal{A}}_{Y}(X)$, we perform the following form of computation
\[
\mathcal{A}(Q,K,V)=\mathrm{softmax}\left(
\frac{Q K^\top}{\sqrt{d}}
\right)V,
\]
where we compute a matrix product $Q K^\top$ as above. In the computation of $\bar{\mathcal{A}}_{X}(Y)$, we need another matrix product, but it is merely the transposed matrix $K Q^\top$ due to the symmetry between $X$ and $Y$. 
For the computational efficiency, we perform computation of $\bar{\mathcal{A}}_{Y}(X)$ and $\bar{\mathcal{A}}_{X}(Y)$ simultaneously; see  $\texttt{MultiHeadAttention}(X, Y)$ in our code. 
Further, following \cite{nguyen2018improved}, we also pad $X$ and $Y$ with two $d$-dimensional vectors that are randomly initialized with He normal initialization.  This implements ``no-where-to-attend'' features as memory features in the computation of $\bar{\mathcal{A}}_{Y}(X)$ and $\bar{\mathcal{A}}_{X}(Y)$.
\subsection{Interactions between All Utilities}
We have designed a basic block as shown in Fig.~\ref{fig:comparison_models}(b) that deals with attention from many sources to a single target. We wish to consider all possible interactions between all the utilities, not a single utility being the only target. To do this, we use $U$ basic blocks to consider all the source-to-target attention. Using the basic block as a building block, we show how an architecture is designed for visual dialog having three utilities, visual features $V$, question features $Q$, and dialog history features $R$, in Fig.~\ref{fig:simplesymbol}(b).

\section{Implementation Details for Visual Dialog}

\subsection{Problem Definition}
The problem of Visual Dialog is stated as follows. An agent is given the image of a scene and a dialog history containing $T$ entities, which consists of a caption and question-answer pairs at $T-1$ rounds. Then, the agent is further given a new question at round $T$ along with 100 candidate answers for it and requested to answer the question by choosing one or scoring each of the candidate answers.

\subsection{Representation of Utilities}
We first extract features from an input image, a dialog history, and a new question at round $T$ to obtain their representations. For this, we follow the standard method employed in many recent studies. For the image utility, we use the bottom-up mechanism \cite{anderson2018bottom}, which extracts region-level image features using the Faster R-CNN \cite{ren2015faster} pre-trained on the Visual Genome dataset \cite{krishna2017visual}. For each region (i.e., a bounding box = an object), we combine its CNN feature and geometry to get a $d$-dimensional vector $v_i$ ($i=1,\ldots,K$), where $K$ is the predefined number of regions. We then define $V = [v_1, v_2, \cdots, v_K]^\top \in \mathbb{R}^{K \times d}$. For the question utility, after embedding each word using an embedding layer initialized by pre-trained GloVe vectors, we use two-layer Bi-LSTM to transform them to $q_i$ $(i=1,\ldots,N)$, where $N$ is the number of words in the question. We optionally use the positional embedding widely used in NLP studies. We examine its effects in an ablation test. We then define $Q = [q_1,\ldots,q_N]^\top \in \mathbb{R}^{N \times d}$. For the dialog history utility, we choose to represent it as a single utility here. Thus, each of its entities represents the initial caption or the question-answer pair at one round. As with the question utility, we use the same embedding layer and a two-layer Bi-LSTM together with the positional embeddings for the order of dialog rounds to encode them with a slight difference in formation of an entity vector $r_i$ ($i=1,\ldots,T)$, where $T$ is the number of Q\&A plus the caption. We then define $R = [r_1,\ldots,r_T]^\top \in \mathbb{R}^{T \times d}$. 

\paragraph{Image Utility}

The image utility is represented by the standard method employed in many recent studies. It is based on the bottom-up mechanism \cite{anderson2018bottom}, which extracts region-level image features using the Faster R-CNN pre-trained on the Visual Genome dataset \cite{krishna2017visual}. For each input image, we select the top $K$ objects, and represent each of them by a visual feature $v^r_i\in\mathbb{R}^{2048}$ and a bounding box expressed by $(x_{i,1},x_{i,2})$ and $(x_{i,3},x_{i,4})$ (the coordinates of the upper-left and lower-right corners.) 

The feature vector $v^r_i$ is then converted into another vector $v^f_i\in\mathbb{R}^d$ as follows. We introduce the following notation to express a single FC layer with ReLU, to which dropout regularization is applied:
\begin{equation}
    \underset{k \rightarrow d}{\mathrm{MLP}}(x) \equiv \mathrm{Dropout}(\mathrm{ReLU}(W^{\top}x + b)),
\end{equation}
where $x \in \mathbb{R}^{k}$ is the input and $W \in \mathbb{R}^{k \times d}$ and $b\in \mathbb{R}^d$ are the weights and biases. Then, $v^f_i$ is obtained by \begin{equation}
    v^f_i = \mathrm{LayerNorm}(\mathop{\mathrm{MLP}}_{2048 \rightarrow d}(v^r_i)), 
\end{equation}
where LayerNorm is the layer normalization \cite{ba2016layer} applied to the output. 

The bounding box geometry is converted into $v^b_i\in \mathbb{R}^d$ in the following way. First, the image is resized to $600\times 600$ pixels and the bounding box geometry is transformed accordingly. Then, representing each of the four coordinates by a one-hot vector of size 600, we convert them into the embedding vectors  $\hat{x}_{i,1},\ldots,\hat{x}_{i,4}(\in\mathbb{R}^{d})$ using four different embedding layers. Then, we obtain $v^b_i$ as below
\begin{equation}
    v^b_i = \sum_{j=1}^{4}\mathrm{LayerNorm}(
    \mathop{\mathrm{MLP}}_{d\rightarrow d}(\hat{x}_{i,j})).
\end{equation}

Finally, $v^f_i$ encoding the visual feature and $v^b_i$ encoding the spatial feature are aggregated by adding and normalizing as
\begin{equation}\label{eq:agg_visdial}
    v_i = \mathrm{LayerNorm}(v^f_i + v^b_i). 
\end{equation}

The resulting $v_i$'s for the $K$ objects ($i=1,\ldots,K$) comprise a matrix $V = [v_1, v_2, \cdots, v_K]^\top \in \mathbb{R}^{K \times d}$, which gives the representation of the visual utility.

\paragraph{Optional Image Feature Enrichment.} 
In the experiment of comparing ensembles on the test split of Visdial v1.0, we enrich the image features for further improvement. 
To be specific, for each object, we also obtain a class label with highest probability (e.g. {\em`cat', `hair', and `car'}) and the top 20 attributes for each class label (e.g., `curly', `blond', `long', and so on, for the label `hair'). These can be extracted from the Faster R-CNN along with the above CNN features and bounding box geometry. We incorporate these into the image utility representation in the following way.

The class label for the $i$-th object is first encoded into an embedding vector $e_i^c \in \mathbb{R}^{300}$ using the same embedding layer as the question. Then, we convert $e_i^c$ into a $d$-dimensional vector $v_i^c$ by
\begin{equation}
    v_i^c = \mathrm{LayerNorm}(\mathop{\mathrm{MLP}}_{300\rightarrow d}(e_i^c)).
\end{equation}
Similarly, for the top 20 attributes of each object $i$, we encode them into embedding vectors of size $300$, i.e. $e_{i,1}^a, \ldots, e_{i,20}^a$, and then convert them further into $v^a_i \in \mathbb{R}^{d}$ as 
\begin{equation}
    v^a_i = \sum_{j=1}^{20}\mathrm{LayerNorm}(
    \mathop{\mathrm{MLP}}_{300\rightarrow d}(e_{i,j}^a)w_{i,j}^a,
\end{equation}
where $w_{i,j}^a$ is the confidence score extracted from the Faster R-CNN for attribute $j$ of the $i$-th object. Then, the visual feature $v_i^f$, the spatial feature $v_i^b$, the class feature $v_i^c$, and the attribute feature $v_i^a$ are aggregated by their addition followed by normalization as
\begin{equation}
        v_i = \mathrm{LayerNorm}(v^f_i + v^b_i + v_i^c + v_i^a). 
\end{equation}
We then use these vectors to form the matrix $V$ instead of Eq.(\ref{eq:agg_visdial}).

\paragraph{Question Utility} \label{ques_utility}

The question utility is also obtained by the standard method but with one exception, the employment of positional embedding used in NLP studies. Note that we examine its effects in an ablation test shown in the main paper. A given question sentence is first fit into a sequence of $N$ words; zero-padding is applied if necessary.
Each word $w_i$ ($i=1,\ldots,N)$ is embedded into a vector $e_i$ of a fixed size using an embedding layer initialized with pre-trained GloVe vectors \cite{pennington2014glove}. They are then inputted into two-layer Bi-LSTM, obtaining two d-dimensional vectors $\overrightarrow{h_{i}}$ and $\overleftarrow{h_{i}}$ as their higher-layer hidden state: 
\begin{equation}
      \begin{aligned}
    \overrightarrow{h_{i}} & =  \mathrm{LSTM}(e_i, \overrightarrow{h_{i-1}}),\\
        \overleftarrow{h_{i}} & =  \mathrm{LSTM}(e_i, \overleftarrow{h_{i+1}}).
    \label{eq:ques_lstm}
  \end{aligned}
\end{equation}
Their concatenation, $h_{i} = [ \overrightarrow{h_{i}}^\top, \overleftarrow{h_{i}}^\top ]^\top$, is then projected back to a $d$-dimensional space using a linear transformation, yielding a vector $q^f_i$. Positional embedding $q^p_i$ from the paper \cite{vaswani2017attention} is added to get the final representation $q_i \in \mathbb{R}^{d}$ of $w_i$ as 
\begin{equation}
    q_{i} = \mathrm{LayerNorm}(q^f_i + q^p_i).
    \label{eq:ques_layernorm}
\end{equation}
The representation of the question utility is given as $Q = [q_1,\ldots,q_N]^\top \in \mathbb{R}^{N \times d}$.

\paragraph{Dialog History Utility} \label{hist_utility}

In this study, we choose to represent the dialog history as a single utility. Each of its entities represents the question-answer pair at one round. As with previous studies, the caption is treated as the first round of $2N$-word which is padded or truncated if necessary. For each round $t > 1$, the word sequences of the question and the answer at the round is concatenated into $2N$-word sequence with zero padding if necessary.
As with the question utility, after embedding each word into a GloVe vector, the resulting sequence of $2N$ embedded vectors is inputted to two-layer Bi-LSTM, from which only their last (higher-layer) hidden states are extracted to construct $2d$-dimensional vector $[\overrightarrow{h_{0}}^\top, \overleftarrow{h_{2N}}^\top]^\top$. We then project it with a linear transform to a $d$-dimensional space, yielding $r^f_t\in \mathbb{R}^d$. For the linear projection, we use different learnable weights from the question utility. As in Eq.(\ref{eq:ques_layernorm}), we add positional embedding, which represents the order of rounds, and then apply layer normalization, yielding a feature vector of the round $t$ question-answer pair. The history utility is then given by $R = [r_1,\ldots,r_T]^\top \in \mathbb{R}^{T \times d}$.

\subsection{Overall Network Design}
\begin{figure}[t]
\centering
\includegraphics[width=1.0\textwidth]{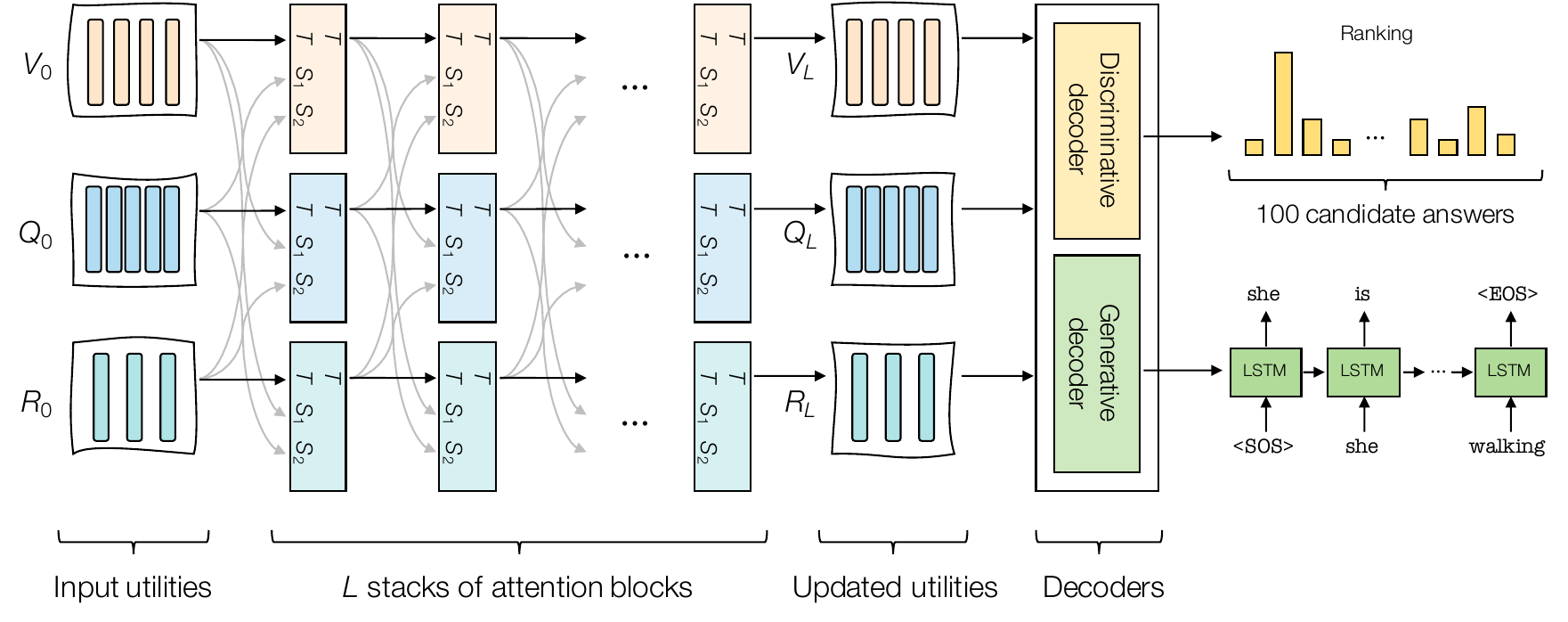}
\caption{The entire network built upon the proposed LTMI for Visual Dialog}
\label{fig:model_overview}
\vskip -0.1in
\end{figure}

Figure \ref{fig:model_overview} shows the entire network. It consists of an encoder and a decoder. The encoder consists of $L$ stacks of the proposed attention blocks; a single stack has $U$ blocks in parallel, as shown in Fig.~\ref{fig:simplesymbol}(b).
We set $V_0 = V$, $Q_0 = Q$, and $R_0 = R$ as the inputs of the first stack. After the $l$-th stack, the representations of the image, question, and dialog history utilities are updated as $V_l$, $Q_l$, and $R_l$, respectively.
In the experiments, we apply dropout with the rate of $0.1$ to the linear layer inside every block. There is a decoder(s) on top of the encoder. We consider a discriminative decoder and a  generative decoder, as in previous studies. Their design is explained below.

\subsection{Design of Decoders}
Decoders receive the updated utility representations, $V_L$, $Q_L$, and $R_L$ at their inputs. We convert them independently into $d$-dimensional vectors $c_V$, $c_Q$, and $c_R$, respectively. This conversion is performed by a simple self-attention computation. We take $c_V$ as an example here. First, attention weights over the entities of $V_L$ are computed by a two-layer network as 
\begin{equation}\label{eqn:a_v_weights}
    a_V = \mathrm{softmax}(\mathrm{ReLU}(V_LW_1 + \mathbf{1}_Kb_1^\top)W_2 + \mathbf{1}_Kb_2),
\end{equation}
where $W_1 \in \mathbb{R}^{d\times d}$, $W_2 \in \mathbb{R}^{d\times1}$, $b_1\in \mathbb{R}^d$, $b_2 \in \mathbb{R}^1$, and $\mathbf{1}_K$ is $K$-vector with all ones.
Then, $c_V$ is given by 
\begin{equation} \label{eqn:self_attn}
     c_V = \sum_{i = 1}^{K}v_{L,i}^\top a_{V,i},
\end{equation}
where $v_{L,i}$ is the $i$-th row vector of $V_L$ and $a_{V,i}$ is the $i$-th attention weight (a scalar). The others, i.e., $c_Q$ and $c_R$, can be obtained similarly. 

These vectors are integrated and used by the decoders. In our implementation for visual dialog, we found that $c_R$ does not contribute to better results; thus we use only $c_V$ and $c_Q$. Note that this does not mean the dialog utility $R$ is not necessary; it is interacted with other utilities inside the attention computation, contributing to the final prediction. The two $d$-vectors $c_V$ and $c_Q$ are concatenated as $[c_V^\top, c_Q^\top]^\top$, and this is projected to $d$-dimensional space, yielding a context vector $c\in \mathbb{R}^d$. 

We design the discriminative and generative decoders following the previous studies. Receiving $c$ and the candidate answers, the two decoders compute the score of each candidate answer in different ways.

\paragraph{Discriminative Decoder}
A discriminative decoder outputs the likelihood score for each of 100 candidate answers for the current question at round $T$ in the following way. We use a similar architecture to the one used to extract question features in Sec.~\ref{ques_utility} to convert each candidate answer (indexed by $i(=1,\ldots,100)$) to a feature vector $a_i \in \mathbb{R}^d$. Specifically, it is two-layer Bi-LSTM receiving a candidate answer at its input, on top of which there is a linear projection layer followed by layer normalization. Using the resulting vectors, the score $p_i$ for $i$-th candidate answer is computed by 
\begin{equation}
p_i=\mathrm{logsoftmax}_i(a_1^\top c,\ldots,a_{100}^\top c). 
\end{equation}
In the test phase, we sort the candidate answers using these scores. In the training phase, the cross-entropy loss $\mathcal{L}_D$ between $p=[p_1,\ldots,p_{100}]^\top$ and the ground truth label encoded by a one-hot vector $y$ is minimized:
\begin{equation}\label{eqn:disc_loss}
    \mathcal{L}_D = -\sum_{i=1}^{100}y_i p_i.
\end{equation}
When relevance scores $s=[s_1,\ldots,s_{100}]^\top$ over the answer candidates are available (called dense annotation in the VisDial dataset) rather than a single ground truth answer, we can use them by setting $y_i=s_i$ for all $i$'s and minimize the above loss. We employ dropout with rate of 0.1 for the LSTM.

\paragraph{Generative Decoder} \label{sec:gen_decoder}

Following \cite{das2017visual}, we also consider a generative decoder to score the candidate answers using the log-likelihood scores. The generative decoder consists of a two-layer LSTM  to generate an answer using the context vector $c$ as the initial hidden state. In the training phase, we predict the next token based on the current token from the ground truth answer. In details, we first append the special token ``SOS" at the beginning of the ground truth answer, then embedding all the sentence into the embedding vectors $a_{gt} = [w_0, w_1, \ldots, w_N]$ where $w_0$ is the embedding vector of ``SOS" token. The hidden state $h_n \in \mathbb{R}^{d}$ at the $n$-th timestep (extracted from the higher-layer LSTM) is computed given $w_{n-1}$ and $h_{n-1}$ as follows:
\begin{equation}
    h_n = \textrm{LSTM}(w_{n-1}, h_{n-1}),   
\end{equation}
where $h_0$ is initialized by $c$. Thus, we compute $p_n$, the log-likelihood of $n$-th word as 
\begin{equation} \label{eq:p_n}
    p_n = \mathrm{logsoftmax}_j(W_n^\top h_n  + b_n),
\end{equation}
where $W_n \in \mathbb{R}^{d \times |V|}$ and $p_n \in \mathbb{R}^{|V|}$, where $|V|$ is the vocabulary size; and $j$ is the index of $n$-th word in the vocabulary.

In the training phase, we minimize $\mathcal{L}_G$, the summation of the negative log-likelihood defined by
\begin{equation}
    \mathcal{L}_G = -\sum_{n=1}^{N}p_n.
\end{equation}
In the validation and test phase, for each candidate answer $A_{T,i}$, we compute $s_i = \sum_{n=1}^{N}p_n^{(A_{T,i})}$ where $p_n^{(A_{T,i})}$ is the log-likelihood of the $n$-th word in the candidate answer $A_{T,i}$ which is computed similarly as in Eq.(\ref{eq:p_n}). Then, the rankings of the candidate answers are derived as $\mathrm{softmax}_{i}(s_1, \ldots, s_{100})$. We employ dropout with rate of 0.1 for the LSTM.

\subsection{Multi-Task Learning}

We observe in our experiments that accuracy is improved by training the entire network using the two decoders simultaneously. This is simply done by minimizing the sum of the losses, $\mathcal{L}_D$ for the discriminative one and $\mathcal{L}_G$ for the generative one (we do not use weights on the losses): 
\begin{equation}
    \mathcal{L} = \mathcal{L}_D + \mathcal{L}_G.
\end{equation}
The increase in performance may be attributable to the synergy of learning two tasks while sharing the same encoder. Details will be given in Sec.~\ref{sec:abalation_study}.

\section{Experiments on Visual Dialog}

\subsection{Experimental Setup}

\paragraph{Dataset}~
We use the VisDial v1.0 dataset in our experiments which consists of the train 1.0 split (123,287 images), the val 1.0 split (2,064 images), and test v1.0 split (8,000 images).
Each image has a dialog composed of 10 question-answer pairs along with a caption. For each question-answer pair, 100 candidate answers are given.
The val v1.0 split and 2,000 images of the train v1.0 split are provided with dense annotations (i.e., relevance scores) for all candidate answers. Although the test v1.0 split was also densely annotated, the information about the ground truth answers and the dense annotations are not publicly available.

\paragraph{Evaluation metrics}~
From the visual dialog challenge 2018, normalized discounted cumulative gain (NDCG) has been used as the principal metric to evaluate methods on the VisDial v1.0 dataset. Unlike other classical retrieval metrics such as R@1, R@5, R@10, mean reciprocal rank (MRR), and mean rank, which are only based on a single ground truth answer, NDCG is computed based on the relevance scores of all candidate answers for each question, which can properly handle the case where each question has more than one correct answer, such as {\it `yes it is'} and {\it `yes'}; such cases do occur frequently. 

\paragraph{Other configurations}~ We employ the standard method used by many recent studies for the determination of hyper-parameters etc. For the visual features, we detect $K=100$ objects from each image. For the question and history features, we first build the vocabulary composed of 11,322 words that appear at least five times in the training split. The captions, questions, and answers are truncated or padded to 40, 20, and 20 words, respectively. Thus, $N=20$ for the question utility $Q$. $T$ for the history utilities varies depending on the number of dialogs. We use pre-trained 300-dimensional GloVe vectors \cite{pennington2014glove} to initialize the embedding layer, which is shared for all the captions, questions, and answers.

For the attention blocks, we set the dimension of the feature space to $d=512$ and the number of heads $H$ in each attention block to $4$. We mainly use models having two stacks of the proposed attention block. 
We train our models on the VisDial v0.9 and VisDial v1.0 dataset using the Adam optimizer \cite{kingma2014adam} with $5$ epochs and $15$ epochs respectively. The learning rate is warmed up from $1\times 10^{-5}$ to $1\times 10^{-3}$ in the first epoch, then halved every $2$ epochs. The batch size is set to $32$ for the both datasets.

\begin{table}[ht]
    \centering
    \caption{Hyper-paramters used in the training procedure.}
    \resizebox{0.5\columnwidth}{!}{
    \begin{tabular}{ll}
    \toprule
    Hyper-parameter          &  Value \\
    \midrule
    Warm-up learning rate   &  $1\mathrm{e}{-5}$ \\
    Warm-up factor & 0.2 \\
    Initial learning rate after the 1st epoch & $1\mathrm{e}{-3}$ \\
    $\beta_1$ in Adam & 0.9 \\
    $\beta_2$ in Adam & 0.997 \\
    $\epsilon$ in Adam & $1\mathrm{e}{-9}$  \\
    Weight decay & $1\mathrm{e}{-5}$  \\
    Number of workers & 8 \\
    Batch size & 32 \\
    \bottomrule
    \end{tabular}
    }
    \label{tab:hyperparams}
\end{table}

Table \ref{tab:hyperparams} shows the hyper-parameters used in our experiments, which are selected following the previous studies. We perform all the experiments on a GPU server that has four Tesla V100-SXM2 of 16GB memory with CUDA version 10.0 and Driver version 410.104. It has Intel(R) Xeon(R) Gold 6148 CPU @ 2.40GHz of 80 cores with the RAM of 376GB memory. We use Pytorch version 1.2 \cite{paszke2017automatic} as the deep learning framework.

\subsection{Comparison with State-of-the-art Methods}
\noindent
{\bf Compared methods}~
We compare our method with previously published methods on the VisDial v0.9 and VisDial v1.0 datasets, including LF, HRE, MN \cite{das2017visual}, LF-Att, MN-Att (with attention) \cite{das2017visual}, SAN \cite{yang2016stacked}, AMEM \cite{seo2017visual}, SF \cite{jain2018two}, HCIAE \cite{lu2017best} and Sequential CoAttention model (CoAtt) \cite{wu2018you}, Synergistic \cite{Guo_2019_CVPR}, FGA \cite{Schwartz2019FactorGA}, GNN \cite{Zheng2019ReasoningVD}, RvA \cite{Niu_2019_CVPR}, CorefNMN \cite{kottur2018visual}, DAN \cite{Kang2019DualAN}, and ReDAN \cite{Gan2019MultistepRV}, all of which were trained without using external datasets or data imposition.
Unless noted otherwise, the results of our models are obtained from the output of discriminative decoders.

\bigskip
\noindent{\bf
Results on the val v1.0 split}~
\begin{table}[t!]
\caption{Comparison of the performances of different methods on the validation set of VisDial v1.0 with discriminative and generative decoders.}
\resizebox{1.0\columnwidth}{!}{
\begin{tabular}{lrccccccccccccc}
  \toprule
  & \multicolumn{1}{c}{\multirow{2}{*}{Model}} & \multicolumn{6}{c}{Discriminative}                    & & \multicolumn{6}{c}{Generative}                         \\
  \cmidrule{3-8} \cmidrule{10-15}
  & \multicolumn{1}{c}{}                       & \NDCG          & \MRR  & \ROne & \RFive& \RTen &\Mean & & \NDCG          & \MRR  & \ROne & \RFive& \RTen & \Mean \\
  \cmidrule{1-8} \cmidrule{10-15}
  & MN  \cite{das2017visual}                   & 55.13          & 60.42 & 46.09 & 78.14 & 88.05 & 4.63 & & 56.99          & 47.83 & 38.01 & 57.49 & 64.08 & 18.76 \\
  & CoAtt  \cite{wu2018you}                    & 57.72          & 62.91 & 48.86 & 80.41 & 89.83 & 4.21 & & 59.24          & 49.64 & 40.09 & 59.37 & 65.92 & 17.86 \\
  & HCIAE  \cite{lu2017best}                   & 57.75          & 62.96 & 48.94 & 80.5  & 89.66 & 4.24 & & 59.70          & 49.07 & 39.72 & 58.23 & 64.73 & 18.43 \\
  & ReDAN  \cite{Gan2019MultistepRV}           & 59.32          & 64.21 & 50.6  & 81.39 & 90.26 & 4.05 & & 60.47          & 50.02 & 40.27 & 59.93 & 66.78 & 17.4  \\
  \midrule
  & \textbf{LTMI}                                       & \textbf{62.72} & 62.32 & 48.94 & 78.65 & 87.88 & 4.86 & & \textbf{63.58} & 50.74 & 40.44 & 61.61 & 69.71 & 14.93 \\
  \bottomrule
\end{tabular}
}
\label{tab:table_val_v1}
\end{table}
We first compare single-model performance on the val v1.0 split. 
We select here MN, CoAtt, HCIAE, and ReDAN for comparison, as their performances from the both decoders in all metrics are available in the literature. To be specific, we use the accuracy values reported in \cite{Gan2019MultistepRV} for a fair comparison, in which these methods are reimplemented using the bottom-up-attention features.
Similar to ours, all these methods employ the standard design of discriminative and generative decoders as in \cite{das2017visual}. Table \ref{tab:table_val_v1} shows the results. It is seen that our method outperforms all the compared methods on the NDCG metric with large margins regardless of the decoder type. Specifically, as compared with ReDAN, the current state-of-the-art on the VisDial v1.0 dataset, our model has improved NDCG from 59.32 to 62.72 and from 60.47 to 63.58 with discriminative and generative decoders, respectively. 

\bigskip
\noindent{\bf Results on the test-standard v1.0 split}~
We next consider performance on the test-standard v1.0 split. In our experiments, we encountered a phenomenon that accuracy values measured by NDCG and other metrics show a trade-off relation (see the supplementary material for details), depending much on the choice of metrics (i.e., NDCG or others) for judging convergence at the training time. This is observed in the results reported in \cite{Gan2019MultistepRV} and is attributable to the inconsistency between the two types of metrics. Thus, we show two results here, the one obtained using NDCG for judging convergence and the one using MRR for it; the latter is equivalent to performing early stopping.

Table \ref{tab:result_visdial_single_model} shows single-model performances on the blind test-standard v1.0 split. With the outputs from the discriminative decoder, our model gains improvement of 3.33pp in NDCG from the best model. When employing the aforementioned early stopping, our model achieves at least comparable or better performance in other metrics as well.

\begin{table}[t]
    \centering
    \caption{Comparison in terms of \textbf{single-model} performance on the blind test-standard v1.0 split of the VisDial v1.0 dataset. The result obtained by early stopping on MRR metric is denoted by $\star$ and those with fine-tuning on dense annotations are denoted by $\dagger$.}
    \resizebox{\textwidth}{!}{
        \begin{tabular}{rcccccc}
        \toprule 
          Model & \textbf{NDCG} $\uparrow $ & MRR $\uparrow$  & R@1 $\uparrow$    & R@5 $\uparrow$  & R@10 $\uparrow$  & Mean $\downarrow$ \\
          \midrule 
            LF \cite{das2017visual}            & 45.31 & 55.42 & 40.95 & 72.45 & 82.83 & 5.95 \\
            HRE \cite{das2017visual}           & 45.46 & 54.16 & 39.93 & 70.45 & 81.50 & 6.41 \\
            MN \cite{das2017visual}            & 47.50 & 55.49 & 40.98 & 72.30 & 83.30 & 5.92 \\
            MN-Att \cite{das2017visual}        & 49.58 & 56.90 & 42.42 & 74.00 & 84.35 & 5.59 \\
            LF-Att \cite{das2017visual}        & 49.76 & 57.07 & 42.08 & 74.82 & 85.05 & 5.41 \\
            FGA \cite{Schwartz2019FactorGA}    & 52.10 & 63.70 & 49.58 & \textbf{80.97} & 88.55 & 4.51 \\
            GNN \cite{Zheng2019ReasoningVD}    & 52.82 & 61.37 & 47.33 & 77.98 & 87.83 & 4.57 \\
            CorefNMN \cite{kottur2018visual}   & 54.70 & 61.50 & 47.55 & 78.10 & 88.80 & 4.40 \\
            RvA \cite{Niu_2019_CVPR}           & 55.59 & 63.03 & 49.03 & 80.40 & 89.83 & 4.18 \\
            Synergistic \cite{Guo_2019_CVPR}   & 57.32 & 62.20 & 47.90 & 80.43 & 89.95 & 4.17 \\
            DAN  \cite{Kang2019DualAN}         & 57.59 & 63.20 & 49.63 & 79.75 & 89.35 & 4.30 \\
            \midrule
             \textbf{LTMI}$^\star$                       & 59.03 & \textbf{64.08} & \textbf{50.20} & 80.68 & \textbf{90.35} & \textbf{4.05} \\
            \textbf{LTMI}                               & \textbf{60.92} & 60.65 & 47.00 & 77.03 & 87.75 & 4.90  \\
          \bottomrule
        \end{tabular}
    }
    \label{tab:result_visdial_single_model}
\end{table}

\begin{table}[t]
    \centering
    \caption{Comparison in terms of \textbf{ensemble-model} performance on the blind test-standard v1.0 split of the VisDial v1.0 dataset. The result obtained by early stopping on MRR metric is denoted by $\star$ and those with fine-tuning on dense annotations are denoted by $\dagger$.}
    \small
    \resizebox{0.85\columnwidth}{!}{
        \begin{tabular}{rcccccc}
        \toprule 
        Model                                         & \textbf{NDCG} $\uparrow$  & MRR $\uparrow$  & R@1 $\uparrow$   & R@5 $\uparrow$   & R@10 $\uparrow$   & Mean $\downarrow$ \\
        \midrule
        FGA \cite{Schwartz2019FactorGA}   & 52.10 & 67.30 & 53.40 & 85.28 & 92.70 & 3.54 \\
        Synergistic \cite{Guo_2019_CVPR}         & 57.88 & 63.42 & 49.30 & 80.77 & 90.68 & 3.97 \\
        DAN  \cite{Kang2019DualAN}        & 59.36 & 64.92 & 51.28 & 81.60 & 90.88 & 3.92 \\
        ReDAN \cite{Gan2019MultistepRV}   & 64.47 & 53.73 & 42.45 & 64.68 & 75.68 & 6.63 \\
        \textbf{LTMI}                          & 66.53 & 63.19 & 49.18 & 80.45 & 89.75 & 4.14 \\ 
        \midrule
        P1\_P2\cite{qi2020two}$^{\dagger}$ & 74.91          &	49.13  & 36.68 & 62.98 &    78.55 &    7.03 \\
        VD-BERT\cite{wang2020vd}$^\dagger$  & \textbf{75.13} & 50.00 &	38.28 & 60.93 &	    77.28  &	6.90 \\
        \textbf{LTMI}$^\dagger$              & 74.88 & \textbf{52.14} & \textbf{38.93} & \textbf{66.60} & \textbf{80.65} & \textbf{6.53} \\
        \bottomrule
        \end{tabular}
    }
    \label{tab:result_visdial_ensemble_model}
\end{table}

\begin{table}[t]
    \centering
    \caption{Comparison in terms of the number of parameters of the attention mechanism. The result obtained by early stopping on MRR metric is denoted by $\star$ and those with fine-tuning on dense annotations are denoted by $\dagger$.}
    \resizebox{0.7\columnwidth}{!}{
        \begin{tabular}{rcccccc}
        \toprule 
        \multicolumn{2}{r}{Model} & \multicolumn{2}{r}{\# params} & &\MRR & \textbf{NDCG}$\uparrow$ \\
          \midrule 
          \multicolumn{2}{r}{DAN   \cite{Kang2019DualAN}}           && \multicolumn{2}{l}{12.6M} & 63.20  & 57.59 \\
          \multicolumn{2}{r}{RvA   \cite{Niu_2019_CVPR}}            && \multicolumn{2}{l}{11.9M} & 63.03  & 55.59 \\
          \multicolumn{2}{r}{Naive Transformer}                     && \multicolumn{2}{l}{56.8M} & 62.09  & 55.10 \\
          \midrule
          \multicolumn{2}{r}{\textbf{LTMI}* (MRR-based)}                     && \multicolumn{2}{l}{4.8M} &  \textbf{64.08}  & 59.92 \\
          \multicolumn{2}{r}{\textbf{LTMI} (Q, V)}                           && \multicolumn{2}{l}{4.8M} &  60.65  & 60.92  \\
          \multicolumn{2}{r}{\textbf{LTMI} (Q, V, R)}                        && \multicolumn{2}{l}{4.8M} &  60.76  & \textbf{61.12}  \\
          \bottomrule
        \end{tabular}
    }
\label{tab:result_visdial_param}
\end{table}

Many previous studies report the performance of an ensemble of multiple models. To make a comparison, we create an ensemble of 16 models with some differences, from initialization with different random seeds to whether to use sharing weights across attention blocks or not, the number of attention blocks (i.e. $L$ = 2, 3), and the number of objects in the image (i.e. $K$ = 50, 100).
Aiming at achieving the best performance, we also enrich the image features by incorporating the class label and attributes of each object in an image, which are also obtained from the pre-trained Faster R-CNN model. Details are given in the supplementary material. We take the average of the outputs (probability distributions) from the discriminative decoders of these models to rank the candidate answers. Furthermore, we also test fine-tuning each model with its discriminative decoder on the available dense annotations from the train v1.0 and val v1.0, where the cross-entropy loss with soft labels (i.e. relevance scores) is minimized for two epochs. Table \ref{tab:result_visdial_ensemble_model} shows the results. It is observed that our ensemble model (w/o the fine-tuning) achieves the best $\textrm{NDCG}=66.53$ in all the ensemble models.

With optional fine-tuning, our ensemble model further gains a large improvement in NDCG, resulting in the third place in the leaderboard. 
The gap in NDCG to the first place (VD-BERT) is only 0.25pp, while our model yields performance that is better in all the other metrics, i.e, by 2.14pp, 5.67pp, and 3.37pp in MRR, R@5, and R@10, respectively, and 5.36\% reduction in Mean. 

Table \ref{tab:result_visdial_param} shows the number of parameters of the multi-modal attention mechanism employed in the recent methods along with their NDCG scores on the VisDial v1.0 test-standard set. We exclude the parameters of the networks computing the input utilities and the decoders, as they are basically shared among these methods. `Naive Transformer' consists of two stacks of transformer blocks with simple extension to three utilities as mentioned in Sec.~\ref{introduction}. The efficiency of our models can be observed. 
Note also that the gap between (Q, V) and (Q, V, R) is small, contrary to the argument in \cite{qi2020two}.
\begin{table}[t]
    \centering
    \caption{ Ablation study on the components of our method on the val v1.0 split of VisDial dataset. $\uparrow$ indicates the higher the better.}
    \resizebox{0.8\columnwidth}{!}{
        \begin{tabular}{l|c|ccc}
            \multicolumn{5}{c}{\normalsize \centering (a)  
            }\\
            \toprule
            Component & Details & A-NDCG $\uparrow$ & D-NDCG  $\uparrow$ & G-NDCG $\uparrow$  \\
            \midrule
            Number of           & 1     & 65.37  & 62.06 & 62.95 \\
            attention blocks    & 2     & \textbf{65.75}  & \textbf{62.72} & \textbf{63.58} \\
                                & 3     & 65.42  & 62.48 & 63.22   \\
            \midrule
            Self-Attention      &  No   & 65.38  & 61.76 & 63.31           \\
                                &  Yes  & \textbf{65.75}  & \textbf{62.72} & \textbf{63.58} \\
            \midrule
            Attended features   &  Add     & 64.12  & 60.28 & 61.49           \\
            aggregation         &  Concat  & \textbf{65.75}  & \textbf{62.72} & \textbf{63.58} \\
            
            \midrule 
            Shared Attention    & No    & \textbf{65.75}  & \textbf{62.72} & \textbf{63.58} \\ 
                   weights      & Yes   & 65.57  & 62.50 & 63.24  \\          
                
        \bottomrule
        \end{tabular}
    
    }
    \label{tab:visdial_0.9_ablative_1}
    \end{table}

\begin{table}[t]
    \centering
    \caption{ Ablation study on the components of our method on the val v1.0 split of VisDial dataset. $\uparrow$ indicates the higher the better.}
    \resizebox{0.8\columnwidth}{!}{
        \begin{tabular}{l|c|ccc}
        \multicolumn{5}{c}{\normalsize \centering (b)}\\
        \toprule
        Component & Details & A-NDCG $\uparrow$ & D-NDCG  $\uparrow$ & G-NDCG $\uparrow$  \\
    
        \midrule
        Context feature     &  [Q]       & 65.12  & 61.50 & 63.19           \\
        aggregation         &  [Q, V]    & \textbf{65.75}  & \textbf{62.72} & \textbf{63.58} \\
                            &  [Q, V, R] & 65.53  & 62.37 & 63.38            \\
        \midrule
        Decoder Type        & Gen     & -  & - & 62.35 \\
                            & Disc    & -  & 61.80 & - \\
                            & Both    & \textbf{65.75}  & \textbf{62.72} & \textbf{63.58} \\
        \midrule
        The number of           & 36        & 65.25  & 62.40 & 63.08  \\
        objects in an image     & 50        & 65.24  & 62.29 & 63.12  \\
                                & 100  & \textbf{65.75}  & \textbf{62.72} & \textbf{63.58} \\         
        \midrule 
        Positional and         & No    & 65.18  & 61.84 & 62.96           \\
        spatial embeddings     & Yes   & \textbf{65.75}  & \textbf{62.72} & \textbf{63.58} \\
        
        \bottomrule 
        \end{tabular}
        \label{tab:visdial_0.9_ablative_2}
    }
    \end{table}

\subsection{Ablation Study} \label{sec:abalation_study}
To evaluate the effect of each of the components of our method, we perform the ablation study on the val v1.0 split of VisDial dataset. We evaluate here the accuracy of the discriminative decoder and the generative decoder separately. We denote the former by D-NDCG and the latter by G-NDCG, and the accuracy of their averaged model by A-NDCG (i.e., averaging the probability distributions over the candidate answers obtained by the discriminative and generative decoders). The results are shown in Table \ref{tab:visdial_0.9_ablative_1} and Table \ref{tab:visdial_0.9_ablative_2}.

The first block of Table \ref{tab:visdial_0.9_ablative_1} shows the effect of the number of stacks of the proposed attention blocks. We observe that the use of two to three stacks achieves good performance on all three measures. More stacks did not bring further improvement, and thus are omitted in the table. 

The second block of Table \ref{tab:visdial_0.9_ablative_1} shows the effect of self-attention, which computes the interaction within a utility, i.e., ${\bar{\mathcal{A}}}_X(X)$. 
We examine this because it can be removed from the attention computation. It is seen that self-attention does contribute to good performance. The third block shows the effects of how to aggregate the attended features.
It is seen that their concatenation yields better performance than their simple addition. The fourth block shows the impact of sharing the weights across the stacks of the attention blocks. If the weights can be shared as in \cite{lan2019albert}, it contributes a further decrease in the number of parameters. We observe that the performance does drop if weight sharing is employed, but the drop is not very large.

The first block of Table \ref{tab:visdial_0.9_ablative_2} shows the effect of how to aggregate the context features $c_V$, $c_Q$, and $c_R$ in the decoder(s), which are obtained from the outputs of our encoder. As mentioned above, the context vector $c_R$ of the dialog history does not contribute to the performance. However, the context vector $c_v$ of the image is important for achieving the best performance. The second block of Table \ref{tab:visdial_0.9_ablative_2} shows the effects of simultaneously training the both decoders (with the entire model). It is seen that this contributes greatly to the performance; this indicates the synergy of learning two tasks while sharing the encoder, resulting better generalization as compared with those trained with a single decoder. 

We have also confirmed that the use of fewer objects leads to worse results. Besides, the positional embedding for representing the question and history utilities as well as the spatial embedding (i.e., the bounding box geometry of objects) for image utility representation have a certain amount of contribution. 

\subsection{Qualitative Results}  
\paragraph{Visualization of Generated Attention}\label{sec:visualization}
\begin{figure}[t]
\centering
\includegraphics[width=1.0\textwidth]{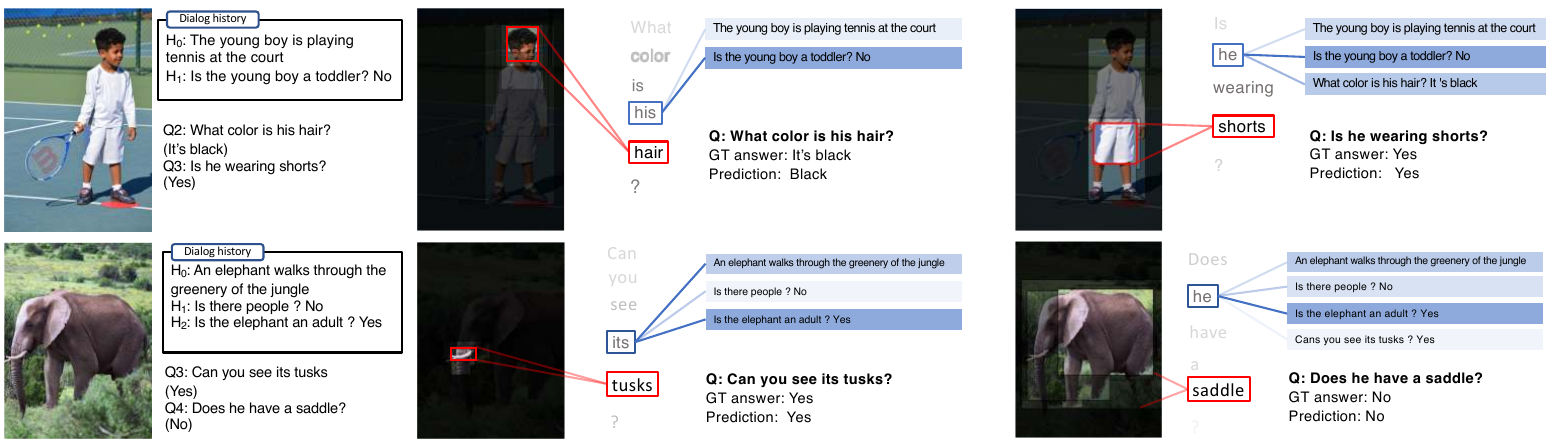}
\caption{\color{black}Examples of visualization for the attention weights
generated in our model at two Q\&A rounds on two images. See Sec.~\ref{sec:visualization} for details.
}
\label{fig:qualitative_results}
\vskip -0.1in
\end{figure}
Figure \ref{fig:qualitative_results} shows attention weights generated in our model on two rounds of Q\&A on two images. We show here two types of attention. One is the self-attention weights used to compute the context vectors $c_V$ and $c_Q$. For $c_V$, the attention weights $a_{V}$ are generated over image regions (i.e., bounding boxes), as in Eq.(\ref{eqn:a_v_weights}). Similarly, for $c_Q$, the attention weights are generated over question words. 
These two sets of attention weights are displayed by brightness of the image bounding-boxes and darkness of question words, respectively, in the center and the rightmost columns. It can be observed from these that the relevant regions and words are properly highlighted at each Q\&A round. 

The other attention we visualize is the source-to-target attention computed inside the proposed block. We choose here the image-to-question attention $\bar{\mathcal{A}}_V(Q)$ and the history-to-question attention $\bar{\mathcal{A}}_R(Q)$. For each, we compute the average of the attention weights over all the heads computed inside the block belonging to the upper stack. 
In Fig.~\ref{fig:qualitative_results}, the former is displayed by the red boxes connected between an image region and a question word; only the region with the largest weight is shown for the target word; the word with the largest self-attention weight is chosen for the target. The history-to-question attention is displayed by the Q\&As highlighted in blue color connected to a selected question word that is semantically ambiguous, e.g., {\em `its'}, {\em `he'}, and {\em `his'}. It is seen that the model performs proper visual grounding for the important words, {\em`hair'}, {\em`shorts'}, and {\em'tusks'}. It is also observed that the model properly resolves the co-reference for the words, {\em `he'} and {\em `its'}. 

We provide additional examples of the results obtained by our method in Figs.~\ref{fig:s1}-\ref{fig:d2}. They are divided into two groups, results for which the top-1 prediction coincides with the ground truth answer (Figs.~\ref{fig:s1}-\ref{fig:s2}) and those for which they do not coincide (Figs.~\ref{fig:d1}-\ref{fig:d2}). For each result, we show the attention maps created on the input image and question, respectively.

\newcommand{\figw}{1.0\textwidth}

\begin{figure*}[ht]
\centering
\centerline{
\includegraphics[width=\figw]{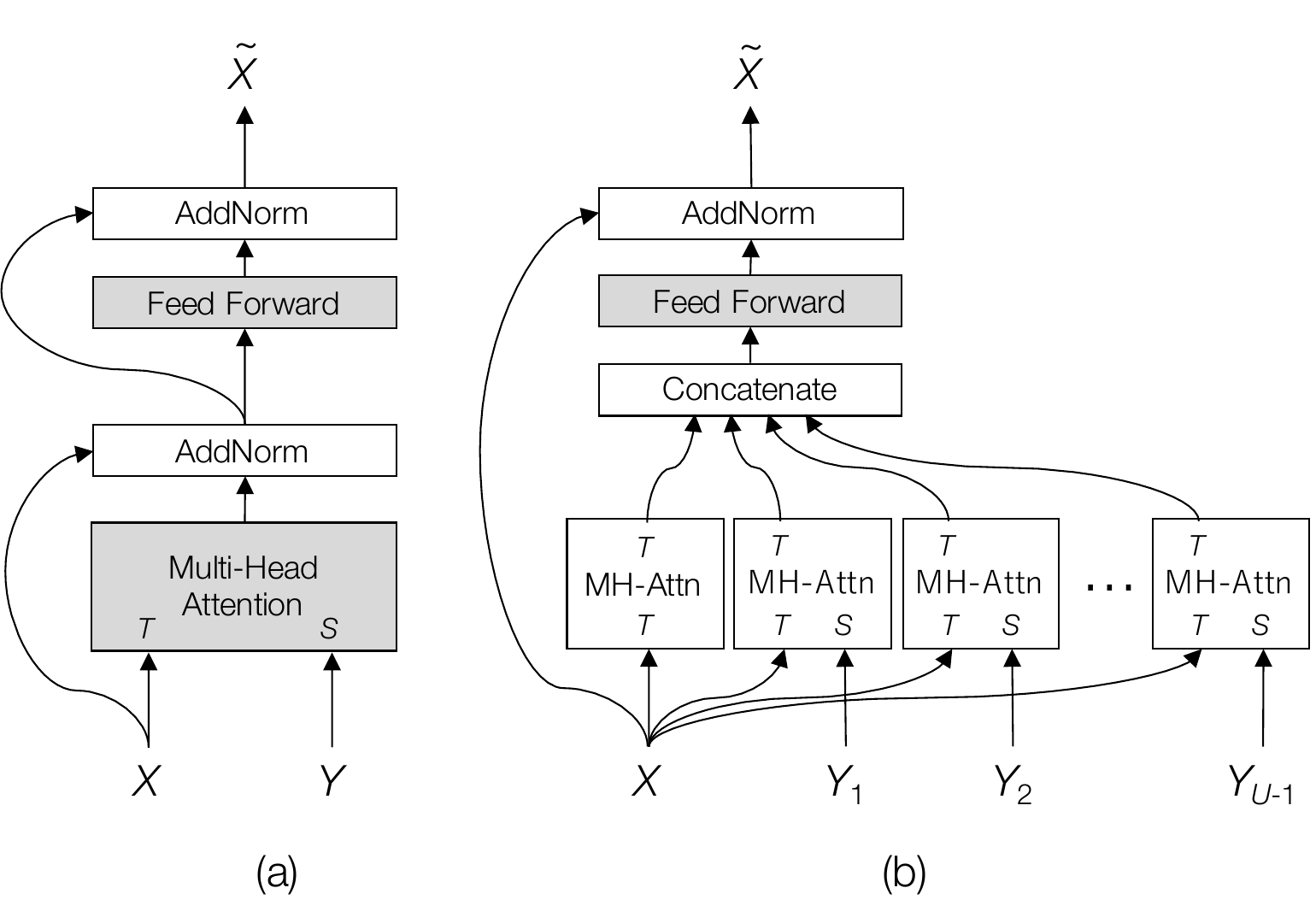}
}
\caption{Examples of results for which the top-1 prediction is the same as the ground truth answer on the validation split of Visdial v1.0. Each row shows selected two rounds of Q\&A for one image.}

\label{fig:s1}
\end{figure*}

\begin{figure*}[ht]
\centering
\centerline{
\includegraphics[width=\figw]{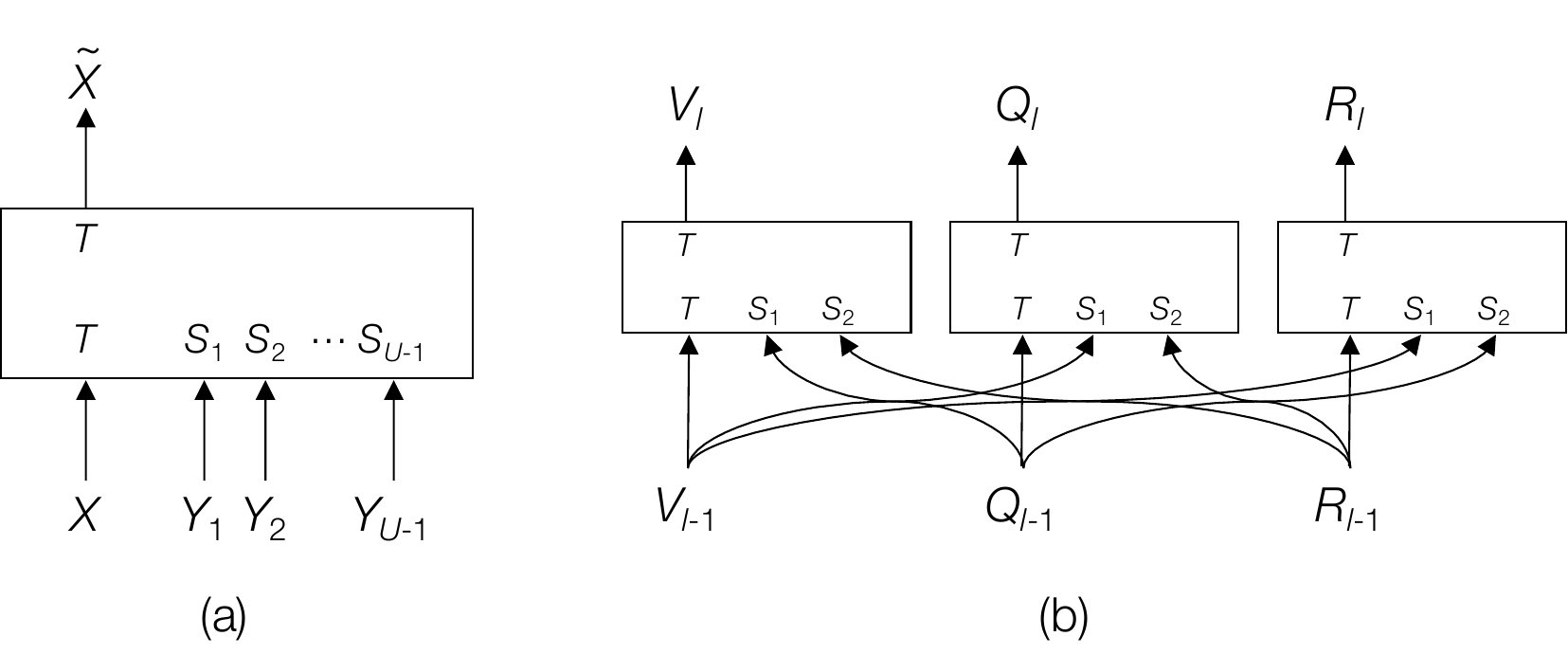}
}
\caption{Examples of results for which the top-1 prediction is the same as the ground truth answer on the validation split of Visdial v1.0. Each row shows selected two rounds of Q\&A for one image.}
\label{fig:s2}
\end{figure*}

\begin{figure*}[ht]
\centering
\centerline{
\includegraphics[width=\figw]{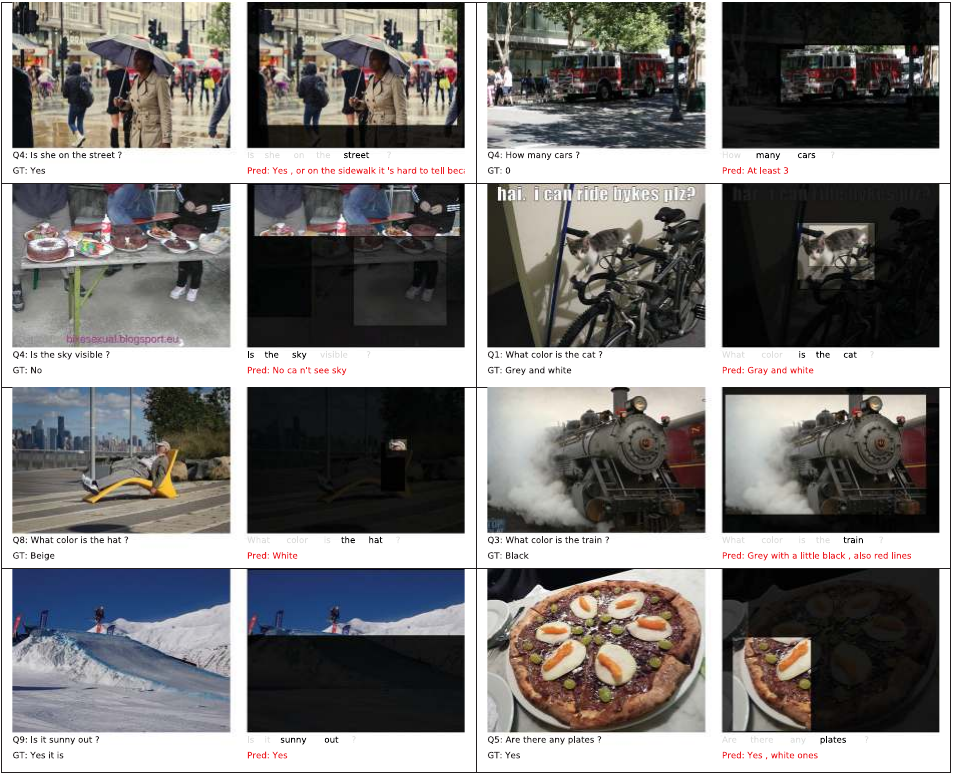}
}
\caption{
Examples of results for which the top-1 prediction is different from the ground truth answer on the validation split of Visdial v1.0.}
\label{fig:d1}
\end{figure*}

\begin{figure*}[ht]
\centering
\centerline{
\includegraphics[width=\figw]{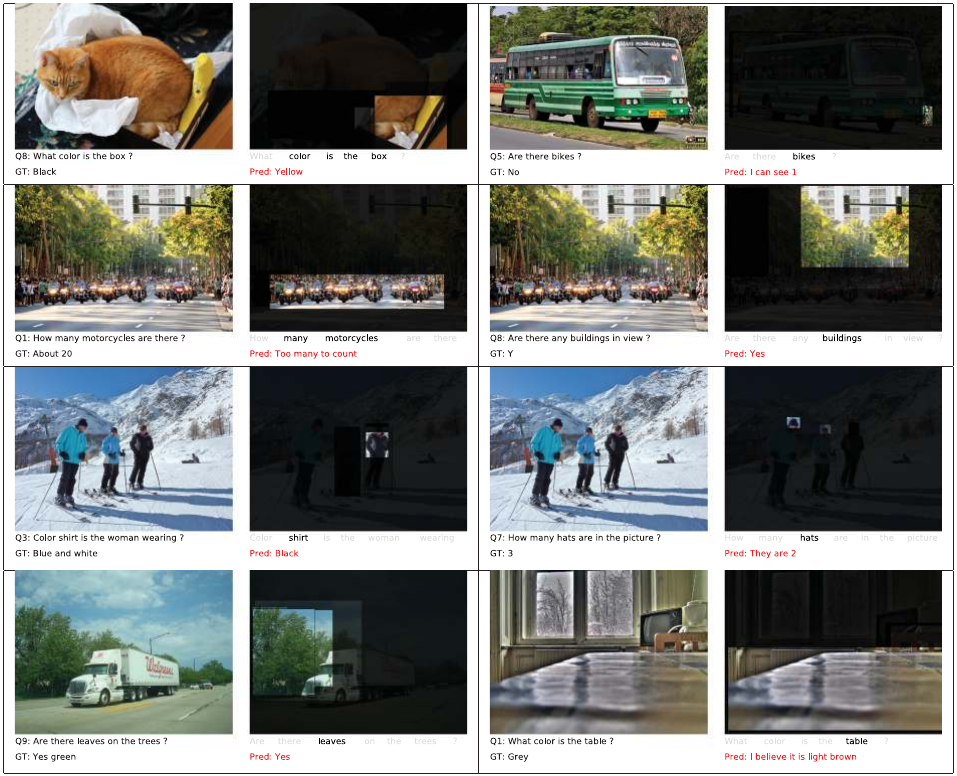}
}
\caption{
Examples of results for which the top-1 prediction is different from the ground truth answer on the validation split of Visdial v1.0.}
\label{fig:d2}
\end{figure*}

\section{Experiments on Audio Visual Scene-aware Dialog}
To test the generality of the proposed method on other tasks as well as its performance on a greater number of utilities, we additionally apply it to the Audio Visual Scene-aware Dialog (AVSD) task \cite{hori2019end}. This task requires a system to generate an answer to a question about events seen in a video given with a previous dialog. AVSD provides more utilities than Visual Dialog, i.e.,  audio features and video features, such as VGG or I3D features (I3D RGB sequence and I3D flow sequence). We build a network by simply replacing the multi-modal attention mechanism in the baseline model of \cite{hori2019end} with a simple extension of the proposed attention mechanism. Details are given below.

\subsection{Network Design}
\begin{table*}[t]
\caption{Comparison of response generation evaluation results with objective measures.}
\resizebox{1.0\columnwidth}{!}{
    \begin{tabular}{lc|ccccccc}
    \toprule
    Model                                        & Video Feat. & CIDEr          & BLEU1          & BLEU2          & BLEU3          & BLEU4          & METEOR         & ROUGE\_L       \\
    \midrule
    Baseline \cite{hori2019end} & VGG         & 0.618          & 0.231          & 0.141          & 0.095          & 0.067          & 0.102          & 0.259          \\
    Ours                        & VGG         & \textbf{0.841} & \textbf{0.266} & \textbf{0.172} & \textbf{0.118} & \textbf{0.086} & \textbf{0.117} & \textbf{0.296} \\
    \midrule
    Baseline \cite{hori2019end} & I3D         & 0.727          & 0.256          & 0.161          & 0.109          & 0.078          & 0.113          & 0.277          \\
    Ours                        & I3D         & \textbf{0.851} & \textbf{0.277} & \textbf{0.178} & \textbf{0.122} & \textbf{0.088} & \textbf{0.119} & \textbf{0.302} \\
    \bottomrule
    \end{tabular}
}
\label{tab:avsd}
\end{table*}
Following the baselines \cite{hori2019end}, we extract the question utility $Q$ using a two-layer LSTM. We separate the caption from the dialog history and feed it into another two-layer LSTM to obtain the caption utility $C$. Similar to \cite{hori2019end}, the dialog history consisting of previous question-answer pairs is inputted into a hierarchical LSTM network; specifically, we encode each question-answer pair with one LSTM and summarize the obtained encodings with another LSTM, yielding a final vector representation $c_r$. All LSTMs used for language encoding have $d$ units. We convert words into vectors with a shared embedding layer initialized with GLoVe vectors.

The video provides two sources of features, i.e., video features and audio features. We use the audio features extracted from the pre-trained VGGish model \cite{hori2019end}, which are fed to a projection layer, providing the audio utility $A$; it is represented as a collection of $d$-dimensional vectors. For video processing, following \cite{hori2019end}, we consider two models with different features: i) VGG features extracted from four uniformly sampled frames in the video, giving the video utility $V$, and ii) I3D features extracted by the I3D network pre-trained on an action recognition task, which are forwarded to  projection layers to obtain an I3D-rgb utility and an I3D-flow utility denoted by $V$ and $F$.

To compute the multi-modal attention between $U$ utilities, we add a stack of $U$ proposed attention blocks; $U = 4$ for the model (i) and $U=5$ for (ii).
To make the designs of two models (i) and (ii) similar, we use only $A$ utility to attend language utilities; and only $Q$ and $C$ are allowed to attend audio and video utilities. After obtaining the updated representations of all utilities, we summarize each utility into a single vector by the self-attention mechanism, in which the summarized vector of question utility is denoted by $c_q$. We concatenate all these vectors together with $c_r$, projecting it into a $d$-dimensional vector of context representation $c$. 

The decoder architecture is similar to the generative decoder described in Sec. \ref{sec:gen_decoder} except that the input of the decoder at the $i$-th step is the concatenation of $w_{i-1}$, $c_q$, and $c_r$. At the time of inference, we use the beam search technique to efficiently find the most likely hypothesis generated by the decoder.

\subsection{Experimental Setup}
Following \cite{hori2019end}, we perform the experiment on the AVSD prototype which is split into training, validation, and test sets with 6172, 732, and 733 videos, respectively. Each video is collected from the Charades dataset, annotated with a caption and 10 dialog rounds.
The hidden size $d$ is set to 512; the GLoVe vectors are $300$-dimensional. We train the models in 15 epochs using the Adam optimizer with initial learning rate $1\times10^{-3}$ in all the experiments. The dropout with rate of 0.2 is applied for the LSTMs.

\subsection{Experimental Results}
   
Table \ref{tab:avsd} shows 
the results, which include evaluation on a number of metrics to measure the quality of generated answers, i.e. CIDEr, BLEU, METEOR, ROUGE\_L. It is seen that our models outperform the baselines presented in \cite{hori2019end} over all the metrics; {
specifically, it improves the CIDEr score by 22.3\% (from 0.618 to 0.841) with VGG features and by 12.4\% (from 0.727 to 0.851) with I3D features.}

\section{Summary and Conclusion}

In this chapter, we have proposed LTMI (Lightweight Transformer for Many Inputs)
that can deal with all the interactions between multiple input utilities in an efficient way. As compared with other methods, the proposed architecture is much simpler in terms of the number of parameters as well as the way of handling inputs (i.e., their equal treatment), and nevertheless surpasses the previous methods in accuracy; it achieves the new state-of-the-art results on the VisDial datasets, e.g., high NDCG scores on the VisDial v1.0 dataset. Thus, we believe our method can be used as a simple yet strong baseline.

\cleardoublepage
\chapter{
LWIT: Improving Performance on Instruction Following Tasks}
\label{chapter:ch5}

\section{Introduction}

The two preceding chapters established two core capabilities that an intelligent vision-language agent must possess.
Chapter~\ref{chapter:ch3} showed that extracting complementary grid-based and region-based visual features significantly improves how accurately an agent can represent and describe a scene.
Chapter~\ref{chapter:ch4} extended the scope to multi-round dialog, proposing a lightweight mechanism to model all pairwise interactions among the image, the question, and the dialog history simultaneously, and demonstrating that doing so leads to substantial improvements over methods that consider only a subset of these interactions.
Despite these advances, both tasks share a critical limitation that was noted in Chapter~\ref{chapter:ch1}: the agent is fundamentally passive.
It observes a static scene and produces responses, but it neither moves through an environment nor manipulates the objects within it.
True embodied intelligence requires closing this loop between perception and action: an agent must be able to follow a sequence of natural language instructions, navigate through environments it may not have seen before, and carry out physical interactions with objects in order to accomplish a goal.
This is precisely the challenge we address in this chapter through the ALFRED benchmark for interactive instruction following.
ALFRED represents the most demanding task in this dissertation, requiring the agent to integrate the visual perception capabilities from Chapter~\ref{chapter:ch3}, the multi-input reasoning strategies from Chapter~\ref{chapter:ch4}, and a new capacity for sequential decision-making, all while remaining robust over long action horizons and in unseen environments.

To appreciate why ALFRED presents such a substantial leap in difficulty, it is helpful to trace the line of work it builds upon.
Vision-language navigation (VLN) tasks have made significant progress in recent years \cite{anderson2018vision,fried2018speaker,zhu2020vision}, but they operate in static environments with a simplified action space and no object interaction.
ALFRED \cite{shridhar2020alfred} extends this setting into fully interactive household environments, which introduces three compounding challenges: the agent must (1) reason over a greater number of instructions and align them with the correct moment in an execution horizon that is far longer than in VLN; (2) predict actions from a substantially larger action space spanning both navigation and manipulation; and (3) localize specific objects by predicting pixel-wise masks to carry out manipulation actions.
Approaches that transfer well to VLN, such as Seq2Seq models with soft attention and a progress monitor \cite{shridhar2020alfred,ma2019selfmonitoring}, perform poorly on ALFRED, and overall, existing methods leave a large gap relative to human performance.

In this chapter, we propose a new method that leads to significant performance improvements. It is based on several ideas. Firstly, we propose to choose a single instruction to process at each timestep from the given series of instructions. This approach contrasts with previous methods that encode them into a single long sequence of word features and use soft attention to specify which instruction to consider at each timestep implicitly \cite{shridhar2020alfred,legg2020eccv,singh2020eccv}. Our method chooses individual instructions explicitly by learning to predict when the agent completes an instruction. This makes it possible to utilize constraints on parsing instructions, leading to a more accurate alignment of instructions and action prediction. 
Secondly, we propose a two-stage approach to the interpretation of the selected instruction. In its first stage, the method interprets the  instruction without using visual inputs from the environment,  yielding a tentative prediction of an action-object sequence. In the second stage, the prediction is integrated with the visual inputs to predict the action to do and the object to manipulate. The tentative interpretation makes it clear to interact with what class of objects, contributing to an accurate selection of objects to interact with. 

Moreover, we acquire multiple agent egocentric views of a scene as visual inputs and integrate them using a hierarchical attention mechanism. This allows the agent to have a wider field of views, leading to more accurate navigation. To be specific, converting each view into an object-centric representation, we integrate those for the multiple views into a single feature vector using hierarchical attention conditioned on the current instruction. 

Besides, we propose a module for predicting precise pixel-wise masks of objects to interact with, referred to as the mask decoder. It employs the object-centric representation of the center view, i.e., multiple object masks detected by the object detector. The module selects one of these candidate masks to specify the object to interact with. In the selection,  self-attention is applied to the candidate masks to weight them; they are multiplied with the tentative prediction of the pairs of action and an object class and the detector's confidence scores for the candidate masks.

The experimental results show that the proposed method outperforms all the existing methods by a large margin and ranks first in the challenge leaderboard as of the time of submission. A preliminary version of the method won an international competition held last year. The present version further improved the task success rate in unseen and seen environments to 8.37\% and 29.16\%, respectively, which are significantly higher than the previously published SOTA (0.39\% and 3.98\%, respectively) \cite{shridhar2020alfred}. 

\section{Related Work}

\subsection{Embodied Vision-Language Tasks}

Many studies have been recently conducted on the problem of making an embodied AI agent follow natural language directives and accomplish the specified tasks in a three-dimensional environment while properly interacting with it. Vision-language navigation (VLN) tasks have been the most extensively studied, which require an agent to follow navigation directions in an environment.

Several frameworks and datasets for simulating real-world environments have been developed to study the VLN tasks. The early ones lack photo-realism and/or natural language directions \cite{kempka2016vizdoom,ai2thor,wu2018building}. Recent studies consider perceptually-rich simulated environments and natural language navigation directions  \cite{anderson2018vision,chen2019touchdown,hermann2020learning}.
In particular, since the release of the Room-to-Room (R2R) dataset \cite{anderson2018vision} that is based on real imagery \cite{chang2017matterport3d}, VLN has attracted increasing attention, leading to the development of many methods \cite{fried2018speaker,wang2019reinforced,ma2019selfmonitoring,tan2019learning,majumdar2020improving}.

Several variants of VLN tasks have been proposed. A study \cite{Nguyen_2019_CVPR} allows the agent to communicate with an adviser using natural language to accomplish a given goal.          
In a study \cite{thomason2020vision}, the agent placed in an environment attempts to find a specified object by communicating with a human by natural language dialog. 
{
A recent study \cite{suhr-etal-2019-executing} proposes the interactive environments where users can collaborate with an agent by not only instructing it to complete tasks, but also acting alongside it.}
Another study \cite{krantz2020beyond} introduces a continuous environment based on the R2R dataset that enables an agent to take more fine-grained navigation actions.
A number of other embodied vision-language tasks have been proposed such as visual semantic planning \cite{zhu2017visual,gordon2019should} and embodied question answering \cite{embodiedqa,gordon2018iqa,wijmans2019embodied,puig2018virtualhome}.

\subsection{Existing Methods for ALFRED}
As mentioned earlier, ALFRED was developed to consider more complicated interactions with environments, which are missing in the above tasks, such as manipulating objects. Several methods for it have been proposed so far. A baseline method \cite{shridhar2020alfred} employs a Seq2Seq model with an attention mechanism and a progress monitor \cite{ma2019selfmonitoring}, which is prior art for the VLN tasks. In \cite{singh2020eccv}, a pre-trained Mask R-CNN is employed to generate object masks. It is proposed in \cite{legg2020eccv} to train the agent to follow instructions and reconstruct them. In \cite{corona2020modularity}, a modular architecture is proposed to exploit the compositionality of instructions. These methods have brought about only modest performance improvements over the baseline. A concurrent study \cite{singh2020moca} proposes a modular architecture design in which the prediction of actions and object masks are treated separately, as with ours. Although it achieves notable performance improvements, the study's ablation test indicates that the separation of the two is not the primary source of the improvements. {
Closely related to ALFRED, ALFWorld \cite{shridhar2021alfworld} has been recently proposed to combine TextWorld \cite{cote18textworld} and ALFRED for creating aligned environments, which enable transferring high-level policies learned in the text world to the embodied world.}

\section{Proposed Method}

The proposed model consists of three decoders (i.e., instruction, mask, and action decoders) with the modules extracting features from the inputs, i.e., the visual observations of the environment and the language directives. We first summarize ALFRED and then explain the components one by one.

\begin{figure*}[t!]
    \centering
    \includegraphics[width=\linewidth]{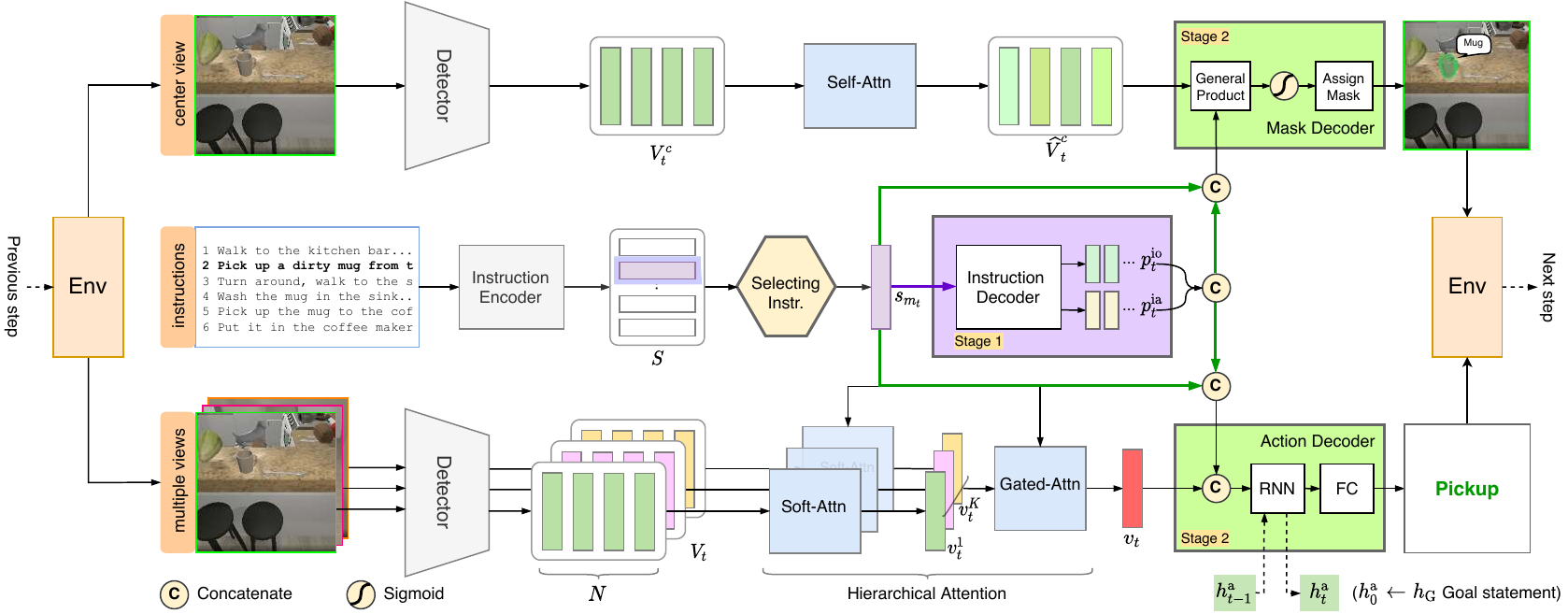}
    \caption{
        {Architecture overview of the proposed model.} It consists of the modules encoding the visual inputs and the language directives (Sec.~\ref{sec:feat_representation}), the instruction decoder with an instruction selector (Sec.~\ref{sec:inst_decoder}), the action decoder (Sec.~\ref{sec:action_decoder}), and the mask decoder (Sec.~\ref{sec:mask_decoder}).}         
    \vspace{-0.5em}
    \label{fig_overview}
\end{figure*}

\subsection{Summary of ALFRED}
ALFRED is built upon AI2Thor \cite{ai2thor}, a simulation environment for embodied AI. An agent performs seven types of tasks in 120 indoor scenes that require interaction with 84 classes of objects, including 26 receptacle object classes.
For each object class, there are multiple visual instances with different shapes, textures, and colors. 

The dataset contains 8,055 expert demonstration  episodes of task instances.
They are sequences of actions, whose average length is 50, and they are used as a ground truth action sequence at training time. For each of them, language directives annotated by AMT workers are provided, which consist of a goal statement $G$ and a set of step-by-step instructions, $S_{1},\ldots, S_{L}$. The alignment between each instruction and a segment of the action sequence is known. As multiple AMT workers annotate the same demonstrations, there are 25,743 language directives in total. 

We wish to predict the sequence of agent's actions, given $G$ and $S_1,\ldots,S_L$ of a task instance. 
There are two types of actions, navigation actions and manipulation actions. There are five navigation actions (e.g., \texta{MoveAhead} and \texta{RotateRight}) and seven manipulation actions (e.g., \texta{Pickup} and \texta{ToggleOn}). The manipulation actions accompany an object. The agent specifies it using a pixel-wise mask in the egocentric input image. Thus, the {\em outputs} are a sequence of actions with, if necessary, the object masks.

\subsection{Feature Representations} \label{sec:feat_representation}
\subsubsection{Object-centric Visual Representations}

Unlike previous studies \cite{shridhar2020alfred,singh2020eccv,legg2020eccv}, we employ the object-centric representations of a scene \cite{devin2018deep}, which are extracted from a pre-trained object detector (i.e., Mask R-CNN \cite{he2017mask}). It provides richer spatial information about the scene at a more fine-grained level and thus allows the agent to localize the target objects better. Moreover, we make the agent look wider by capturing the images of its surroundings, aiming to enhance its navigation ability.

Specifically, at timestep $t$, the agent obtains visual observations from $K$ egocentric views. For each view $k$, we encode the visual observation
by a bag of $N$ object features,
which are extracted the object detector. Every detected object 
is associated with a visual feature, a mask, and its confidence score. We project the visual feature
into $\mathbb{R}^{d}$ with a linear layer, followed by a ReLU activation and dropout regularization \cite{srivastava2014dropout} to obtain a single vector; thus, we get 
a set of $N$ object features for view $k$, $V^{k}_{t} = (v^{k}_{t,1}, \ldots, v^{k}_{t,N})$. 
We obtain $V^{1}_{t}, \ldots, V^{K}_{t}$ 
for all the views.

\subsubsection{Language Representations}
We encode the language directives as follows. We use an embedding layer initialized with pre-trained GloVe \cite{pennington2014glove} vectors to embed each word of the $L$ step-by-step instructions and the goal statement. For each instruction $i(=1,\ldots,L)$, the embedded feature sequence is inputted to a two-layer LSTM \cite{hochreiter1997long}, and its last hidden state is used as the feature $s_i\in\mathbb{R}^d$ of the instruction. We use the same LSTM for all the instructions with dropout regularization. We encode the goal statement $G$ in the same manner using an LSTM with the same architecture different weights, obtaining $h_\text{G}\in\mathbb{R}^d$.

\subsection{Instruction Decoder} \label{sec:inst_decoder} 

\subsubsection{Selecting Instructions}
Previous studies \cite{shridhar2020alfred,singh2020eccv,legg2020eccv} employ a Seq2Seq model in which all the language directives are represented as a {\em single sequence} of word features, and soft attention is generated over it to specify the portion to deal with at each timestep. 
We think this method could fail to correctly segment instructions with time, even with the employment of progress monitoring \cite{ma2019selfmonitoring}.
This method does not use a few constraints on parsing the step-by-step instructions that they should be processed in the given order and when dealing with one of them, the other instructions, especially the future ones, will be of little importance.

We propose a simple method that can take the above constraints into account, which explicitly represents which instruction to consider at the current timestep $t$. The method introduces an integer variable $m_t(\in[1,L])$ storing the index of the instruction to deal with at $t$. 

To update $m_t$ properly, we introduce a virtual action representing the {\em completion of a single instruction}, which we treat equally to the original twelve actions defined in ALFRED. Defining a new token \texttt{COMPLETE} to represent this virtual action, we augment each instruction's action sequence provided in the expert demonstrations 
always end with \texttt{COMPLETE}.
At training time, we train the action
decoder to predict the augmented sequences. At test time, the same decoder predicts an action at each timestep; if it predicts \texttt{COMPLETE}, this means completing the current instruction. 
The instruction index $m_t$ is updated as follows:%
\begin{equation}
    m_t= 
    \begin{cases}
        m_{t-1} + 1,& \text{if } \mathrm{argmax}(p^\text{a}_{t-1}) = \texttt{COMPLETE}\\
        m_{t-1},              & \text{otherwise},
    \end{cases}
\end{equation}
where $p^\text{a}_{t-1}$ is the predicted probability distribution over all the actions at time $t-1$, which will be explained in Sec.~\ref{sec:action_decoder}. 
The encoded feature $s_{m_t}$ of the selected instruction is used in all the subsequent components, as shown in Fig.~\ref{fig_overview}. 

\subsubsection{Decoder Design} 
\begin{figure}[!ht]
    \centering
    \includegraphics[width=1.0\linewidth]{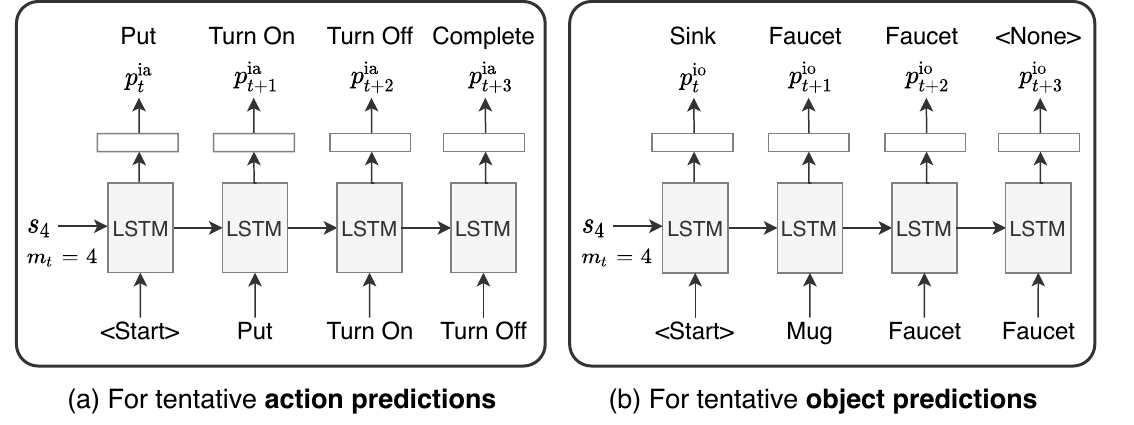}
    \caption{
        {
        An example illustrates how we reinitialize the hidden states of the two LSTMs in the instruction encoder by $s_{m_t}$ when $m_t = m_{t-1} + 1$ ($m_t = 4)$.}}    
    \label{fig:inst_decoder}
    \vspace{-0.2cm}
\end{figure}
As explained earlier, our method employs a two-stage approach for interpreting the instructions. The instruction decoder (see Fig.~\ref{fig_overview}) runs the first stage, where it interprets the instruction encoded as $s_{m_t}$ {\em without any visual input}. To be specific, it transforms $s_{m_t}$ into the sequence of action-object pairs without additional input. In this stage, objects mean the {\em classes} of objects. 

As it is not based on visual inputs, the predicted action-object sequence has to be tentative. The downstream components in the model (i.e., the mask decoder and the action decoder) interpret $s_{m_t}$ again, yielding the final prediction of an action-object sequence, which are grounded on the visual inputs. Our intention of this two-stage approach is to increase prediction accuracy; we expect that using a prior prediction of (action, object class) pairs helps more accurate grounding.

In fact, many instructions in the dataset, particularly those about interactions with objects, are sufficiently specific so that they are uniquely translated into (action, object class) sequences with a perfect accuracy, even without visual inputs. For instance,
``
Wash the mug in the sink'' can be  translated into 
(\texta{Put}, \texta{Sink}), (\texta{TurnOn}, \texta{Faucet}), 
(\texta{TurnOff}, \texta{Faucet}), (\texta{PickUp}, \texta{Mug}).
However, this is not the case with navigation instructions. 
For instance, 
``{Go straight to the sink}'' may be translated into a variable number of repetition of \texta{MoveAhead}; it is also hard to translate ``{Walk into the drawers}'' when it requires to navigate to the left/right. Therefore, we separately deal with the manipulation actions and the navigation actions. In what follows, we first explain the common part and then  the different parts.

Given the encoded feature $s_{m_t}$ of the selected instruction, the instruction decoder predicts the action and the object class to choose at $t$. To be precise, it outputs the probability distributions $p^\text{ia}_t(\in\mathbb{R}^{N_\text{a}})$ and $p^\text{io}_t(\in\mathbb{R}^{N_\text{o}})$ over all the actions and the object classes, respectively; $N_\text{a}$ and $N_\text{o}$ are the numbers of the actions and the object classes. 

These probabilities $p^\text{ia}_t$ and $p^\text{io}_t$ are predicted separately by two LSTMs in an autoregressive fashion.
The two LSTMs are initialized whenever a new instruction is selected; to be precise, we reset their internal states as $h^\text{ia}_{t-1}=h^\text{io}_{t-1}=s_{m_t}$ for $t$ when we increment $m_t$ as $m_t=m_{t-1}+1$ ({
see the example in Fig. \ref{fig:inst_decoder}}). Then, $p^\text{ia}_t$ and $p^\text{io}_t$  are predicted as follows:
\begin{subequations}
\begin{align}
        p^\text{ia}_{t} &= \mathrm{softmax}(W_\text{ia}\text{LSTM}(E_{\text{a}}(p^\text{ia}_{t-1}), h^\text{ia}_{t-1}) + b_\text{ia}),\\ 
        p^\text{io}_{t} &= \mathrm{softmax}(W_\text{io}\text{LSTM}(E_{\text{o}}(p^\text{io}_{t-1}), h^\text{io}_{t-1}) + b_\text{io}), 
\end{align}
\end{subequations}
where $W_\text{ia} \in \mathbb{R}^{N_\text{a} \times d}$, 
$b_\text{ia}\in \mathbb{R}^{N_\text{a}}$, 
$W_\text{io} \in \mathbb{R}^{N_\text{o} \times d}$, 
and $b_\text{io}\in \mathbb{R}^{N_\text{o}}$ are learnable parameters; 
$E_\text{a}$ maps the most likely action into the respective vectors according to the last predictions $p^\text{ia}_{t-1}$ using a dictionary with $N_\text{a}\times d$ learnable parameters; $E_\text{o}$ does the same for the object classes. The predicted $p^\text{ia}_{t}$ and $p^\text{io}_{t}$ are transferred to the input of these LSTMs at the next timestep and also inputted to the downstream components, the mask decoder and the action decoder. 

Now, as they do not need visual inputs, we can train the two LSTMs in a supervised fashion using the pairs of instructions and the corresponding ground truth action-object sequences. We denote this supervised loss, i.e., the sum of the losses for the two LSTMs, by $\mathcal{L}_{\text{aux}}$. Although it is independent of the environment and we can train the LSTMs offline, we simultaneously train them along with other components in the model by adding $\mathcal{L}_{\text{aux}}$ to the overall loss. We think this contributes to better learning of instruction representation 
$s_{m_t}$, which is also used by the mask decoder and the action decoder.

As mentioned above, we treat the navigation actions differently from the manipulation actions. There are three differences. First, we simplify the ground truth action sequence for the navigation actions if necessary. For instance, suppose an instruction ``{Turn left, go ahead to the counter and turn right}'' with a ground truth action sequence ``\texta{RotateLeft}, \texta{MoveAhead}, \texta{MoveAhead}, \texta{MoveAhead}, \texta{MoveAhead}, \texta{RotateRight}''. The repetition of \texta{MoveAhead} reflects the environment and cannot be predicted without visual inputs. Thus, by eliminating the repeated actions, we convert the sequence into the minimum-length one, ``\texta{RotateLeft}, \texta{MoveAhead}, \texta{RotateRight}'', and regard it as the ground truth sequence, training the instruction decoder. Second, as there is no 
accompanied object for the  navigation actions, we use the object-class sequence ``\texta{None, None, None}'' as the ground truth. Third, in the case of navigation actions, we do not transfer the outputs $p^\text{ia}_{t}$ and $p^\text{io}_{t}$ to the mask decoder and the action decoder and instead feed constant (but learnable) vectors $p^\text{ia}_{\text{nav}} \in \mathbb{R}^{N_\text{a}}$ and $p^\text{io}_{\text{nav}}  \in \mathbb{R}^{N_\text{o}}$ to them. 
As the instruction decoder learns to predict the minimum-length action sequences as above, providing such predictions will be harmful for the action decoder. We avoid this by feeding $p^\text{ia}_{\text{nav}}$ and $p^\text{io}_{\text{nav}}$.

\subsection{Action Decoder} \label{sec:action_decoder}

The action decoder receives four inputs and predicts the action at $t$. The inputs are as follows: the encoded instruction $s_{m_t}$, the output $p_t^\text{ia}$ and $p_t^\text{io}$ of the instruction decoder\footnote{These are replaced with $p_{\text{nav}}^\text{ia}$ and $p_{\text{nav}}^\text{ia}$ if $\mathrm{argmax}(p^\text{ia}_t)$ is not a manipulation action, as mention above.}
and aggregated feature $v_t$ of visual inputs, which will be described below.

\subsubsection{Hierarchical Attention over Visual Features} 

As explained in Sec.~\ref{sec:feat_representation}, we use the multi-view object-centric representation of visual inputs. To be specific, we aggregate $N\times K$ outputs of Mask R-CNN from $K$ ego-centric images, obtaining a single vector $v_t$. The Mask R-CNN outputs for view $k(=1,\ldots,K)$ are the visual features  $(v^k_{t,1},\ldots,v^k_{t,N})$ and 
the confidence scores 
$(\rho^k_{t,1},\ldots,\rho^k_{t,N})$ of $N$ detected objects.

To do this feature aggregation, we employ a hierarchical approach, where we first search for the objects relevant to the current instruction in each view and then merge the features over the views to a single feature vector. In the first step, we compute and apply soft-attentions over $N$ objects for each view. To be specific, we compute attention weights
${\alpha}^{k}_{\text{s}} \in \mathbb{R}^N$ across
$v^{k}_{t,1},\ldots,v^{k}_{t,N}$ 
guided by $s_{m_t}$ as
\begin{equation}
    \alpha^{k}_{\text{s},n} = \mathrm{softmax}(({v}^{k}_{t,n})^\top W^k_\text{s} s_{m_t}), \\
\end{equation}
where $W^k_\text{s} \in \mathbb{R}^{d \times d}$ is a learnable matrix, for $k=1,\ldots,K$. 
We then apply the weights to the $N$ visual features multiplied with their confidence scores for this view, yielding a single $d$-dimensional vector as 
\begin{equation}
    {v}^{k}_{t} = \sum_{n=1}^{N} {\alpha}^{k}_{\text{s},n} 
    {v}^{k}_{t,n} \rho^k_{t,n},
\end{equation}
where $\rho^k_{t,n}$ is the confidence score associated with
$v^k_{t,n}$.

In the second step, we merge the above features $v^1_t,\ldots,v^K_t$ using 
\emph{gated-attention}. We compute the weight $\alpha_{g}^k (\in \mathbb{R})$ 
of view $k(=1,\ldots,K)$ guided by $s_{m_t}$ as
\begin{equation}
\label{eq:gate-attn}
    \alpha^{k}_{\text{g}} = \mathrm{sigmoid}(({v}^{k}_{t})^\top W_\text{g} s_{m_t}),
\end{equation}
where $W_\text{g}\in \mathbb{R}^{d \times d}$ is a learnable matrix.
Finally, we apply the weights to $\{v^k_t\}_{k=1,\ldots,K}$ to have the visual feature $v_t \in \mathbb{R}^d$ as
\begin{equation}
    v_t = \sum_{k=1}^{K} {\alpha}^{k}_{\text{g}} {v}^{k}_{t}.
\end{equation}
As shown in the ablation test in the supplementary,
the performance drops significantly when replacing the above gated-attention by soft-attention, indicating the necessity for merging observations of different views, not selecting one of them. 

\subsubsection{Decoder Design}

The decoder predicts the action at $t$ from $v_t$, $s_{m_t}$, $p_t^\text{ia}$ and $p_t^\text{io}$. We employ an LSTM, which outputs the hidden state $h^\text{a}_{t}\in\mathbb{R}^d$ at $t$ from the previous state $h^\text{a}_{t-1}$ along with the above four inputs as
\begin{equation}\label{eq:action_pred}
    h^\text{a}_{t} = \mathrm{LSTM}([v_t; s_{m_t}; p^\text{ia}_{t}; p^\text{io}_{t}], h^\text{a}_{t-1}),
\end{equation}
where $[;]$ denotes concatenation operation. We initialize the LSTM by setting the initial hidden state $h^\text{a}_0$ to  $h_\text{G}$, the encoded feature of the goal statement; see Sec.~\ref{sec:feat_representation}. The updated state $h^\text{a}_{t}$ is fed into a fully-connected layer to yield the probabilities over the $N_\text{a} +1$ actions including \texttt{COMPLETE} as follows:
\begin{equation}
    p^\text{a}_t = \mathrm{softmax}(W_\text{a} h^\text{a}_{t}  + b_\text{a}),
\end{equation}
where $W_\text{a} \in \mathbb{R}^{(N_\text{a} + 1)\times d}$
and 
$b_\text{a} \in \mathbb{R}^{N_\text{a} + 1}$.
We choose the action with the maximum probability for the predicted action. In the training of the model, we use cross entropy loss $\mathcal{L}_\text{action}$ computed between $p^\text{a}_t$ and the one-hot representation of the true action.

\subsection{Mask Decoder} \label{sec:mask_decoder}

To predict the mask specifying an object to interact with, we utilize the object-centric representations 
$V^c_t=(v^c_{t,1},\ldots,v^c_{t,N})$
of the visual inputs of the central view ($k=c$).
Namely, we have only to select one of the $N$ detected objects. This enables more accurate specification of an object mask than predicting a class-agnostic binary mask as in the prior work \cite{shridhar2020alfred}.

To do this, we first apply simple self-attention to the visual features $V^c_t$, aiming at capturing the relation between objects in the central view. We employ the attention mechanism inside the light-weight Transformer with a single head proposed in \cite{nguyenefficient} for this purpose, obtaining $\bar{\mathcal{A}}_{V^c_t}(V^c_t) \in \mathbb{R}^{N\times d}$. We then apply linear transformation to $\bar{\mathcal{A}}_{V^c_t}(V^c_t)$ 
using a single fully-connected layer having weight $W\in \mathbb{R}^{N\times d}$ and bias $b\in \mathbb{R}^d$, 
with a residual connection as 
\begin{equation}
    \hat{V^c_t} = \mathrm{ReLU}(W \bar{\mathcal{A}}_{V^c_t}
    (V^c_t) +\mathbf{1}_{K} \cdot b^\top) + V^c_t,
\end{equation}
where $\mathbf{1}_K$ is  $K$-vector with all ones.

We then compute the probability $p^\text{m}_{t,n}$ of selecting $n$-th object from the $N$ candidates using the above self-attended object features along with other inputs $s_{m_t}$, $p^\text{ia}_{t}$, and $p^\text{io}_{t}$. We concatenate the latter three inputs into a vector $g^\text{m}_t = [s_{m_t}; p^\text{ia}_{t}; p^\text{io}_{t}]$ and then compute the probability as 
\begin{equation}\label{eq:mask_pred}
    p^\text{m}_{t,n} = \mathrm{sigmoid}((g^\text{m}_t)^\top  W_\text{m} \hat{v}^{c}_{t,n}),
\end{equation}
where $W_\text{m}\in\mathbb{R}^{d + N_\text{a} + N_\text{o} \times d}$ is a learnable matrix. We select the object mask with the highest probability (i.e., $\mathrm{argmax}_{n=1,\ldots,N}(p^\text{m}_{t,n})$) at inference time.
At training time, we first match the ground truth object mask with the object mask having the highest IoU. Then, we calculate the BCE loss $\mathcal{L}_\text{mask}$ between the two. 

\section{Experiments}

\subsection{Experimental Configuration}

\paragraph{Dataset.}

We follow the standard procedure of ALFRED; 25,743 language directives over 8,055 expert demonstration episodes are split into the training, validation, and test sets. The latter two are further divided into two splits, called {\em seen} and {\em unseen}, depending on whether the scenes are included in the training set. 

\paragraph{Evaluation metrics.}
Following ~\cite{shridhar2020alfred}, we report the standard metrics, i.e.,  the scores of Task Success Rate, denoted by \textbf{Task} and Goal Condition Success Rate, denoted by \textbf{Goal-Cond}. The \textbf{Goal-Cond} score is the ratio of goal conditions being completed at the end of an episode. The \textbf{Task} score is defined to be one if all the goal conditions are completed, and otherwise 0. Besides, each metric is accompanied by a path-length-weighted (PLW) score \cite{Anderson2018OnEO}, which measures the agent's efficiency by penalizing scores with the length of the action sequence.

\paragraph{Implementation details.}

\begin{table*}[t!]
    \centering
    \caption{{Task and Goal-Condition Success Rate.}
        For each metric, the corresponding path weighted metrics are given in (parentheses).
        The highest values per fold and metric are shown in \B{bold}.
    }
    \resizebox{1.00\textwidth}{!}{
        \begin{tabular}{@{}laarrcaarr@{}}
            \toprule
            \multicolumn{1}{l}{\multirow{3}{*}{Model}}
                             & \mcp{4}{\textbf{Validation}} & & \mcc{4}{\textbf{Test}} \\
                             
                             & \mcc{2}{\textit{Seen}}   & \mcc{2}{\textit{Unseen}}
                             & 
                             & \mcc{2}{\textit{Seen}}   & \mcc{2}{\textit{Unseen}}  \\
                             
                             & \multicolumn{1}{d}{Task} & \multicolumn{1}{d}{Goal-Cond} 
                             & \multicolumn{1}{c}{Task} & \multicolumn{1}{c}{Goal-Cond} 
                             & 
                             & \multicolumn{1}{d}{Task} & \multicolumn{1}{d}{Goal-Cond} 
                             & \multicolumn{1}{c}{Task} & \multicolumn{1}{c}{Goal-Cond} \\
            \cmidrule{1-5} \cmidrule{7-10}

            \multicolumn{1}{l}{Shridhar et al. \cite{shridhar2020alfred}}   & $3.70$ ($2.10$)    & $10.00$  ($7.00$)    & $0.00$ ($0.00$)   & $6.90$ ($5.10$)  & & $3.98$ ($2.02$)   & $9.42$ ($6.27$)   & $0.39$ ($0.80$) & $7.03$ ($4.26$) \\[1pt]            
            \multicolumn{1}{l}{Legg et al. \cite{legg2020eccv}}          &\multicolumn{1}{d}{-} & \multicolumn{1}{d}{-} & \multicolumn{1}{c}{-} & \multicolumn{1}{c}{-}     & & ${3.85}$ (${1.50}$)   & ${8.87}$ (${5.52}$)   & ${0.85}$ (${0.36}$) & ${7.68}$ (${4.31}$)        \\[1pt]
            \multicolumn{1}{l}{Singh et al. \cite{singh2020eccv}}        & ${4.50}$ (${2.20}$)    & ${12.20}$  (${8.10}$)    & ${0.70}$ (${0.30}$)   & ${9.50}$ (${6.10}$)        & & ${5.41}$ (${2.51}$)   & ${12.32}$ (${8.27}$)   & ${1.50}$ (${0.7}$) & ${8.08}$ (${5.20}$)        \\[1pt]
            \multicolumn{1}{l}{MOCA \cite{singh2020moca}}        & ${19.15}$ (${13.60}$)    & ${28.50}$  (${22.30}$)    & ${3.78}$ (${2.00}$)   & ${13.40}$ (${8.30}$)    & & ${22.05}$ (${15.10}$)   & ${28.29}$ (${22.05}$)   & ${5.30}$ (${2.72}$) & ${14.28}$ (${9.99}$) \\[1pt]
            
            \cmidrule{1-5} \cmidrule{7-10}
            \multicolumn{1}{l}{Ours (single view)}& ${18.90}$ (${13.90}$)    & ${26.80}$  (${21.90}$)     & ${3.90}$ (${2.50}$)    & ${15.30}$ (${10.90}$)      & & $15.20$ ($11.79$) & $23.95$ ($20.27$)   & $4.45$ ($2.37$) & $14.71$ ($10.88$) \\[1pt]
            \multicolumn{1}{l}{Ours (multiple views)}              & {$\B{33.70}$ ($\B{28.40}$)}    & $\B{43.10}$  ($\B{38.00}$)    & $\B{9.70}$ ($\B{7.30}$)    & $\B{23.10}$ ($\B{18.10}$)      & & $\B{29.16}$ ($\B{24.67}$) & $\B{38.82}$ ($\B{34.85}$)   & $\B{8.37}$ ($\B{5.06}$) & $\B{19.13}$ ($\B{14.81}$) \\[1pt]
            \multicolumn{1}{l}{Ours (winning entry)$^\diamond$}   & ${14.30}$ (${10.80}$)    & ${22.40}$  (${19.60}$)     & ${4.60}$ (${2.80}$)    & ${11.40}$ (${8.70}$)         & & $12.39$ ($8.20$) & $20.68$ ($18.79$)   & $4.45$ ($2.24$) & $12.34$ ($9.44$) \\[1pt]

            \cmidrule{1-5}\cmidrule{7-10}
            \multicolumn{1}{l}{Human}     & \multicolumn{1}{d}{-} & \multicolumn{1}{d}{-} & \multicolumn{1}{c}{-} & \multicolumn{1}{c}{-} & & \multicolumn{1}{d}{-} & \multicolumn{1}{d}{-} & $91.00$ ($85.80$) & $94.50$ ($87.60$) \\
            \bottomrule
        \end{tabular}
    }
    \label{tab:results}
\end{table*}

We use $K=5$ views:
the center view, \emph{up} and \emph{down} views with the elevation degrees of $\pm 15^{\circ}$, and \emph{left} and \emph{right} views with the angles of $\pm 90^{\circ}$.
We employ a Mask R-CNN model with ResNet-50 backbone that receives a $300 \times 300$ 
image and outputs $N = 32$ 
object candidates. 
We train it before training the proposed model with 800K frames and corresponding instance segmentation masks collected by replaying the expert demonstrations of the training set. We set the feature dimensionality $d=512$.
We train the model using imitation learning on the expert demonstrations by minimizing the following loss:
\begin{equation}
    \mathcal{L} = \mathcal{L}_\text{mask} + \mathcal{L}_\text{action} + 
    \mathcal{L}_\text{aux}.
\end{equation}
We use the Adam optimizer with an initial learning rate of $10^{-3}$, which is halved at epoch 5, 8, and 10, and a batch size of 32 for 15 epochs in total. We use a dropout with the dropout probability 0.2 for the both visual features and LSTM decoder hidden states.

\subsection{Experimental Results} \label{sec:result}

\begin{table}[t!]
    \centering
    \caption{
        {Sub-goal success rate.} All values are in percentage. The agent is evaluated on the Validation set. Highest values per fold are indicated in \B{bold}.
    }
    \resizebox{0.65\textwidth}{!}{
        \begin{tabular}{@{}ldcdcdcdc@{}}
        \toprule
        \multirow{2}{*}{Sub-goal}
                         & \multicolumn{2}{c}{\small\cite{shridhar2020alfred}}
                         & \multicolumn{2}{c}{\small\cite{singh2020moca}} 
                         & \multicolumn{2}{c}{\textbf{Ours}}
                         \\
                          \cmidrule(lr){2-3} \cmidrule(lr){4-5}  \cmidrule(lr){6-7}
                         & \multicolumn{1}{a}{Seen} & \multicolumn{1}{c}{Unseen} 
                         & \multicolumn{1}{a}{Seen} & \multicolumn{1}{c}{Unseen}
                         & \multicolumn{1}{a}{Seen} & \multicolumn{1}{c}{Unseen}
                         \\
                         
        \hline
        {Goto}   & $51$ & $22$ & ${54}$ & $ {32}$ & $\B{59}$ & $ \B{39}$ \\
        \hline
        {Pickup} & $32$ & $21$ & ${53}$ & $ {44}$ & $\B{84}$ & $ \B{79}$ \\
        {Put}    & ${81}$ & ${46}$ & $62$ & $39$  & $\B{82}$ & $ \B{66}$ \\
        {Slice}  & $25$ & $12$ & ${51}$ & ${55}$  & $\B{89}$ & $ \B{85}$ \\
        \hline
        {Cool}   & ${88}$ & ${92}$ & $87$ & $38$  & $\B{92}$ & $ \B{94}$ \\
        {Heat}   & ${85}$ & ${89}$ & $84$ & $86$  & $\B{99}$ & $ \B{95}$ \\
        {Clean}  & ${81}$ & $57$ & $79$ & $\B{71}$  & $\B{94}$ & $ {68}$ \\
        {Toggle} & $\B{100}$ & ${32}$ & $93$ & $11$ & ${99}$ & $ \B{66}$ \\
        \hline
        {Average}    & $68$ & $46$ & ${70}$ & ${47}$ &\B{87} & \B{74} \\
        \bottomrule
        \end{tabular}
    }
    \label{tab:res_subgoal}
\end{table}

Table~\ref{tab:results} shows the results. It is seen that our method shows significant improvement over the previous methods~\cite{shridhar2020alfred,legg2020eccv,singh2020eccv,singh2020moca} on all metrics. Our method also achieves better PLW (path length weighted) scores in all the metrics (indicated in the parentheses), showing its efficiency. Notably, our method attains \textbf{8.96\%} success rate on the unseen test split, improving approximately 20 times compared with the published result in \cite{shridhar2020alfred}. The higher success rate in the unseen scenes indicates its ability to generalize in novel environments. Detailed results for each of the seven task types are shown in the supplementary.

The preliminary version of our method won an international competition, whose performance is lower than the present version. It differs in that $(p^\text{ia}_{t}, p^\text{io}_{t})$ are not forwarded to the mask decoder and the action decoder and the number of Mask R-CNN's outputs is set to $N=20$. It is noted that even with a single view (i.e., $K=1$), our model still outperforms  ~\cite{shridhar2020alfred,legg2020eccv,singh2020eccv} in all the metrics.

\paragraph{Sub-goal success rate.}

Following~\cite{shridhar2020alfred}, we evaluate the performance on individual sub-goals.
Table~\ref{tab:res_subgoal} shows the results.
It is seen that our method shows higher success rates in almost all of  the sub-goal categories. 

\paragraph{Performance by Task Type}

Table \ref{tab:tasktype} shows the success rates across the 7 task types achieved by the existing methods including ours on the validation set of ALFRED. It is seen that our method outperforms others by a large margin in both seen and unseen environments. 

\begin{table}[!ht]
    \centering
    \resizebox{0.65\textwidth}{!}{
        \begin{tabular}{@{}ldcdcdr@{}}
        \toprule
        \multirow{2}{*}{Task-Type}
                         & \multicolumn{2}{c}{\small\cite{shridhar2020alfred}}
                         & \multicolumn{2}{c}{\small\cite{singh2020moca}} 
                         & \multicolumn{2}{c}{\textbf{Ours}}
                         \\
                          \cmidrule(lr){2-3} \cmidrule(lr){4-5}  \cmidrule(lr){6-7}
                         & \multicolumn{1}{a}{Seen} & \multicolumn{1}{c}{Unseen} 
                         & \multicolumn{1}{a}{Seen} & \multicolumn{1}{c}{Unseen}
                         & \multicolumn{1}{a}{Seen} & \multicolumn{1}{c}{Unseen}
                         \\
                         
        \midrule
        {Pick \& Place}     & $7.0$ & $0.0$             & ${29.5}$ & $ {5.0}$
        & $\B{40.1}$ & $ \B{13.0}$\\

        {Stack \& Place}    & $0.9$ & $0.0$             & $5.2$ & $1.8$      
        & 
        $\B{17.4}$ & $ \B{11.9}$ \\
        
        {Pick Two}          & ${0.8}$ & ${0.0}$         & $11.2$ & $1.1$
        & 
        $\B{21.8}$ & $ \B{1.1}$ \\
        
        {Clean \& Place}    & $1.8$ & $0.0$             & 22.3 & 2.4        
        & 
        $\B{40.2}$ & $ \B{15.0}$ \\
        
        {Heat \& Place}     & $1.9$ & ${0.0}$           & ${15.8}$ & ${2.7}$  
        & 
        $\B{41.2}$ & $ \B{9.6}$ \\
        
        {Cool \& Place}     & ${4.0}$ & $0.0$           & $26.1$ & $0.7$         
        & 
        $\B{40.0}$ & $ \B{13.8}$ \\
        
        {Examine}           & ${9.6}$ & ${0.0}$       & $20.2$ & $\B{13.2}$
        & 
        $\B{34.4}$ & $ {12.9}$ \\
        \midrule
        {Average}    & $3.7$ & $0.0$  & ${18.6}$ & ${3.8}$          
        & 
        \B{33.6} & \B{11.0} \\
        \bottomrule
        \end{tabular}
    }

    \caption{
        {Success rate across 7 task types.} All values are in percentages. The agent is evaluated on the validation set. Highest values per split 
        are indicated in \B{bold}.
    }
    \label{tab:tasktype}
\end{table}

\subsection{Ablation Study}
We conduct an ablation test to validate the effectiveness of the components by incrementally adding each component to the proposed model. The results are shown in Table \ref{tab:ablation_lwit}. 

\begin{table}[!ht]
    \centering
    \caption{
        {Ablation study for the components of the proposed model.}
        We report the success rate (Task score) on the validation seen and unseen splits. The {\xmark} mark denotes that a corresponding component is removed from the proposed model. 
    }
    
    \resizebox{0.8\textwidth}{!}{
        \begin{tabular}{@{}cccccc@{}}
        \toprule
        \multirow{3}{*}{Model} & \multicolumn{4}{c}{\bf Components} &  \multicolumn{1}{c}{\bf Validation} \\
        \cmidrule(r){2-5} \cmidrule(l){6-6}
        & {Instruction} & {Two-stage} & {Multi-view} & Mask  & \multicolumn{1}{c}{\multirow{2}{*}{Seen / Unseen}} \\
        & {Selection} & {Interpretation} & {Hier. Attn} & Decoder   \\

        \cmidrule(r){1-5} \cmidrule(l){6-6}
        
        1 & \xmark  & \xmark     &  \xmark    & \cmark & \multicolumn{1}{r}{2.8 / 0.5}  \\ 

        2 & \cmark  & \xmark     &       \xmark & \cmark & \multicolumn{1}{r}{12.9 / 2.9} \\  

        3 & \cmark  & \cmark     &  \xmark      & \cmark & \multicolumn{1}{r}{18.9 / 3.9} \\
        
        4 & \cmark  & \cmark     &  \xmark      & \xmark &  \multicolumn{1}{r}{3.8 / 0.7} \\

        5 & \cmark & \cmark     &       \cmark & \cmark &  \multicolumn{1}{r}{33.7 / 9.7} \\  
        \bottomrule
        \end{tabular}
    }
    \label{tab:ablation_lwit}
\end{table}

The model variants 1-4 use a single-view input ($K=1$); they do not use multi-view inputs and the hierarchical attention method. Model 1 further discards the instruction decoder by replacing it with the soft-attention-based approach \cite{shridhar2020alfred}, which yields a different language feature $s_{\text{att}}$ at each timestep. Accordingly, $p^\text{io}_t$ and $p^\text{ia}_t$ are not fed to the mask/action decoders; we use $g^\text{m}_t = [s_\text{att}; h^\text{a}_t]$. These changes will make the method almost unworkable. Model 2 retains only the instruction selection module, yielding $s_{m_t}$. It performs much better than Model 1. Model 3 has the instruction decoder, which feeds $p^\text{io}_t$ and $p^\text{ia}_t$ to the subsequent decoders. It performs better than Model 2 by a large margin, showing the effectiveness of the two-stage method. 

Model 4 replaces the mask decoder with the counterpart of the baseline method \cite{shridhar2020alfred}, which upsamples a concatenated vector $[g^m_t; v_t]$ by deconvolution layers. This change results in inaccurate mask prediction, yielding a considerable performance drop. Model 5 is the full model. The difference from Model 3 is the use of multi-view inputs with the hierarchical attention mechanism. It contributes to a notable performance improvement, validating its effectiveness. 

Table \ref{tab:full_ablation1} shows the full results of the ablation test  reported in the main paper. We also provide additional results in Table \ref{tab:full_ablation2} with different activation functions (i.e. {sigmoid} or {softmax}) in the second step of the proposed hierarchical attention mechanism, and with different $K$'s; $K$ is selected from 1 (only `center' view), 3 (`center', `left', and `right' views), or 5 (`center', `left', `right', `up', and `down' views)). The results show that the use of \emph{gated-attention} in Eq.(5) (of the main paper) is essential. We also confirm the number of views also affect the success rate.

\begin{table*}[ht]
    \centering
    \begin{small}
    \resizebox{1.0\textwidth}{!}{
        \begin{tabular}{@{}ccccrrrr@{}}
        \toprule
        \multicolumn{4}{c}{\bf Components} & \multicolumn{2}{c}{\bf Validation-Seen} & \multicolumn{2}{c}{\bf Validation-Unseen} \\
        \cmidrule(r){1-4} \cmidrule(lr){5-6} \cmidrule(l){7-8}
        {Instruction}  & {Two-stage} & {Multi-view} & {Mask} &
        \multicolumn{1}{c}{\multirow{2}{*}{Task}}   & \multicolumn{1}{c}{\multirow{2}{*}{Goal-Cond.}} &  \multicolumn{1}{c}{\multirow{2}{*}{Task}} &  \multicolumn{1}{c}{\multirow{2}{*}{Goal-Cond.}} \\
        {Selection} & {Interpretation} & {Hier. Attn} & {Decoder} & & & & \\

        \cmidrule(r){1-4} \cmidrule(lr){5-6} \cmidrule(l){7-8}
        \xmark & \xmark &     \xmark    & \cmark     & ${2.8}$ (${1.3}$)    & ${9.7}$  (${6.5}$)     & ${0.5}$ (${0.2}$)    & ${9.2}$ (${5.4}$)  \\ 
        
        \cmark &    \xmark  & \xmark     &  \cmark    & ${12.9}$ (${9.4}$)    & ${21.6}$  (${17.3}$)     & ${2.9}$ (${1.6}$)    & ${13.1}$ (${9.4}$) \\

        \cmark &    \cmark  & \xmark     &  \cmark      &${18.9}$ (${13.9}$)    & ${26.8}$  (${21.9}$)     & ${3.9}$ (${2.5}$)    & ${15.3}$ (${10.9}$) \\
        
        \cmark &    \cmark  & \cmark     &  \xmark      &${3.8}$ (${2.4}$)& ${14.9}$ (${11.2}$) & ${0.7}$ (${0.3}$)& ${10.4}$ (${6.9}$) \\ 
        
        \cmark & \cmark & \cmark     &       \cmark         & {${33.7}$ (${28.4}$)}    & ${43.1}$  (${38.0}$)    & ${9.7}$ (${7.3}$)    & ${23.1}$ (${18.1}$) \\ 
        \bottomrule
        \end{tabular}
    }
    \end{small}
    \caption{
        Results of an ablation test for examining the effectiveness of each component of the proposed model. The path weighted scores   are reported in the parentheses.
    }
    \label{tab:full_ablation1}
\end{table*}

\begin{table*}[ht]
    \centering
    \begin{small}
    \resizebox{1.0\textwidth}{!}{
        \begin{tabular}{@{}ccccrrrr@{}}
        \toprule
        \multicolumn{4}{c}{\multirow{2}{*}{\bf Configurations}} & \multicolumn{2}{c}{\bf Validation-Seen} & \multicolumn{2}{c}{\bf Validation-Unseen} \\
        \cmidrule(lr){5-6} \cmidrule(l){7-8}
        & & & & \multicolumn{1}{c}{{Task}}   & \multicolumn{1}{c}{{Goal-Cond.}} &  \multicolumn{1}{c}{{Task}} &  \multicolumn{1}{c}{{Goal-Cond.}} \\

        \cmidrule(r){1-4} \cmidrule(lr){5-6} \cmidrule(l){7-8}
        \multicolumn{2}{l}{\multirow{2}{*}{Activation Function}} &   \multicolumn{2}{c}{\hspace{0.5cm}Softmax}     &  ${11.9}$ (${9.3}$)    & ${20.8}$ (${17.3}$)  & ${4.1}$ (${2.2}$)    & ${14.0}$  (${10.2}$) \\ 
        
        & &  \multicolumn{2}{c}{\hspace{0.5cm}\textbf{Sigmoid}}    & {${33.7}$ (${28.4}$)}    & ${43.1}$  (${38.0}$)    & ${9.7}$ (${7.3}$)    & ${23.1}$ (${18.1}$) \\ 
        
        \cmidrule(r){1-4} \cmidrule(lr){5-6} \cmidrule(l){7-8}
        & &  \multicolumn{2}{c}{\hspace{0.5cm}center}      &${18.9}$ (${13.9}$)    & ${26.8}$  (${21.9}$)     & ${3.9}$ (${2.5}$)    & ${15.3}$ (${10.9}$) \\
        
        \multicolumn{2}{l}{{Ego-centric views}}  & \multicolumn{2}{c}{\hspace{0.5cm}center, left, right}      &${25.9}$ (${21.2}$)& ${34.4}$ (${30.0}$) & ${6.2}$ (${3.8}$)& ${17.0}$ (${12.3}$) \\ 
        
        & & \multicolumn{2}{c}{\hspace{0.5cm}\textbf{center, left, right, up, down} \hspace{3cm}}       & {${33.7}$ (${28.4}$)}    & ${43.1}$  (${38.0}$)    & ${9.7}$ (${7.3}$)    & ${23.1}$ (${18.1}$) \\ 
        \bottomrule
        \end{tabular}
    }
    \end{small}
    \caption{
        Results of experiments comparing activation functions in the module for aggregating and encoding multi-view visual inputs. 
        The path weighted scores are reported in the parentheses.
    }
    \label{tab:full_ablation2}
\end{table*}

\subsection{Qualitative Results}

\subsubsection{Entire Task Completion}
\begin{figure*}[t]
\centering
\includegraphics[width=\linewidth]{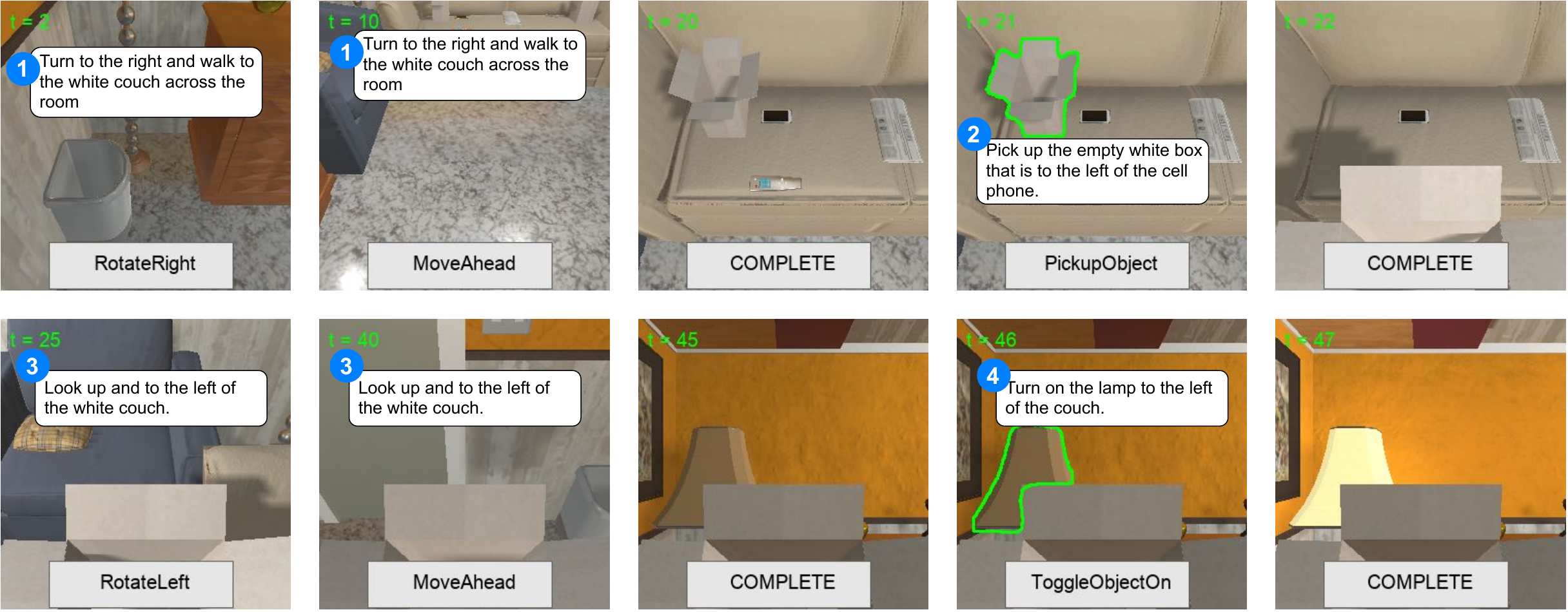}
\vspace{-0.1cm}
\caption{Our agent completes an \textbf{Examine} task ``\textit{Examine an empty box by the light of a floor lamp}" in an unseen environment.}
\label{fig:tasktype0}
\vspace{-0.1cm}
\end{figure*}

\begin{figure*}[t!]
  \includegraphics[width=\linewidth]{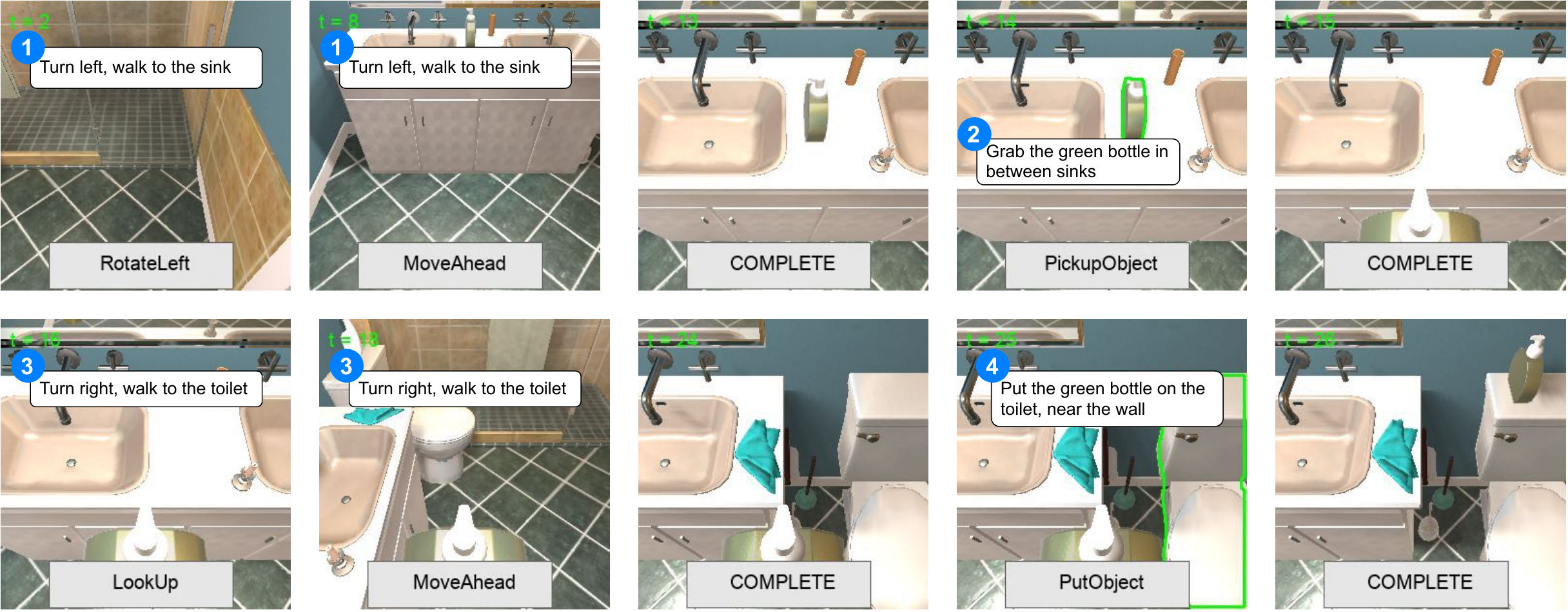}
\vspace{-0.1cm}
\caption{Our agent completes a \textbf{Pick} \& \textbf{Place} task ``\textit{Place the green bottle on the toilet basin}" in an unseen environment.}
\label{fig:tasktype1}
\vspace{-0.1cm}
\end{figure*}

\begin{figure*}[t]
  \includegraphics[width=\linewidth]{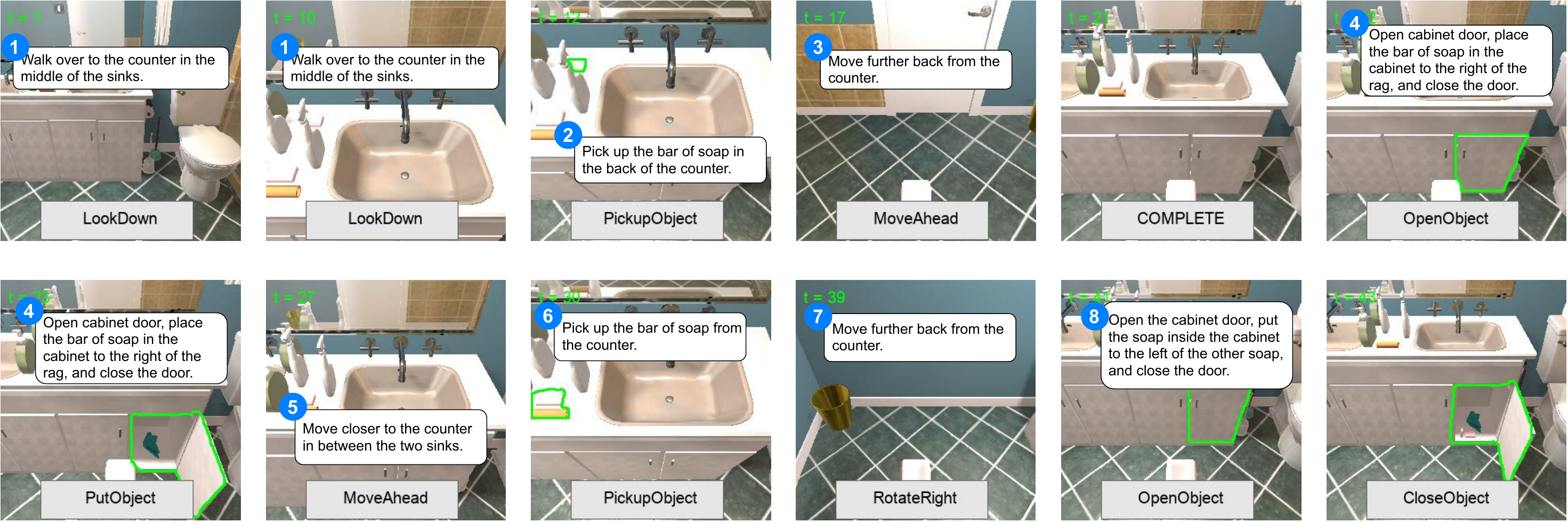}
\vspace{-0.1cm}
\caption{Our agent completes a \textbf{Pick Two} \& \textbf{Place} task ``\textit{To move two bars of soap to the cabinet}" in an unseen environment.}
\label{fig:tasktype2}
\vspace{-0.1cm}
\end{figure*}

\begin{figure*}[t]
  \includegraphics[width=\linewidth]{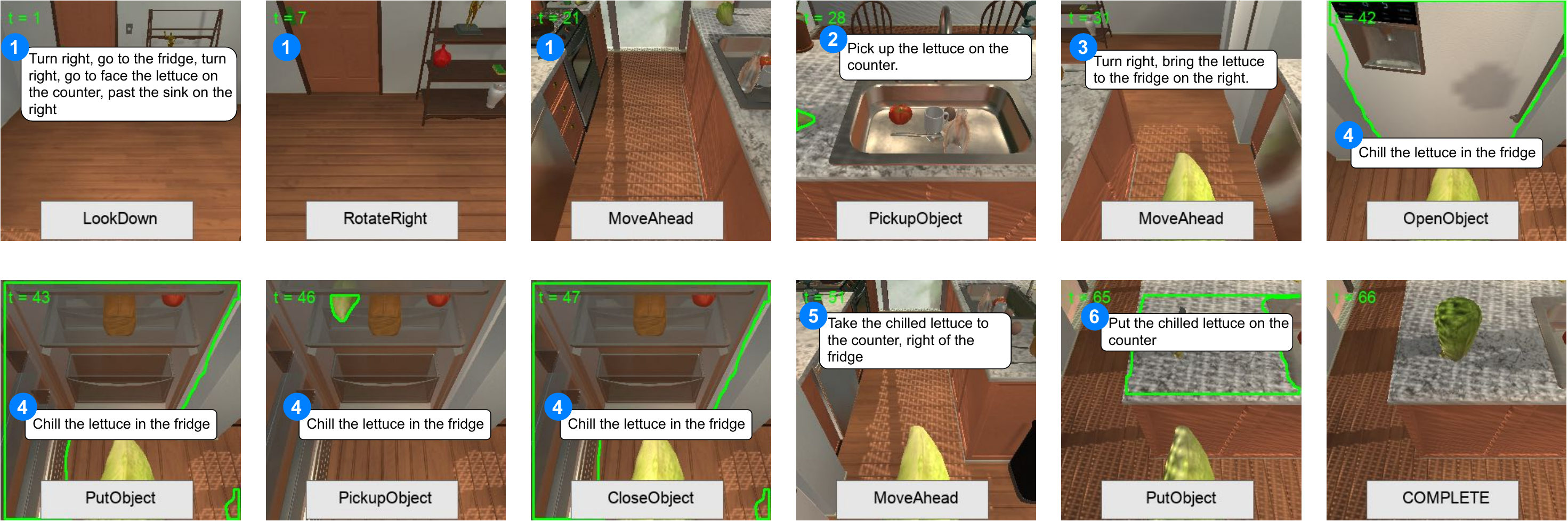}
\vspace{-0.1cm}
\caption{Our agent completes a \textbf{Cool} \& \textbf{Place} task ``\textit{Put chilled lettuce on the counter}" in an unseen environment.}
\label{fig:tasktype3}
\vspace{-0.1cm}
\end{figure*}

Figures \ref{fig:tasktype0}-\ref{fig:tasktype3} show the visualization of how the agent completes one of the seven types of tasks. These are the results for the unseen environment of the validation set. Each panel shows the agent's center view with the predicted action and object mask (if existing) at different time-steps. 

\begin{figure*}[tbh]
  \includegraphics[width=\linewidth]{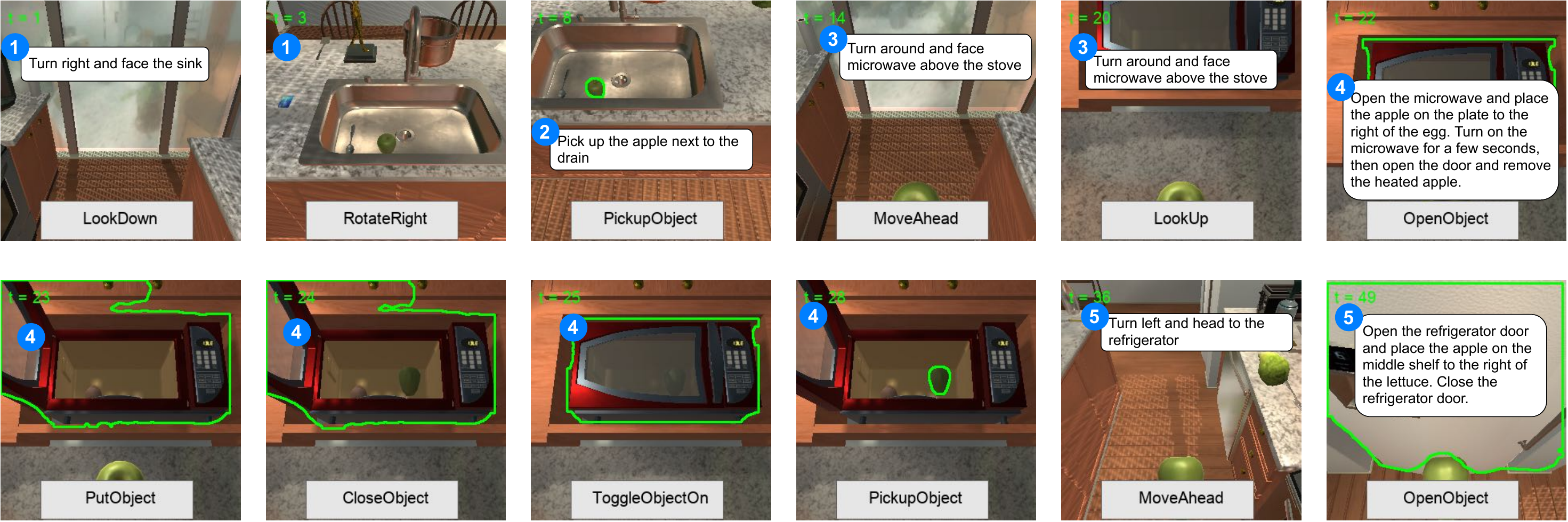}
\vspace{-0.1cm}
\caption{Our agent completes a \textbf{Heat} \& \textbf{Place} task ``\textit{Put a heated apple next to the lettuce on the middle shelf in the refrigerator}" in an unseen environment.}
\label{fig:tasktype4}
\vspace{-0.1cm}
\end{figure*}




\subsubsection{Mask Prediction for Sub-goal Completion}

\begin{figure}[!ht]
\centering
\begin{subfigure}{.4\columnwidth}
  \centering
  \includegraphics[width=\linewidth]{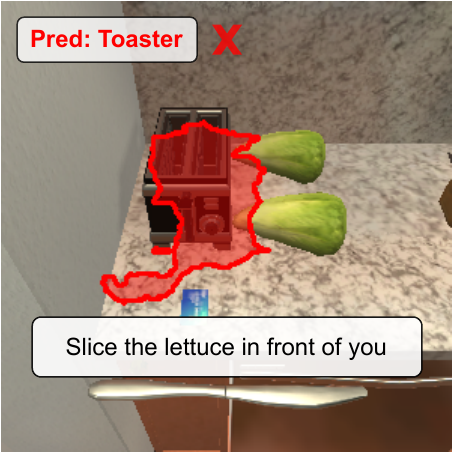}
  \caption{Shridhar et al. \cite{shridhar2020alfred}}
  \label{fig:sfig1}
\end{subfigure} \hfill 
\begin{subfigure}{.4\columnwidth}
  \centering
  \includegraphics[width=\linewidth]{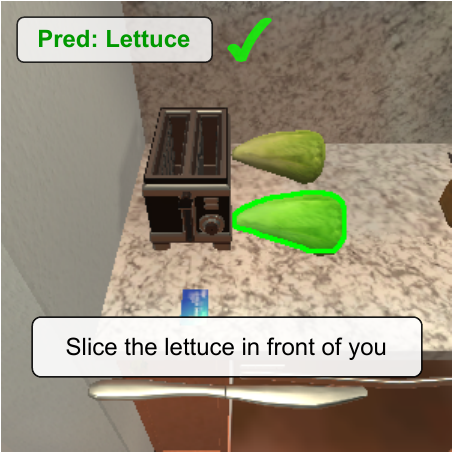}
  \caption{Ours}
  \label{fig:sfig2}
\end{subfigure}
\vspace{-0.1cm}
\caption{The prediction masks generated by  Shridhar \etal and our method where the agents are moved to the same location to accomplish {\bf Slice} sub-goal.}
\label{fig:vis_comparison}
\vspace{-0.1cm}
\end{figure}

Figure \ref{fig:vis_comparison} shows {
an example of the mask prediction} 
by the baseline \cite{shridhar2020alfred} and the proposed method.
It shows our method can predict a more accurate object mask when performing {\bf Slice} sub-goal. 
More examples are shown in the supplementary material.
Overall, our method shows better results, especially for difficult sub-goals like \textbf{Pickup}, \textbf{Put}, and \textbf{Clean}, for which a target object needs to be chosen from a wide range of candidates.

We also provide seven video clips as independent files, which contain several examples of the agent's entire task completion for seven above task instances in unseen environments. 

\section{Analyses of Failure Cases}

We analyze the failure cases of our method using the results on the validation splits.
We categorize them into navigation failures and manipulation failures. 

\subsection{Navigation Failures} 

It is seen from the sub-goal results of Table 2 in the main paper that the \textbf{Goto} sub-goal is the most challenging. Failures with it tend to make it hard to complete the entire goal, since they will inevitably affect the subsequent actions to take. We think there are three major cases for the navigation failures.  

The first case, which occurs most frequently, is that 
the agent follows a navigation instruction and reaches a position that should be fine as far as the instruction goes; nevertheless, it is not the right position for the next manipulation action to take. For instance, following the instruction ``Go to the table,'' the agent goes to the table. The next instruction is `` Pickup the remote control at the table,'' but the remote lies on the other side of the table. This is counted as a failure of completing the \text{Goto} sub-goal. 

The second case is when the instructions are either  abstract or misleading. An example is that when the agent has to take several left and right turns together with multiple \texttt{MoveAhead} steps to reach the destination, e.g., a drawer, the provided instruction is simply``Go to the drawer.''

The third case, which occurs less frequently, is that
while there is an obstacle in front of the agent, e.g., wall, it attempts to take the {\tt MoveAhead} action. This occurs because of the lack of proper visual inputs. 
This is demonstrated by the fact that when we reduce the number of views, the task success rate drops significantly, as shown in the second block of Table \ref{tab:full_ablation2}.

\subsection{Manipulation Failures} 
As shown in Table 2, after it has moved to the ideal position right before performing any interaction sub-goals (i.e., all the sub-goals but \textbf{Goto}), the agent can manipulates objects with high success rates of 91\% and 69\% in the seen and unseen environments, respectively. 
However, the success rates for completing the Goto sub-goal in the seen and unseen environments are only 59\% and 39\%, respectively. Therefore, the primary cause of the manipulation failures is that the agent cannot find the target object because it fails to reach the right destination due to a navigation failure. 


Even if the agent has successfully navigated to the right destination, it can fail to detect the target object. This seem to happen mostly because the object is either too small or indistinguishable from the surroundings. The agent tends to fail to detect, for example, a small knife placed on the steel/metal-made sink of the same color. 

The agent also fails to detect an object that has not seen in the training. This is confirmed by the fact that the performance drops considerably in unseen environments for some interaction sub-goals (including \textbf{Put}, \textbf{Clean}, and \textbf{Toggle}). 
There are also a small number of cases where failures are attributable to bad instructions, e.g., incorrect statement of objects. 
\section{Summary and Conclusion}

This chapter has presented a new method for interactive instruction following tasks and applied it to ALFRED. The method is built upon several new ideas, including the explicit selection of one of the provided instructions, the two-stage approach to the interpretation of each instruction (i.e., the instruction decoder), the employment of the object-centric representation of visual inputs obtained by hierarchical attention from multiple surrounding views (i.e., the action decoder), and the precise specification of objects to interact with based on the object-centric representation (i.e., the mask decoder). The experimental results have shown that the proposed method achieves superior performances in both seen and unseen environments compared with all the existing methods. We believe this study provides a useful baseline framework for future studies.

\cleardoublepage
\chapter{Conclusion} \label{chapter:ch6}
In this chapter, we summarize the contributions of our work and discuss some remaining problems as well as the future directions.

Over the past decades, we have experienced unprecedented progress in Artificial Intelligence, especially in Computer Vision and Natural Language Processing. Recently, there has been an increasing interest in solving problems that integrate visual and linguistic learning to unlock many practical applications potentially. In this dissertation, we have developed the models to understand visual and linguistic information and interact with humans and physical environments.

Concretely, in Chapter \ref{chapter:ch3}, we proposed a better and faster image captioning model. 
It is arguably believed that extracting good visual representations from the input image plays a crucial role in generating captions. We then pointed out that region-based features extracted by an object detector, such as Faster R-CNN, employed by previous methods, have some limitations: 1) lack of contextual information, 2) the risk of false detection, and 3) the high computational cost. We observed that grid-based features could alleviate the first two issues, while DETR-based detectors could help overcome the third issue. Thus, we introduced GRIT, which effectively utilizes the two visual features to generate better captions. Under the same training condition, we found that our proposed method outperforms previous methods significantly in terms of accuracy and speed. 

We believe that integrating the two visual features would also benefit other tasks in which the contextual information provides further clues for more accurate predictions. The tasks include but are not limited to several vision-language tasks (e.g., image-text retrieval, VQA, multi-modal verification, visual dialog, etc.), referring expression, scene graph generation, etc. 
Although this method may produce promising results for image captioning and maybe other tasks without vision-language pre-training, it is unclear how to utilize these two features in pre-training large models on large-scale datasets. First, using these two elements simultaneously complicates the model design. Second, the datasets with object detection annotations appearing marginal mitigate the advantage of region features compared to  pre-training simple grid-based models on large-scale datasets of image-text pairs. We hope future research will leverage the two visual features for pre-training on large-scale datasets and fine-tuning on downstream tasks.

In Chapter \ref{chapter:ch4}, we studied the visual dialog task, which requires agents to answer multiple questions in the form of dialog with humans. Previous methods considered attention from one input to another based on different hypotheses, such as ``question $\rightarrow$ history $\rightarrow$ image" path in \cite{Kang2019DualAN,lu2017best}, and ``question $\rightarrow$ image $\rightarrow$ history $\rightarrow$ question" path in \cite{Gan2019MultistepRV,wu2018you}, etc. 
These methods cannot take all the interactions between inputs into account. We argued that a plausible approach to solving the problem is effectively modeling all the interactions between inputs. Therefore, we proposed LTMI, a neural architecture that can efficiently deal with all the interactions between multiple such inputs in visual dialog. It treats all the utilities equally and simultaneously computes all their interactions. It has a design structure similar to Transformers, yet is much light-weight with less than one-tenth of the number of parameters. We found that the models built upon the proposed LTMI block achieve considerable performance improvement in Visual Dialog and Audio Visual Scene-aware Dialog with more inputs. 

It is worth mentioning that recent state-of-the-art methods \cite{wang2020vd,murahari2019large} for Visual Dialog employ pre-trained vision-language transformers. Specifically, these methods  concatenate the dialog history and the question into a unified text before forwarding them into the pre-trained transformers and fine-tuning for the task. We expect that placing additional LTMI layers on the top of transformer outputs may disentangle the inputs and capture their interactions for better prediction. Since the LTMI layer is light-weight, its introduction adds only minimal computation. We leave it for future research.

In Chapter \ref{chapter:ch5}, we tackled the ALFRED, an interactive instruction-following benchmark that requires performing household tasks by following human instructions. We pointed out that the previous methods that perform well on related tasks, Vision Language Navigation (VLN), fail to achieve good performance on ALFRED due to a number of challenging problems in ALFRED compared with VLN. We proposed LWIT to overcome the shortcomings of previous methods, which outperforms them by a large margin. We found several ideas that contribute to better performance: 1) selecting explicitly one instruction from a sequence of given instructions to accomplish at a time; 2) interpreting each instruction in two steps which refine the initial predictions; 3) employing region-based features obtained from multiple surrounding views and specifying objects to interact with rather than generating pixel-wise predictions. 

Despite significantly improving the ALFRED tasks, our method performs much worse than humans. 
However, the method serves as a strong baseline for ALFRED for further development. We strongly believe that several ideas, such as selecting masks instead of pixel-wise predictions and two-stage interpretation, are still applicable to future research to improve the accuracy. 
It is also noted that in this work, we assume that agents perform low-level actions (i.e., navigation and interaction actions with objects and environments) with perfect accuracy, which is beyond our research scope. We suspect that building a complete intelligent robot that works in real environments is much more complicated than simulation. Research on simulation-to-real embodied agents is still active and attracting attention from the community \cite{deitke2020robothor,bharadhwaj2019data,anderson2021sim}. 

%
%
\cleardoublepage
\bibliographystyle{unsrt}

\bibliography{References/references}

\begin{thebibliography}{100}

\bibitem{shridhar2020alfred}
M.~Shridhar, J.~Thomason, D.~Gordon, Y.~Bisk, W.~Han, R.~Mottaghi, L.~Zettlemoyer, and D.~Fox.
\newblock Alfred: A benchmark for interpreting grounded instructions for everyday tasks.
\newblock In {\em Proceedings of the IEEE Conference on Computer Vision and Pattern Recognition}, 2020.

\bibitem{lecun1998mnist}
Yann LeCun, L{\'e}on Bottou, Yoshua Bengio, Patrick Haffner, et~al.
\newblock Gradient-based learning applied to document recognition.
\newblock In {\em Proceedings of the IEEE}, 1998.

\bibitem{vaswani2017attention}
Ashish Vaswani, Noam Shazeer, Niki Parmar, Jakob Uszkoreit, Llion Jones, Aidan~N Gomez, {\L}ukasz Kaiser, and Illia Polosukhin.
\newblock Attention is all you need.
\newblock In {\em Advances in Neural Information Processing Systems}, pages 5998--6008, 2017.

\bibitem{krizhevsky2012imagenet}
Alex Krizhevsky, Ilya Sutskever, and Geoffrey~E Hinton.
\newblock Imagenet classification with deep convolutional neural networks.
\newblock In {\em Advances in Neural Information Processing Systems}, volume~25, 2012.

\bibitem{simonyan2014very}
Karen Simonyan and Andrew Zisserman.
\newblock Very deep convolutional networks for large-scale image recognition.
\newblock In {\em arXiv preprint arXiv:1409.1556}, 2014.

\bibitem{he2016deep}
Kaiming He, Xiangyu Zhang, Shaoqing Ren, and Jian Sun.
\newblock Deep residual learning for image recognition.
\newblock In {\em Proceedings of the IEEE Conference on Computer Vision and Pattern Recognition}, pages 770--778, 2016.

\bibitem{dosovitskiy2020image}
Alexey Dosovitskiy, Lucas Beyer, Alexander Kolesnikov, Dirk Weissenborn, Xiaohua Zhai, Thomas Unterthiner, Mostafa Dehghani, Matthias Minderer, Georg Heigold, Sylvain Gelly, et~al.
\newblock An image is worth 16x16 words: Transformers for image recognition at scale.
\newblock In {\em arXiv:2010.11929}, 2020.

\bibitem{ren2015faster}
Shaoqing Ren, Kaiming He, Ross Girshick, and Jian Sun.
\newblock Faster r-cnn: Towards real-time object detection with region proposal networks.
\newblock In {\em Advances in Neural Information Processing Systems}, pages 91--99, 2015.

\bibitem{he2017mask}
Kaiming He, Georgia Gkioxari, Piotr Doll{\'{a}}r, and Ross~B. Girshick.
\newblock Mask {R-CNN}.
\newblock In {\em Proceedings of the IEEE International Conference on Computer Vision}, pages 2980--2988. {IEEE} Computer Society, 2017.

\bibitem{deng2009imagenet}
Jia Deng, Wei Dong, Richard Socher, Li-Jia Li, Kai Li, and Li~Fei-Fei.
\newblock Imagenet: A large-scale hierarchical image database.
\newblock In {\em Proceedings of the Conference on Computer Vision and Pattern Recognition}, 2009.

\bibitem{jing:2019}
Longlong Jing and Yingli Tian.
\newblock Self-supervised visual feature learning with deep neural networks: {A} survey.
\newblock In {\em CoRR}, volume abs/1902.06162, 2019.

\bibitem{devlin2018bert}
Jacob Devlin, Ming-Wei Chang, Kenton Lee, and Kristina Toutanova.
\newblock Bert: Pre-training of deep bidirectional transformers for language understanding.
\newblock In {\em arXiv preprint arXiv:1810.04805}, 2018.

\bibitem{radford2018improving}
Alec Radford, Karthik Narasimhan, Tim Salimans, and Ilya Sutskever.
\newblock Improving language understanding by generative pre-training.
\newblock In {\em Technical report}. OpenAI, 2018.

\bibitem{brown2020language}
Tom~B Brown, Benjamin Mann, Nick Ryder, Melanie Subbiah, Jared Kaplan, Prafulla Dhariwal, Arvind Neelakantan, Pranav Shyam, Girish Sastry, Amanda Askell, et~al.
\newblock Language models are few-shot learners.
\newblock In {\em arXiv preprint arXiv:2005.14165}, 2020.

\bibitem{karpathy2014deep}
Andrej Karpathy, Armand Joulin, and Li~F Fei-Fei.
\newblock Deep fragment embeddings for bidirectional image sentence mapping.
\newblock In {\em Advances in Neural Information Processing Systems}, volume~27, 2014.

\bibitem{vinyals2015show}
Oriol Vinyals, Alexander Toshev, Samy Bengio, and Dumitru Erhan.
\newblock Show and tell: A neural image caption generator.
\newblock In {\em Proceedings of the IEEE Conference on Computer Vision and Pattern Recognition}, pages 3156--3164, 2015.

\bibitem{antol2015vqa}
Stanislaw Antol, Aishwarya Agrawal, Jiasen Lu, Margaret Mitchell, Dhruv Batra, C~Lawrence~Zitnick, and Devi Parikh.
\newblock Vqa: Visual question answering.
\newblock In {\em Proceedings of the IEEE International Conference on Computer Vision}, pages 2425--2433, 2015.

\bibitem{das2017visual}
Abhishek Das, Satwik Kottur, Khushi Gupta, Avi Singh, Deshraj Yadav, Jos{\'e}~MF Moura, Devi Parikh, and Dhruv Batra.
\newblock Visual dialog.
\newblock In {\em Proceedings of the IEEE Conference on Computer Vision and Pattern Recognition}, pages 326--335, 2017.

\bibitem{anderson2018vision}
P.~Anderson, Q.~Wu, D.~Teney, J.~Bruce, M.~Johnson, N.~S{\"u}nderhauf, I.~Reid, S.~Gould, and A.~van~den Hengel.
\newblock Vision-and-language navigation: Interpreting visually-grounded navigation instructions in real environments.
\newblock In {\em Proceedings of the IEEE Conference on Computer Vision and Pattern Recognition}, 2018.

\bibitem{fried2018speaker}
D.~Fried, R.~Hu, V.~Cirik, A.~Rohrbach, J.~Andreas, L.-P. Morency, T.~Berg-Kirkpatrick, K.~Saenko, D.~Klein, and T.~Darrell.
\newblock Speaker-follower models for vision-and-language navigation.
\newblock In {\em Advances in Neural Information Processing Systems}, 2018.

\bibitem{zhu2020vision}
F.~Zhu, Y.~Zhu, X.~Chang, and X.~Liang.
\newblock Vision-language navigation with self-supervised auxiliary reasoning tasks.
\newblock In {\em Proceedings of the IEEE Conference on Computer Vision and Pattern Recognition}, 2020.

\bibitem{mogadala:2015}
Aditya Mogadala.
\newblock Polylingual multimodal learning.
\newblock In {\em ECML PKDD Doctoral Consortium}, page 155. Citeseer, 2015.

\bibitem{ma2019selfmonitoring}
C.-Y. Ma, J.~Lu, Z.~Wu, G.~AlRegib, Z.~Kira, R.~Socher, and C.~Xiong.
\newblock Self-monitoring navigation agent via auxiliary progress estimation.
\newblock In {\em Proceedings of International Conference on Learning Representations}, 2019.

\bibitem{goodfellow2016deep}
Ian Goodfellow, Yoshua Bengio, Aaron Courville, and Yoshua Bengio.
\newblock {\em Deep learning}, volume~1.
\newblock MIT Press, 2016.

\bibitem{polyak1964some}
Boris~T Polyak.
\newblock Some methods of speeding up the convergence of iteration methods.
\newblock In {\em Ussr computational mathematics and mathematical physics}, volume~4, pages 1--17. Elsevier, 1964.

\bibitem{sutskever2013importance}
Ilya Sutskever, James Martens, George Dahl, and Geoffrey Hinton.
\newblock On the importance of initialization and momentum in deep learning.
\newblock In {\em Proceedings of International Conference on Machine Learning}, pages 1139--1147. PMLR, 2013.

\bibitem{kingma2014adam}
Diederik~P Kingma and Jimmy Ba.
\newblock Adam: A method for stochastic optimization.
\newblock In {\em arXiv preprint arXiv:1412.6980}, 2014.

\bibitem{duchi2011adaptive}
John Duchi, Elad Hazan, and Yoram Singer.
\newblock Adaptive subgradient methods for online learning and stochastic optimization.
\newblock In {\em Journal of machine learning research}, volume~12, 2011.

\bibitem{Hinton06}
Geoffrey~E. Hinton, Simon Osindero, and Yee~Whye Teh.
\newblock A fast learning algorithm for deep belief nets.
\newblock {\em Neural Computation}, 18:1527--1554, 2006.

\bibitem{rosenblatt1958perceptron}
Frank Rosenblatt.
\newblock The perceptron: a probabilistic model for information storage and organization in the brain.
\newblock {\em Psychological review}, 65(6):386, 1958.

\bibitem{rumelhart1985learning}
David~E Rumelhart, Geoffrey~E Hinton, and Ronald~J Williams.
\newblock Learning internal representations by error propagation.
\newblock Technical report, California Univ San Diego La Jolla Inst for Cognitive Science, 1985.

\bibitem{rumelhart1988learning}
David~E Rumelhart, Geoffrey~E Hinton, Ronald~J Williams, et~al.
\newblock Learning representations by back-propagating errors.
\newblock In {\em Cognitive modeling}, volume~5, page~1, 1988.

\bibitem{hochreiter1997long}
Sepp Hochreiter and J{\"u}rgen Schmidhuber.
\newblock Long short-term memory.
\newblock In {\em Neural computation}, volume~9, pages 1735--1780, 1997.

\bibitem{liu2021swin}
Ze~Liu, Yutong Lin, Yue Cao, Han Hu, Yixuan Wei, Zheng Zhang, Stephen Lin, and Baining Guo.
\newblock Swin transformer: Hierarchical vision transformer using shifted windows.
\newblock In {\em Proceedings of the IEEE International Conference on Computer Vision}, pages 10012--10022, 2021.

\bibitem{xu2015show}
Kelvin Xu, Jimmy Ba, Ryan Kiros, Kyunghyun Cho, Aaron Courville, Ruslan Salakhudinov, Rich Zemel, and Yoshua Bengio.
\newblock Show, attend and tell: Neural image caption generation with visual attention.
\newblock In {\em Proceedings of International Conference on Machine Learning}, pages 2048--2057, 2015.

\bibitem{rennie2017self}
Steven~J Rennie, Etienne Marcheret, Youssef Mroueh, Jerret Ross, and Vaibhava Goel.
\newblock Self-critical sequence training for image captioning.
\newblock In {\em Proceedings of the IEEE Conference on Computer Vision and Pattern Recognition}, pages 7008--7024, 2017.

\bibitem{lu2017knowing}
Jiasen Lu, Caiming Xiong, Devi Parikh, and Richard Socher.
\newblock Knowing when to look: Adaptive attention via a visual sentinel for image captioning.
\newblock In {\em Proceedings of the IEEE Conference on Computer Vision and Pattern Recognition}, pages 375--383, 2017.

\bibitem{anderson2018bottom}
Peter Anderson, Xiaodong He, Chris Buehler, Damien Teney, Mark Johnson, Stephen Gould, and Lei Zhang.
\newblock Bottom-up and top-down attention for image captioning and visual question answering.
\newblock In {\em Proceedings of the IEEE Conference on Computer Vision and Pattern Recognition}, pages 6077--6086, 2018.

\bibitem{luo2021dual}
Yunpeng Luo, Jiayi Ji, Xiaoshuai Sun, Liujuan Cao, Yongjian Wu, Feiyue Huang, Chia-Wen Lin, and Rongrong Ji.
\newblock Dual-level collaborative transformer for image captioning.
\newblock In {\em Proceedings of the AAAI Conference on Artificial Intelligence}, pages 2286--2293, 2021.

\bibitem{xian2022dual}
Tiantao Xian, Zhixin Li, Canlong Zhang, and Huifang Ma.
\newblock Dual global enhanced transformer for image captioning.
\newblock In {\em Neural Networks}, volume 148, pages 129--141, 2022.

\bibitem{carion2020end}
Nicolas Carion, Francisco Massa, Gabriel Synnaeve, Nicolas Usunier, Alexander Kirillov, and Sergey Zagoruyko.
\newblock End-to-end object detection with transformers.
\newblock In {\em Proceedings of the European Conference on Computer Vision}, pages 213--229, 2020.

\bibitem{zhu2021deformable}
Xizhou Zhu, Weijie Su, Lewei Lu, Bin Li, Xiaogang Wang, and Jifeng Dai.
\newblock Deformable detr: Deformable transformers for end-to-end object detection.
\newblock In {\em Proceedings of International Conference of Learning Representations}, 2021.

\bibitem{lin2014microsoft}
Tsung-Yi Lin, Michael Maire, Serge Belongie, James Hays, Pietro Perona, Deva Ramanan, Piotr Doll{\'a}r, and C~Lawrence Zitnick.
\newblock Microsoft coco: Common objects in context.
\newblock In {\em Proceedings of the European Conference on Computer Vision}, pages 740--755. Springer, 2014.

\bibitem{wang2021simvlm}
Zirui Wang, Jiahui Yu, Adams~Wei Yu, Zihang Dai, Yulia Tsvetkov, and Yuan Cao.
\newblock Simvlm: Simple visual language model pretraining with weak supervision.
\newblock In {\em arXiv:2108.10904}, 2021.

\bibitem{karpathy2015deep}
Andrej Karpathy and Li~Fei-Fei.
\newblock Deep visual-semantic alignments for generating image descriptions.
\newblock In {\em Proceedings of the IEEE Conference on Computer Vision and Pattern Recognition}, pages 3128--3137, 2015.

\bibitem{jiang2020defense}
Huaizu Jiang, Ishan Misra, Marcus Rohrbach, Erik Learned-Miller, and Xinlei Chen.
\newblock In defense of grid features for visual question answering.
\newblock In {\em Proceedings of the IEEE Conference on Computer Vision and Pattern Recognition}, pages 10267--10276, 2020.

\bibitem{zhang2021rstnet}
Xuying Zhang, Xiaoshuai Sun, Yunpeng Luo, Jiayi Ji, Yiyi Zhou, Yongjian Wu, Feiyue Huang, and Rongrong Ji.
\newblock Rstnet: Captioning with adaptive attention on visual and non-visual words.
\newblock In {\em Proceedings of the IEEE Conference on Computer Vision and Pattern Recognition}, pages 15465--15474, 2021.

\bibitem{fang2021you}
Yuxin Fang, Bencheng Liao, Xinggang Wang, Jiemin Fang, Jiyang Qi, Rui Wu, Jianwei Niu, and Wenyu Liu.
\newblock You only look at one sequence: Rethinking transformer in vision through object detection.
\newblock In {\em Advances in Neural Information Processing Systems}, 2021.

\bibitem{song2021vidt}
Hwanjun Song, Deqing Sun, Sanghyuk Chun, Varun Jampani, Dongyoon Han, Byeongho Heo, Wonjae Kim, and Ming-Hsuan Yang.
\newblock Vidt: An efficient and effective fully transformer-based object detector.
\newblock In {\em arXiv:2110.03921}, 2021.

\bibitem{xu2021e2e}
Haiyang Xu, Ming Yan, Chenliang Li, Bin Bi, Songfang Huang, Wenming Xiao, and Fei Huang.
\newblock E2e-vlp: End-to-end vision-language pre-training enhanced by visual learning.
\newblock In {\em arXiv:2106.01804}, 2021.

\bibitem{yang2019learning}
Xu~Yang, Hanwang Zhang, and Jianfei Cai.
\newblock Learning to collocate neural modules for image captioning.
\newblock In {\em Proceedings of the IEEE International Conference on Computer Vision}, pages 4250--4260, 2019.

\bibitem{li2019entangled}
Guang Li, Linchao Zhu, Ping Liu, and Yi~Yang.
\newblock Entangled transformer for image captioning.
\newblock In {\em Proceedings of the IEEE International Conference on Computer Vision}, pages 8928--8937, 2019.

\bibitem{huang2019attention}
Lun Huang, Wenmin Wang, Jie Chen, and Xiao-Yong Wei.
\newblock Attention on attention for image captioning.
\newblock In {\em Proceedings of the IEEE International Conference on Computer Vision}, pages 4634--4643, 2019.

\bibitem{pan2020x}
Yingwei Pan, Ting Yao, Yehao Li, and Tao Mei.
\newblock X-linear attention networks for image captioning.
\newblock In {\em Proceedings of the IEEE International Conference on Computer Vision}, pages 10971--10980, 2020.

\bibitem{cornia2020meshed}
Marcella Cornia, Matteo Stefanini, Lorenzo Baraldi, and Rita Cucchiara.
\newblock Meshed-memory transformer for image captioning.
\newblock In {\em Proceedings of the IEEE Conference on Computer Vision and Pattern Recognition}, pages 10578--10587, 2020.

\bibitem{herdade2019image}
Simao Herdade, Armin Kappeler, Kofi Boakye, and Joao Soares.
\newblock Image captioning: Transforming objects into words.
\newblock In {\em Advances in Neural Information Processing Systems}, 2019.

\bibitem{guo2020normalized}
Longteng Guo, Jing Liu, Xinxin Zhu, Peng Yao, Shichen Lu, and Hanqing Lu.
\newblock Normalized and geometry-aware self-attention network for image captioning.
\newblock In {\em Proceedings of the IEEE Conference on Computer Vision and Pattern Recognition}, pages 10327--10336, 2020.

\bibitem{zhang2021vinvl}
Pengchuan Zhang, Xiujun Li, Xiaowei Hu, Jianwei Yang, Lei Zhang, Lijuan Wang, Yejin Choi, and Jianfeng Gao.
\newblock Vinvl: Revisiting visual representations in vision-language models.
\newblock In {\em Proceedings of the IEEE Conference on Computer Vision and Pattern Recognition}, pages 5579--5588, 2021.

\bibitem{krishnavisualgenome}
Ranjay Krishna, Yuke Zhu, Oliver Groth, Justin Johnson, Kenji Hata, Joshua Kravitz, Stephanie Chen, Yannis Kalantidis, Li-Jia Li, David~A Shamma, Michael Bernstein, and Li~Fei-Fei.
\newblock {Visual Genome: Connecting Language and Vision Using Crowdsourced Dense Image Annotations}.
\newblock In {\em International Journal of Computer Vision}, volume 123, pages 32--73, 2017.

\bibitem{OpenImages}
Alina Kuznetsova, Hassan Rom, Neil Alldrin, Jasper Uijlings, Ivan Krasin, Jordi Pont-Tuset, Shahab Kamali, Stefan Popov, Matteo Malloci, Alexander Kolesnikov, Tom Duerig, and Vittorio Ferrari.
\newblock The open images dataset v4: Unified image classification, object detection, and visual relationship detection at scale.
\newblock In {\em International Journal of Computer Vision}, volume 128, pages 1956--1981, 2020.

\bibitem{shao2019objects365}
Shuai Shao, Zeming Li, Tianyuan Zhang, Chao Peng, Gang Yu, Xiangyu Zhang, Jing Li, and Jian Sun.
\newblock Objects365: A large-scale, high-quality dataset for object detection.
\newblock In {\em Proceedings of the IEEE International Conference on Computer Vision}, pages 8430--8439, 2019.

\bibitem{agrawal2019nocaps}
Harsh Agrawal, Karan Desai, Yufei Wang, Xinlei Chen, Rishabh Jain, Mark Johnson, Dhruv Batra, Devi Parikh, Stefan Lee, and Peter Anderson.
\newblock nocaps: novel object captioning at scale.
\newblock In {\em Proceedings of the IEEE International Conference on Computer Vision}, pages 8948--8957, 2019.

\bibitem{caesar2018cvpr}
Holger Caesar, Jasper Uijlings, and Vittorio Ferrari.
\newblock Coco-stuff: Thing and stuff classes in context.
\newblock In {\em Proceedings of the IEEE Conference on Computer Vision and Pattern Recognition}. IEEE, 2018.

\bibitem{karpathy}
Karpathy.
\newblock Karpathy/neuraltalk: Neuraltalk is a python+numpy project for learning multimodal recurrent neural networks that describe images with sentences.

\bibitem{achlioptas2021artemis}
Panos Achlioptas, Maks Ovsjanikov, Kilichbek Haydarov, Mohamed Elhoseiny, and Leonidas~J Guibas.
\newblock Artemis: Affective language for visual art.
\newblock In {\em Proceedings of the IEEE Conference on Computer Vision and Pattern Recognition}, pages 11569--11579, 2021.

\bibitem{papineni2002bleu}
Kishore Papineni, Salim Roukos, Todd Ward, and Wei-Jing Zhu.
\newblock Bleu: a method for automatic evaluation of machine translation.
\newblock In {\em Proceedings of the Annual Meeting of the Association for Computational Linguistics}, pages 311--318, 2002.

\bibitem{banerjee2005meteor}
Satanjeev Banerjee and Alon Lavie.
\newblock Meteor: An automatic metric for mt evaluation with improved correlation with human judgments.
\newblock In {\em Proceedings of the ACL Workshop on Intrinsic and Extrinsic Evaluation Measures for Machine Translation and/or Summarization}, pages 65--72, 2005.

\bibitem{lin2004rouge}
Chin-Yew Lin.
\newblock Rouge: A package for automatic evaluation of summaries.
\newblock In {\em Text Summarization Branches Out}, pages 74--81, 2004.

\bibitem{vedantam2015cider}
Ramakrishna Vedantam, C~Lawrence~Zitnick, and Devi Parikh.
\newblock Cider: Consensus-based image description evaluation.
\newblock In {\em Proceedings of the IEEE Conference on Computer Vision and Pattern Recognition}, pages 4566--4575, 2015.

\bibitem{anderson2016spice}
Peter Anderson, Basura Fernando, Mark Johnson, and Stephen Gould.
\newblock Spice: Semantic propositional image caption evaluation.
\newblock In {\em Proceedings of the European Conference on Computer Vision}, pages 382--398, 2016.

\bibitem{spacy2}
Matthew Honnibal and Ines Montani.
\newblock {spaCy 2}: Natural language understanding with {B}loom embeddings, convolutional neural networks and incremental parsing.
\newblock To appear, 2017.

\bibitem{kingma2015adam}
Diederik~P Kingma and Jimmy Ba.
\newblock {Adam: A Method for Stochastic Optimization}.
\newblock In {\em Proceedings of International Conference on Representation Learning}, 2015.

\bibitem{zhou2020unified}
Luowei Zhou, Hamid Palangi, Lei Zhang, Houdong Hu, Jason Corso, and Jianfeng Gao.
\newblock Unified vision-language pre-training for image captioning and vqa.
\newblock In {\em Proceedings of the AAAI Conference on Artificial Intelligence}, pages 13041--13049, 2020.

\bibitem{li2020oscar}
Xiujun Li, Xi~Yin, Chunyuan Li, Pengchuan Zhang, Xiaowei Hu, Lei Zhang, Lijuan Wang, Houdong Hu, Li~Dong, Furu Wei, et~al.
\newblock Oscar: Object-semantics aligned pre-training for vision-language tasks.
\newblock In {\em Proceedings of the European Conference on Computer Vision}, pages 121--137, 2020.

\bibitem{yao2017boosting}
Ting Yao, Yingwei Pan, Yehao Li, Zhaofan Qiu, and Tao Mei.
\newblock Boosting image captioning with attributes.
\newblock In {\em Proceedings of the IEEE International Conference on Computer Vision}, pages 4894--4902, 2017.

\bibitem{ke2019reflective}
Lei Ke, Wenjie Pei, Ruiyu Li, Xiaoyong Shen, and Yu-Wing Tai.
\newblock Reflective decoding network for image captioning.
\newblock In {\em Proceedings of the IEEE International Conference on Computer Vision}, pages 8888--8897, 2019.

\bibitem{yao2018exploring}
Ting Yao, Yingwei Pan, Yehao Li, and Tao Mei.
\newblock Exploring visual relationship for image captioning.
\newblock In {\em Proceedings of the European Conference on Computer Vision}, pages 684--699, 2018.

\bibitem{qin2019look}
Yu~Qin, Jiajun Du, Yonghua Zhang, and Hongtao Lu.
\newblock Look back and predict forward in image captioning.
\newblock In {\em Proceedings of the IEEE Conference on Computer Vision and Pattern Recognition}, pages 8367--8375, 2019.

\bibitem{yang2019auto}
Xu~Yang, Kaihua Tang, Hanwang Zhang, and Jianfei Cai.
\newblock Auto-encoding scene graphs for image captioning.
\newblock In {\em Proceedings of the IEEE Conference on Computer Vision and Pattern Recognition}, pages 10685--10694, 2019.

\bibitem{ji2021improving}
Jiayi Ji, Yunpeng Luo, Xiaoshuai Sun, Fuhai Chen, Gen Luo, Yongjian Wu, Yue Gao, and Rongrong Ji.
\newblock Improving image captioning by leveraging intra-and inter-layer global representation in transformer network.
\newblock In {\em Proceedings of the AAAI Conference on Artificial Intelligence}, pages 1655--1663, 2021.

\bibitem{fan2021tcic}
Zhihao Fan, Zhongyu Wei, Siyuan Wang, Ruize Wang, Zejun Li, Haijun Shan, and Xuanjing Huang.
\newblock Tcic: Theme concepts learning cross language and vision for image captioning.
\newblock In {\em arXiv:2106.10936}, 2021.

\bibitem{wang2019hierarchical}
Weixuan Wang, Zhihong Chen, and Haifeng Hu.
\newblock Hierarchical attention network for image captioning.
\newblock In {\em Proceedings of the AAAI Conference on Artificial Intelligence}, pages 8957--8964, 2019.

\bibitem{mathews2016senticap}
Alexander Mathews, Lexing Xie, and Xuming He.
\newblock Senticap: Generating image descriptions with sentiments.
\newblock In {\em Proceedings of the AAAI conference on artificial intelligence}, pages 3574–--3580, 2016.

\bibitem{lu2018neural}
Jiasen Lu, Jianwei Yang, Dhruv Batra, and Devi Parikh.
\newblock Neural baby talk.
\newblock In {\em Proceedings of the IEEE Conference on Computer Vision and Pattern Recognition}, pages 7219--7228, 2018.

\bibitem{kim2021vilt}
Wonjae Kim, Son Bokyung, Kim Ildoo, and Wonjae Kim.
\newblock Vilt: Vision-and-language transformer without convolution or region supervision.
\newblock In {\em Proceedings of International Conference on Machine Learning}, 2021.

\bibitem{Schwartz2019FactorGA}
Idan Schwartz, Seunghak Yu, Tamir Hazan, and Alexander~G Schwing.
\newblock Factor graph attention.
\newblock In {\em Proceedings of the IEEE Conference on Computer Vision and Pattern Recognition}, pages 2039--2048, 2019.

\bibitem{Kang2019DualAN}
Gi-Cheon Kang, Jaeseo Lim, and Byoung-Tak Zhang.
\newblock Dual attention networks for visual reference resolution in visual dialog.
\newblock In {\em Proceedings of the Conference on Empirical Methods in Natural Language Processing}, pages 2024--2033, 2019.

\bibitem{lu2017best}
Jiasen Lu, Anitha Kannan, Jianwei Yang, Devi Parikh, and Dhruv Batra.
\newblock Best of both worlds: Transferring knowledge from discriminative learning to a generative visual dialog model.
\newblock In {\em Advances in Neural Information Processing Systems}, pages 314--324, 2017.

\bibitem{Gan2019MultistepRV}
Zhe Gan, Yu~Cheng, Ahmed~El Kholy, Linjie Li, Jingjing Liu, and Jianfeng Gao.
\newblock Multi-step reasoning via recurrent dual attention for visual dialog.
\newblock In {\em Proceedings of the Conference of the Association for Computational Linguistics}, pages 6463--6474, 2019.

\bibitem{wu2018you}
Qi~Wu, Peng Wang, Chunhua Shen, Ian Reid, and Anton van~den Hengel.
\newblock Are you talking to me? reasoned visual dialog generation through adversarial learning.
\newblock In {\em Proceedings of the IEEE Conference on Computer Vision and Pattern Recognition}, pages 6106--6115, 2018.

\bibitem{chen2019uniter}
Yen-Chun Chen, Linjie Li, Licheng Yu, Ahmed~El Kholy, Faisal Ahmed, Zhe Gan, Yu~Cheng, and Jingjing Liu.
\newblock Uniter: Learning universal image-text representations.
\newblock In {\em arXiv preprint arXiv:1909.11740}, 2019.

\bibitem{gao2019dynamic}
Peng Gao, Zhengkai Jiang, Haoxuan You, Pan Lu, Steven~CH Hoi, Xiaogang Wang, and Hongsheng Li.
\newblock Dynamic fusion with intra-and inter-modality attention flow for visual question answering.
\newblock In {\em Proceedings of the IEEE Conference on Computer Vision and Pattern Recognition}, pages 6639--6648, 2019.

\bibitem{li2019visualbert}
Liunian~Harold Li, Mark Yatskar, Da~Yin, Cho-Jui Hsieh, and Kai-Wei Chang.
\newblock Visualbert: A simple and performant baseline for vision and language.
\newblock In {\em arXiv preprint arXiv:1908.03557}, 2019.

\bibitem{lu2019vilbert}
Jiasen Lu, Dhruv Batra, Devi Parikh, and Stefan Lee.
\newblock Vilbert: Pretraining task-agnostic visiolinguistic representations for vision-and-language tasks.
\newblock In {\em arXiv preprint arXiv:1908.02265}, 2019.

\bibitem{yu2019deep}
Zhou Yu, Jun Yu, Yuhao Cui, Dacheng Tao, and Qi~Tian.
\newblock Deep modular co-attention networks for visual question answering.
\newblock In {\em Proceedings of the IEEE Conference on Computer Vision and Pattern Recognition}, pages 6281--6290, 2019.

\bibitem{chen2015abc}
Kan Chen, Jiang Wang, Liang-Chieh Chen, Haoyuan Gao, Wei Xu, and Ram Nevatia.
\newblock Abc-cnn: An attention based convolutional neural network for visual question answering.
\newblock In {\em arXiv preprint arXiv:1511.05960}, 2015.

\bibitem{ilievski2016focused}
Ilija Ilievski, Shuicheng Yan, and Jiashi Feng.
\newblock A focused dynamic attention model for visual question answering.
\newblock In {\em arXiv preprint arXiv:1604.01485}, 2016.

\bibitem{kim2018bilinear}
Jin-Hwa Kim, Jaehyun Jun, and Byoung-Tak Zhang.
\newblock Bilinear attention networks.
\newblock In {\em Advances in Neural Information Processing Systems}, pages 1564--1574, 2018.

\bibitem{lu2016hierarchical}
Jiasen Lu, Jianwei Yang, Dhruv Batra, and Devi Parikh.
\newblock Hierarchical question-image co-attention for visual question answering.
\newblock In {\em Advances in Neural Information Processing Systems}, pages 289--297, 2016.

\bibitem{nguyen2018improved}
Duy-Kien Nguyen and Takayuki Okatani.
\newblock Improved fusion of visual and language representations by dense symmetric co-attention for visual question answering.
\newblock In {\em Proceedings of the IEEE Conference on Computer Vision and Pattern Recognition}, pages 6087--6096, 2018.

\bibitem{yang2016stacked}
Zichao Yang, Xiaodong He, Jianfeng Gao, Li~Deng, and Alex Smola.
\newblock Stacked attention networks for image question answering.
\newblock In {\em Proceedings of the IEEE Conference on Computer Vision and Pattern Recognition}, pages 21--29, 2016.

\bibitem{yu2017multi}
Zhou Yu, Jun Yu, Jianping Fan, and Dacheng Tao.
\newblock Multi-modal factorized bilinear pooling with co-attention learning for visual question answering.
\newblock In {\em Proceedings of the IEEE International Conference on Computer Vision}, pages 1821--1830, 2017.

\bibitem{yu2018beyond}
Zhou Yu, Jun Yu, Chenchao Xiang, Jianping Fan, and Dacheng Tao.
\newblock Beyond bilinear: Generalized multimodal factorized high-order pooling for visual question answering.
\newblock In {\em IEEE Transactions on Neural Networks and Learning Systems}, volume~29, pages 5947--5959. IEEE, 2018.

\bibitem{deng2018visual}
Chaorui Deng, Qi~Wu, Qingyao Wu, Fuyuan Hu, Fan Lyu, and Mingkui Tan.
\newblock Visual grounding via accumulated attention.
\newblock In {\em Proceedings of the IEEE Conference on Computer Vision and Pattern Recognition}, pages 7746--7755, 2018.

\bibitem{yu2018mattnet}
Licheng Yu, Zhe Lin, Xiaohui Shen, Jimei Yang, Xin Lu, Mohit Bansal, and Tamara~L Berg.
\newblock Mattnet: Modular attention network for referring expression comprehension.
\newblock In {\em Proceedings of the IEEE Conference on Computer Vision and Pattern Recognition}, pages 1307--1315, 2018.

\bibitem{zhuang2018parallel}
Bohan Zhuang, Qi~Wu, Chunhua Shen, Ian Reid, and Anton van~den Hengel.
\newblock Parallel attention: A unified framework for visual object discovery through dialogs and queries.
\newblock In {\em Proceedings of the IEEE Conference on Computer Vision and Pattern Recognition}, pages 4252--4261, 2018.

\bibitem{tan2019lxmert}
Hao Tan and Mohit Bansal.
\newblock Lxmert: Learning cross-modality encoder representations from transformers.
\newblock In {\em Proceedings of the Conference on Empirical Methods in Natural Language Processing}, 2019.

\bibitem{chen2015microsoft}
Xinlei Chen, Hao Fang, Tsung-Yi Lin, Ramakrishna Vedantam, Saurabh Gupta, Piotr Doll{\'a}r, and C~Lawrence Zitnick.
\newblock Microsoft coco captions: Data collection and evaluation server.
\newblock In {\em arXiv preprint arXiv:1504.00325}, 2015.

\bibitem{sharma2018conceptual}
Piyush Sharma, Nan Ding, Sebastian Goodman, and Radu Soricut.
\newblock Conceptual captions: A cleaned, hypernymed, image alt-text dataset for automatic image captioning.
\newblock In {\em Proceedings of the Annual Meeting of the Association for Computational Linguistics}, pages 2556--2565, 2018.

\bibitem{de2017guesswhat}
Harm De~Vries, Florian Strub, Sarath Chandar, Olivier Pietquin, Hugo Larochelle, and Aaron Courville.
\newblock Guesswhat?! visual object discovery through multi-modal dialogue.
\newblock In {\em Proceedings of the IEEE Conference on Computer Vision and Pattern Recognition}, pages 5503--5512, 2017.

\bibitem{kottur2019clevr}
Satwik Kottur, Jos{\'e}~MF Moura, Devi Parikh, Dhruv Batra, and Marcus Rohrbach.
\newblock Clevr-dialog: A diagnostic dataset for multi-round reasoning in visual dialog.
\newblock In {\em arXiv preprint arXiv:1903.03166}, 2019.

\bibitem{sutskever2014sequence}
Ilya Sutskever, Oriol Vinyals, and Quoc~V Le.
\newblock Sequence to sequence learning with neural networks.
\newblock In {\em Advances in Neural Information Processing Systems}, pages 3104--3112, 2014.

\bibitem{Guo_2019_CVPR}
Dalu Guo, Chang Xu, and Dacheng Tao.
\newblock Image-question-answer synergistic network for visual dialog.
\newblock In {\em Proceedings of the IEEE Conference on Computer Vision and Pattern Recognition}, pages 10434--10443, 2019.

\bibitem{kim2020modality}
Hyounghun Kim, Hao Tan, and Mohit Bansal.
\newblock Modality-balanced models for visual dialogue.
\newblock In {\em arXiv preprint arXiv:2001.06354}, 2020.

\bibitem{Niu_2019_CVPR}
Yulei Niu, Hanwang Zhang, Manli Zhang, Jianhong Zhang, Zhiwu Lu, and Ji-Rong Wen.
\newblock Recursive visual attention in visual dialog.
\newblock In {\em Proceedings of the IEEE Conference on Computer Vision and Pattern Recognition}, pages 6679--6688, 2019.

\bibitem{kottur2018visual}
Satwik Kottur, Jos{\'e}~MF Moura, Devi Parikh, Dhruv Batra, and Marcus Rohrbach.
\newblock Visual coreference resolution in visual dialog using neural module networks.
\newblock In {\em Proceedings of the European Conference on Computer Vision}, pages 153--169, 2018.

\bibitem{seo2017visual}
Paul~Hongsuck Seo, Andreas Lehrmann, Bohyung Han, and Leonid Sigal.
\newblock Visual reference resolution using attention memory for visual dialog.
\newblock In {\em Advances in Neural Information Processing Systems}, pages 3719--3729, 2017.

\bibitem{Zhang2019}
Heming Zhang, Shalini Ghosh, Larry Heck, Stephen Walsh, Junting Zhang, Jie Zhang, and C-C~Jay Kuo.
\newblock Generative visual dialogue system via weighted likelihood estimation.
\newblock In {\em Proceedings of the International Joint Conference on Artificial Intelligence}, pages 1025--1031, 2019.

\bibitem{Zheng2019ReasoningVD}
Zilong Zheng, Wenguan Wang, Siyuan Qi, and Song-Chun Zhu.
\newblock Reasoning visual dialogs with structural and partial observations.
\newblock In {\em Proceedings of the IEEE Conference on Computer Vision and Pattern Recognition}, pages 6669--6678, 2019.

\bibitem{chattopadhyay2017evaluating}
Prithvijit Chattopadhyay, Deshraj Yadav, Viraj Prabhu, Arjun Chandrasekaran, Abhishek Das, Stefan Lee, Dhruv Batra, and Devi Parikh.
\newblock Evaluating visual conversational agents via cooperative human-ai games.
\newblock In {\em Proceedings of AAAI Conference on Human Computation and Crowdsourcing}, 2017.

\bibitem{das2017learning}
Abhishek Das, Satwik Kottur, Jos{\'e}~MF Moura, Stefan Lee, and Dhruv Batra.
\newblock Learning cooperative visual dialog agents with deep reinforcement learning.
\newblock In {\em Proceedings of the IEEE International Conference on Computer Vision}, pages 2951--2960, 2017.

\bibitem{Yang2019MakingHM}
Tianhao Yang, Zheng-Jun Zha, and Hanwang Zhang.
\newblock Making history matter: History-advantage sequence training for visual dialog.
\newblock In {\em Proceedings of the IEEE International Conference on Computer Vision}, pages 2561--2569, 2019.

\bibitem{wang2020vd}
Yue Wang, Shafiq Joty, Michael~R Lyu, Irwin King, Caiming Xiong, and Steven~CH Hoi.
\newblock Vd-bert: A unified vision and dialog transformer with bert.
\newblock In {\em arXiv preprint arXiv:2004.13278}, 2020.

\bibitem{murahari2019large}
Vishvak Murahari, Dhruv Batra, Devi Parikh, and Abhishek Das.
\newblock Large-scale pretraining for visual dialog: A simple state-of-the-art baseline.
\newblock In {\em arXiv preprint arXiv:1912.02379}, 2019.

\bibitem{qi2020two}
Jiaxin Qi, Yulei Niu, Jianqiang Huang, and Hanwang Zhang.
\newblock Two causal principles for improving visual dialog.
\newblock In {\em Proceedings of the IEEE Conference on Computer Vision and Pattern Recognition}, pages 10860--10869, 2020.

\bibitem{krishna2017visual}
Ranjay Krishna, Yuke Zhu, Oliver Groth, Justin Johnson, Kenji Hata, Joshua Kravitz, Stephanie Chen, Yannis Kalantidis, Li-Jia Li, David~A Shamma, et~al.
\newblock Visual genome: Connecting language and vision using crowdsourced dense image annotations.
\newblock In {\em International Journal of Computer Vision}, volume 123, pages 32--73. Springer, 2017.

\bibitem{ba2016layer}
Jimmy~Lei Ba, Jamie~Ryan Kiros, and Geoffrey~E Hinton.
\newblock Layer normalization.
\newblock In {\em arXiv preprint arXiv:1607.06450}, 2016.

\bibitem{pennington2014glove}
Jeffrey Pennington, Richard Socher, and Christopher Manning.
\newblock Glove: Global vectors for word representation.
\newblock In {\em Proceedings of the Conference on Empirical Methods in Natural Language Processing}, pages 1532--1543, 2014.

\bibitem{paszke2017automatic}
Adam Paszke, Sam Gross, Soumith Chintala, Gregory Chanan, Edward Yang, Zachary DeVito, Zeming Lin, Alban Desmaison, Luca Antiga, and Adam Lerer.
\newblock Automatic differentiation in pytorch.
\newblock In {\em https://pytorch.org}, 2017.

\bibitem{jain2018two}
Unnat Jain, Svetlana Lazebnik, and Alexander~G Schwing.
\newblock Two can play this game: visual dialog with discriminative question generation and answering.
\newblock In {\em Proceedings of the IEEE Conference on Computer Vision and Pattern Recognition}, pages 5754--5763, 2018.

\bibitem{lan2019albert}
Zhenzhong Lan, Mingda Chen, Sebastian Goodman, Kevin Gimpel, Piyush Sharma, and Radu Soricut.
\newblock Albert: A lite bert for self-supervised learning of language representations.
\newblock In {\em arXiv preprint arXiv:1909.11942}, 2019.

\bibitem{hori2019end}
Chiori Hori, Huda Alamri, Jue Wang, Gordon Wichern, Takaaki Hori, Anoop Cherian, Tim~K Marks, Vincent Cartillier, Raphael~Gontijo Lopes, Abhishek Das, et~al.
\newblock End-to-end audio visual scene-aware dialog using multimodal attention-based video features.
\newblock In {\em Proceedings of the IEEE International Conference on Acoustics, Speech and Signal Processing}, pages 2352--2356, 2019.

\bibitem{legg2020eccv}
L.~Yeung, Y.~Bisk, and O.~Polozov.
\newblock Alfred speaks: Automatic instruction generation for egocentric skill learning.
\newblock In {\em https://askforalfred.com/EVAL}, 2020.

\bibitem{singh2020eccv}
K.~P. Singh, S.~Bhambri, B.~Kim, , and J.~Choi.
\newblock Improving mask prediction for long horizon instruction following.
\newblock In {\em https://askforalfred.com/EVAL}, 2020.

\bibitem{kempka2016vizdoom}
M.~Kempka, M.~Wydmuch, G.~Runc, J.~Toczek, and W.~Ja{\'s}kowski.
\newblock Vizdoom: A doom-based ai research platform for visual reinforcement learning.
\newblock In {\em Proceedings of the IEEE Conference on Computational Intelligence and Games}, 2016.

\bibitem{ai2thor}
E.~Kolve, R.~Mottaghi, W.~Han, E.~VanderBilt, L.~Weihs, A.~Herrasti, D.~Gordon, Y.~Zhu, A.~Gupta, and A.~Farhadi.
\newblock {AI2-THOR: An Interactive 3D Environment for Visual AI}.
\newblock In {\em arXiv:1712.05474}, 2017.

\bibitem{wu2018building}
Yi~Wu, Yuxin Wu, Georgia Gkioxari, and Yuandong Tian.
\newblock Building generalizable agents with a realistic and rich 3d environment.
\newblock In {\em Proceedings of International Conference on Learning Representations}, 2018.

\bibitem{chen2019touchdown}
H.~Chen, A.~Suhr, D.~Misra, N.~Snavely, and Y.~Artzi.
\newblock Touchdown: Natural language navigation and spatial reasoning in visual street environments.
\newblock In {\em Proceedings of the IEEE Conference on Computer Vision and Pattern Recognition}, 2019.

\bibitem{hermann2020learning}
K.~M. Hermann, M.~Malinowski, P.~Mirowski, A.~Banki-Horvath, K.~Anderson, and R.~Hadsell.
\newblock Learning to follow directions in street view.
\newblock In {\em Proceedings of the AAAI Conference on Artificial Intelligence}, 2020.

\bibitem{chang2017matterport3d}
A.~Chang, A.~Dai, T.~Funkhouser, M.~Halber, M.~Niessner, M.~Savva, S.~Song, A.~Zeng, and Y.~Zhang.
\newblock Matterport3d: Learning from rgb-d data in indoor environments.
\newblock In {\em Proceedings of International Conference on 3D Vision}, 2017.

\bibitem{wang2019reinforced}
X.~Wang, Q.~Huang, A.~Celikyilmaz, J.~Gao, D.~Shen, Y.-F. Wang, W.~Y. Wang, and L.~Zhang.
\newblock Reinforced cross-modal matching and self-supervised imitation learning for vision-language navigation.
\newblock In {\em Proceedings of the IEEE Conference on Computer Vision and Pattern Recognition}, 2019.

\bibitem{tan2019learning}
H.~Tan, L.~Yu, and M.~Bansal.
\newblock Learning to navigate unseen environments: Back translation with environmental dropout.
\newblock In {\em Proceedings of Conference of the North American Chapter of the Association for Computational Linguistics}, 2019.

\bibitem{majumdar2020improving}
A.~Majumdar, A.~Shrivastava, S.~Lee, P.~Anderson, D.~Parikh, and D.~Batra.
\newblock Improving vision-and-language navigation with image-text pairs from the web.
\newblock In {\em Proceedings of the European Conference on Computer Vision}, 2020.

\bibitem{Nguyen_2019_CVPR}
K.~Nguyen, D.~Dey, C.~Brockett, and B.~Dolan.
\newblock Vision-based navigation with language-based assistance via imitation learning with indirect intervention.
\newblock In {\em Proceedings of the IEEE Conference on Computer Vision and Pattern Recognition}, 2019.

\bibitem{thomason2020vision}
J.~Thomason, M.~Murray, M.~Cakmak, and L.~Zettlemoyer.
\newblock Vision-and-dialog navigation.
\newblock In {\em Proceedings of Conference on Robot Learning}, 2020.

\bibitem{suhr-etal-2019-executing}
Alane Suhr, Claudia Yan, Jacob Schluger, Stanley Yu, Hadi Khader, Marwa Mouallem, Iris Zhang, and Yoav Artzi.
\newblock Executing instructions in situated collaborative interactions.
\newblock {\em arXiv preprint arXiv:1910.03655}, 2019.

\bibitem{krantz2020beyond}
J.~Krantz, E.~Wijmans, A.~Majumdar, D.~Batra, and S.~Lee.
\newblock Beyond the nav-graph: Vision-and-language navigation in continuous environments.
\newblock In {\em Proceedings of the European Conference on Computer Vision}, 2020.

\bibitem{zhu2017visual}
Y.~Zhu, D.~Gordon, E.~Kolve, D.~Fox, L.~Fei-Fei, A.~Gupta, R.~Mottaghi, and A.~Farhadi.
\newblock Visual semantic planning using deep successor representations.
\newblock In {\em Proceedings of the IEEE International Conference on Computer Vision}, 2017.

\bibitem{gordon2019should}
D.~Gordon, D.~Fox, and A.~Farhadi.
\newblock What should i do now? marrying reinforcement learning and symbolic planning.
\newblock In {\em arXiv:1901.01492}, 2019.

\bibitem{embodiedqa}
A.~Das, S.~Datta, G.~Gkioxari, S.~Lee, D.~Parikh, and D.~Batra.
\newblock {E}mbodied {Q}uestion {A}nswering.
\newblock In {\em Proceedings of the IEEE Conference on Computer Vision and Pattern Recognition}, 2018.

\bibitem{gordon2018iqa}
D.~Gordon, A.~Kembhavi, M.~Rastegari, J.~Redmon, D.~Fox, and A.~Farhadi.
\newblock Iqa: Visual question answering in interactive environments.
\newblock In {\em Proceedings of the IEEE Conference on Computer Vision and Pattern Recognition}, 2018.

\bibitem{wijmans2019embodied}
E.~Wijmans, S.~Datta, O.~Maksymets, A.~Das, G.~Gkioxari, S.~Lee, I.~Essa, D.~Parikh, and D.~Batra.
\newblock Embodied question answering in photorealistic environments with point cloud perception.
\newblock In {\em Proceedings of the IEEE Conference on Computer Vision and Pattern Recognition}, 2019.

\bibitem{puig2018virtualhome}
X.~Puig, K.~Ra, M.~Boben, J.~Li, T.~Wang, S.~Fidler, and A.~Torralba.
\newblock Virtualhome: Simulating household activities via programs.
\newblock In {\em Proceedings of the IEEE Conference on Computer Vision and Pattern Recognition}, 2018.

\bibitem{corona2020modularity}
R.~Corona, D.~Fried, C.~Devin, D.~Klein, and T.~Darrell.
\newblock Modularity improves out-of-domain instruction following.
\newblock In {\em arXiv:2010.12764}, 2020.

\bibitem{singh2020moca}
K.~P. Singh, S.~Bhambri, B.~Kim, R.~Mottaghi, and J.~Choi.
\newblock Moca: A modular object-centric approach for interactive instruction following.
\newblock In {\em arXiv:2012.03208}, 2020.

\bibitem{shridhar2021alfworld}
Mohit Shridhar, Xingdi Yuan, Marc-Alexandre Cote, Yonatan Bisk, Adam Trischler, and Matthew Hausknecht.
\newblock {\{}ALFW{\}}orld: Aligning text and embodied environments for interactive learning.
\newblock In {\em Proceedings of International Conference on Learning Representations}, 2021.

\bibitem{cote18textworld}
Marc-Alexandre C\^ot\'e, \'Akos K\'ad\'ar, Xingdi Yuan, Ben Kybartas, Tavian Barnes, Emery Fine, James Moore, Ruo~Yu Tao, Matthew Hausknecht, Layla~El Asri, Mahmoud Adada, Wendy Tay, and Adam Trischler.
\newblock Textworld: A learning environment for text-based games.
\newblock In {\em CoRR}, volume abs/1806.11532, 2018.

\bibitem{devin2018deep}
C.~Devin, P.~Abbeel, T.~Darrell, and S.~Levine.
\newblock Deep object-centric representations for generalizable robot learning.
\newblock In {\em IEEE International Conference on Robotics and Automation}, 2018.

\bibitem{srivastava2014dropout}
N.~Srivastava, G.~Hinton, A.~Krizhevsky, I.~Sutskever, and R.~Salakhutdinov.
\newblock Dropout: a simple way to prevent neural networks from overfitting.
\newblock In {\em The journal of machine learning research}, volume~15, pages 1929--1958, 2014.

\bibitem{nguyenefficient}
{V. Q.} Nguyen, M.~Suganuma, and T.~Okatani.
\newblock Efficient attention mechanism for visual dialog that can handle all the interactions between multiple inputs.
\newblock In {\em Proceedings of the European Conference on Computer Vision}, 2020.

\bibitem{Anderson2018OnEO}
P.~Anderson, A.~X. Chang, D.~S. Chaplot, A.~Dosovitskiy, S.~Gupta, V.~Koltun, J.~Kosecka, J.~Malik, R.~Mottaghi, M.~Savva, and A.~R. Zamir.
\newblock On evaluation of embodied navigation agents.
\newblock In {\em arXiv:1807.06757}, 2018.

\bibitem{deitke2020robothor}
Matt Deitke, Winson Han, Alvaro Herrasti, Aniruddha Kembhavi, Eric Kolve, Roozbeh Mottaghi, Jordi Salvador, Dustin Schwenk, Eli VanderBilt, Matthew Wallingford, et~al.
\newblock Robothor: An open simulation-to-real embodied ai platform.
\newblock In {\em Proceedings of the IEEE/CVF conference on computer vision and pattern recognition}, pages 3164--3174, 2020.

\bibitem{bharadhwaj2019data}
Homanga Bharadhwaj, Zihan Wang, Yoshua Bengio, and Liam Paull.
\newblock A data-efficient framework for training and sim-to-real transfer of navigation policies.
\newblock In {\em 2019 International Conference on Robotics and Automation (ICRA)}, pages 782--788. IEEE, 2019.

\bibitem{anderson2021sim}
Peter Anderson, Ayush Shrivastava, Joanne Truong, Arjun Majumdar, Devi Parikh, Dhruv Batra, and Stefan Lee.
\newblock Sim-to-real transfer for vision-and-language navigation.
\newblock In {\em Conference on Robot Learning}, pages 671--681. PMLR, 2021.

\bibitem{vanalgorithm}
Nguyen Van~Quang.
\newblock An algorithm adaptation multi-label classification method and experiments on vietnamese text.

\bibitem{pham2017mass}
Thi-Ngan Pham, Van-Quang Nguyen, Duc-Trong Dinh, Tri-Thanh Nguyen, and Quang-Thuy Ha.
\newblock Mass: a semi-supervised multi-label classification algorithm with specific features.
\newblock In {\em Asian Conference on Intelligent Information and Database Systems}, pages 37--47. Springer, 2017.

\bibitem{pham2017semi}
Thi-Ngan Pham, Van-Quang Nguyen, Van-Hien Tran, Tri-Thanh Nguyen, and Quang-Thuy Ha.
\newblock A semi-supervised multi-label classification framework with feature reduction and enrichment.
\newblock {\em Journal of Information and Telecommunication}, 1(4):305--318, 2017.

\bibitem{ha2018newlifelong}
Quang-Thuy Ha, Thi-Ngan Pham, Van-Quang Nguyen, Thi-Cham Nguyen, Thi-Hong Vuong, Minh-Tuoi Tran, and Tri-Thanh Nguyen.
\newblock A new lifelong topic modeling method and its application to vietnamese text multi-label classification.
\newblock In {\em Asian Conference on Intelligent Information and Database Systems}, pages 200--210. Springer, 2018.

\bibitem{ha2018newtext}
Quang-Thuy Ha, Thi-Ngan Pham, Van-Quang Nguyen, Minh-Chau Nguyen, Thanh-Huyen Pham, and Tri-Thanh Nguyen.
\newblock A new text semi-supervised multi-label learning model based on using the label-feature relations.
\newblock In {\em International Conference on Computational Collective Intelligence}, pages 403--413. Springer, 2018.

\bibitem{van2019capsulenet}
Nguyen Van~Quang, Jinhee Chun, and Takeshi Tokuyama.
\newblock Capsulenet for micro-expression recognition.
\newblock In {\em 2019 14th IEEE International Conference on Automatic Face \& Gesture Recognition (FG 2019)}, pages 1--7. IEEE, 2019.

\bibitem{van2019revisiting}
Nguyen Van~Quang and Hiromasa Fujihara.
\newblock Revisiting a single-stage method for face detection.
\newblock In {\em 2019 14th IEEE International Conference on Automatic Face \& Gesture Recognition (FG 2019)}, pages 1--8. IEEE, 2019.

\bibitem{van2019learning}
Nguyen Van~Quang.
\newblock {\em Learning Discriminative and Generative tasks for Visual Dialog}.
\newblock PhD thesis, TOHOKU UNIVERSITY, 2019.

\bibitem{nguyen2020efficient}
Van-Quang Nguyen, Masanori Suganuma, and Takayuki Okatani.
\newblock Efficient attention mechanism for visual dialog that can handle all the interactions between multiple inputs.
\newblock In {\em European Conference on Computer Vision}, pages 223--240. Springer, 2020.

\bibitem{nguyen2021look}
Van-Quang Nguyen, Masanori Suganuma, and Takayuki Okatani.
\newblock Look wide and interpret twice: Improving performance on interactive instruction-following tasks.
\newblock {\em arXiv preprint arXiv:2106.00596}, 2021.

\bibitem{nguyen2022grit}
Van-Quang Nguyen, Masanori Suganuma, and Takayuki Okatani.
\newblock Grit: Faster and better image captioning transformer using dual visual features.
\newblock In {\em European Conference on Computer Vision}, pages 167--184. Springer, 2022.

\bibitem{pham2024ktvic}
Anh-Cuong Pham, Van-Quang Nguyen, Thi-Hong Vuong, and Quang-Thuy Ha.
\newblock Ktvic: A vietnamese image captioning dataset on the life domain.
\newblock {\em arXiv preprint arXiv:2401.08100}, 2024.

\bibitem{charoenpitaks2024exploring}
Korawat Charoenpitaks, Van-Quang Nguyen, Masanori Suganuma, Masahiro Takahashi, Ryoma Niihara, and Takayuki Okatani.
\newblock Exploring the potential of multi-modal ai for driving hazard prediction.
\newblock {\em IEEE Transactions on Intelligent Vehicles}, 2024.

\bibitem{areerob2025multimodal}
Kittitouch Areerob, Van-Quang Nguyen, Xianfeng Li, Shogo Inadomi, Toru Shimada, Hiroyuki Kanasaki, Zhijie Wang, Masanori Suganuma, Keiji Nagatani, Pang-jo Chun, et~al.
\newblock Multimodal artificial intelligence approaches using large language models for expert-level landslide image analysis.
\newblock {\em Computer-Aided Civil and Infrastructure Engineering}, 40(19):2900--2921, 2025.

\bibitem{charoenpitaks2025tb}
Korawat Charoenpitaks, Van-Quang Nguyen, Masanori Suganuma, Kentaro Arai, Seiji Totsuka, Hiroshi Ino, and Takayuki Okatani.
\newblock Tb-bench: Training and testing multi-modal ai for understanding spatio-temporal traffic behaviors from dashcam images/videos.
\newblock In {\em 2025 IEEE/CVF Conference on Computer Vision and Pattern Recognition Workshops (CVPRW)}, pages 2436--2446. IEEE, 2025.

\bibitem{nguyen2026coretab}
Van-Quang Nguyen and Takayuki Okatani.
\newblock Coretab: Improving multimodal table understanding with code-driven reasoning.
\newblock {\em arXiv preprint arXiv:2601.19193}, 2026.

\bibitem{tran2026360}
Huyen~TT Tran, Van-Quang Nguyen, Farros Alferro, Kang-Jun Liu, and Takayuki Okatani.
\newblock 360$^{\circ}$ image perception with mllms: A comprehensive benchmark and a training-free method.
\newblock {\em arXiv preprint arXiv:2603.16179}, 2026.

\end{thebibliography}

%
%
%
\cleardoublepage

\begin{acknowledgments}

I owe a debt of gratitude to many people for making my years as a Ph.D. student meaningful and memorable. This journey would not have been so memorable, and I would not have completed this dissertation without them.
First and foremost, I must express my gratitude to my supervisor, Professor Takayuki Okatani. We have discussed numerous meetings and emails over the last three years. He has patiently listened to all of my progress reports, regardless of whether they were a mixture of messed-up text, invalid results, or inferior ideas. He has constantly given me comments and asked important questions about my research. Professor Okatani spent days and weeks carefully reading my manuscripts during the submission deadline period, and I can recall it vividly. He then distilled the essence, articulated and contextualized ideas, and transformed all manuscripts into greater ones. I admire his pursuit of excellence and clarity and acknowledge his support for students like myself. It must be said that, as of now, I have achieved many fruitful results thanks to his constant support. I am grateful to him for serving as my advisor.
 
During my preliminary and final defenses, other jury members on my dissertation committee, Professor Koichi Hashimoto, Professor Kentaro Inui, and Associate Professor Shingo Kagami, provided insightful comments and suggestions.
 
In addition, I'd like to thank Assistant Professor Masanori Suganuma. He has been an outstanding collaborator and communicator and has assisted me with my research on numerous occasions. He has also offered me a lot of advice on research, career, and everyday life. Mrs. Sakane Akemi, the laboratory secretary, has helped assist other students and me with our paperwork requirements. With her assistance, things have become simpler. I have had the good fortune to meet and become friends with numerous others. All my labmates, including Dang Anh Chuong, Kang Jun Liu, Earth, Kitto, and Luo, have shared joy and happiness with me during our time together. I value all my university and non-university friends, and I cherish the time we spent together in Sendai, where I studied for the past five years.
 
I would like to thank the Japanese government and Tohoku University for selecting me as a MEXT scholarship recipient during my five years at Tohoku University.
 
Dinh Thu Hien, my sweetheart, deserves my heartfelt gratitude. She has made this long journey more enjoyable and meaningful by providing companionship and encouragement. Furthermore, I want to express my enduring gratitude to my parents and siblings for their unconditional love and support throughout the writing of this thesis and my life in general. They have been my inspiration and motivation for every step I have taken.
 
Finally, I dedicate this work to myself, who has previously studied and worked tirelessly through ups and downs. Any errors or deficiencies in my dissertation that may still exist are entirely my fault and responsibility.

\end{acknowledgments}

\cleardoublepage

\chapter*{List of Publications}

The following publications were produced during my five-year Data Science Program at Tohoku University.

During the first two years of the program, I had the opportunity to collaborate with researchers at Vietnam National University on foundational problems in text mining and machine learning. This period produced work on semi-supervised multi-label classification, where we developed algorithms that leverage both labeled and unlabeled data to improve classification performance with limited supervision \cite{vanalgorithm, pham2017mass, pham2017semi}. Extending this line of research, we further explored lifelong topic modeling and label-feature relationships for Vietnamese text multi-label classification, yielding two conference papers that addressed the challenges of continual learning and structural feature exploitation in natural language processing \cite{ha2018newlifelong, ha2018newtext}. These early contributions deepened my understanding of learning under data scarcity and the role of auxiliary information in improving model generalization.

Alongside this collaborative work, I pursued two independent first-author directions in computer vision. The first proposed a CapsuleNet for facial micro-expression recognition, evaluated on the cross-database benchmark combining SMIC, CASME~II, and SAMM \cite{van2019capsulenet}. The second revisited single-stage face detection: building on ResNet-101, the model reduced false positives via context from deeper layers and achieved approximately 26 ms inference on VGA images, with competitive results on Wider Face, AFW, Pascal Face, and FDDB \cite{van2019revisiting}.

The main thrust of my doctoral research, conducted under the supervision of Professor Takayuki Okatani, focused on multi-modal understanding at the intersection of vision and language. Building upon my master's thesis on visual dialog \cite{van2019learning}, I developed an efficient attention mechanism capable of handling all pairwise interactions among multiple heterogeneous inputs, including the image, dialog history, and the current question. This work was published at the European Conference on Computer Vision (ECCV~2020) \cite{nguyen2020efficient}. Motivated by the challenges of interactive instruction-following, I subsequently proposed a method that widens the field of visual attention and interprets language inputs at multiple levels of granularity, published at the International Joint Conference on Artificial Intelligence (IJCAI~2021) \cite{nguyen2021look}. The culminating contribution of my doctoral program was GRIT, a faster and more accurate image captioning transformer that exploits dual visual features---grid features for dense spatial coverage and region features for semantic grounding---published at ECCV~2022 \cite{nguyen2022grit}. This work demonstrated that jointly leveraging complementary visual representations leads to substantial improvements in both captioning quality and inference speed.

The research directions established during my doctoral studies have continued to generate collaborative contributions in subsequent years on vision-language understanding, especially the applications multimoal large language models on various specific domains. These include a Vietnamese image captioning dataset covering diverse life-domain scenarios \cite{pham2024ktvic}, multi-modal artificial intelligence for driving hazard prediction \cite{charoenpitaks2024exploring}, expert-level landslide image analysis using large language models \cite{areerob2025multimodal}, and a training and testing benchmark for understanding spatio-temporal traffic behaviors from dashcam images and videos \cite{charoenpitaks2025tb}. More recently, I have contributed to advancing the capabilities of multimodal large language models, including a method for improving table understanding through code-driven reasoning \cite{nguyen2026coretab} and a comprehensive benchmark and training-free approach for 360-degree image perception \cite{tran2026360}.

\cleardoublepage
\end{document}